\documentclass[10pt,journal,compsoc]{IEEEtran}
\usepackage{amsmath,amssymb,amsfonts}
\usepackage{algorithmic}
\usepackage{graphicx}
\usepackage{textcomp}
\usepackage{epstopdf}
\usepackage{bm}
\usepackage{makecell}
\usepackage{multirow}
\usepackage{array}
\usepackage{booktabs}
\usepackage{color}
\usepackage{float}
\usepackage{subcaption}
\usepackage{bm}

\usepackage[capitalize]{cleveref}
\crefname{section}{Sec.}{Secs.}
\Crefname{section}{Section}{Sections}
\Crefname{table}{Table}{Tables}
\crefname{table}{Tab.}{Tabs.}

\ifCLASSOPTIONcompsoc
  \usepackage[nocompress]{cite}
\else
  \usepackage{cite}
\fi

\begin{document}
%
\title{Representation Separation for Semantic Segmentation with Vision Transformers}
%
%
%
%

\author{Yuanduo Hong, Huihui Pan,~\IEEEmembership{Member,~IEEE}, Weichao Sun,~\IEEEmembership{Senior Member,~IEEE}, Xinghu Yu, and Huijun Gao,~\IEEEmembership{Fellow,~IEEE}
\IEEEcompsocitemizethanks{\IEEEcompsocthanksitem The authors are with Research Institute of Intelligent Control and Systems,
Harbin Institute of Technology, Harbin 150001, China (e-mail: hjgao@hit.edu.cn)}}

%
%

\markboth{Journal of \LaTeX\ Class Files,~Vol.~14, No.~8, August~2015}%
{Shell \MakeLowercase{\textit{et al.}}: Bare Demo of IEEEtran.cls for Computer Society Journals}
%



\IEEEtitleabstractindextext{%
\begin{abstract}
Vision transformers (ViTs) encoding an image as a sequence of patches bring new paradigms for semantic segmentation.
We present an efficient framework of representation separation in local-patch level and global-region level for semantic segmentation with ViTs. It is targeted for the peculiar over-smoothness of ViTs in semantic segmentation, and therefore differs from current popular paradigms of context modeling and most existing related methods reinforcing the advantage of attention. We first deliver the decoupled two-pathway network in which another pathway enhances and passes down local-patch discrepancy complementary to global representations of transformers. We then propose the spatially adaptive separation module to obtain more separate deep representations and the discriminative cross-attention which yields more discriminative region representations through novel auxiliary supervisions. The proposed methods achieve some impressive results: 1) incorporated with large-scale plain ViTs, our methods achieve new state-of-the-art performances on five widely used benchmarks; 2) using masked pre-trained plain ViTs, we achieve 68.9 $\%$ mIoU on Pascal Context, setting a new record; 3) pyramid ViTs integrated with the decoupled two-pathway network even surpass the well-designed high-resolution ViTs on Cityscapes; 4) the improved representations by our framework have favorable transferability in images with natural corruptions. The codes will be released publicly.
\end{abstract}

}

\maketitle

\IEEEdisplaynontitleabstractindextext

%
\IEEEpeerreviewmaketitle

\IEEEraisesectionheading{\section{Introduction}\label{sec:introduction}}

\IEEEPARstart{S}{emantic} segmentation is one of the fundamental tasks related to image understanding and is indispensable for many promising technologies and applications including medical image processing and environment perception of autonomous vehicles and robots. The key to semantic segmentation is allocating correct labels to pixels of objects belonging to the same class. FCN\cite{long2015fully}, a seminal work, utilizes convolutional neural networks (CNNs) to directly predict the dense labels of pixels, having a far-reaching influence on follow-up works. Previous works show that the capacity of capturing long-range context or even multi-scale context plays a significant role in high-accuracy semantic segmentation\cite{chen2017deeplab,chen2017rethinking,liu2015parsenet,zhao2017pyramid,peng2017large}. DeepLab\cite{chen2017deeplab} introduces dilation convolutions for semantic segmentation which augment the receptive fields of vanilla convolutions without extra computation. Larger convolutional kernels\cite{peng2017large} and global pooling\cite{liu2015parsenet,zhao2017pyramid} are also powerful tools to establish long-range relations.

\begin{figure}
\centerline{\includegraphics[width=\columnwidth]{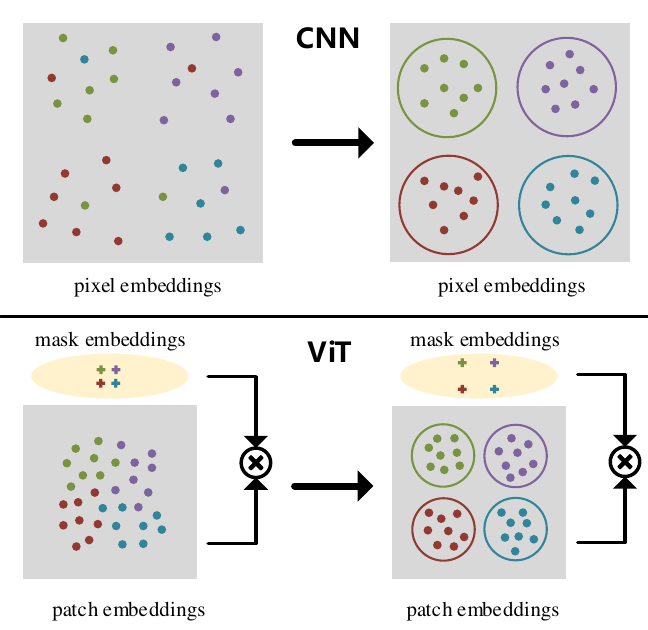}}
\caption{The comparison between our representation separation for ViTs and the context modeling for CNNs. The representations learned by CNNs are usually dispersive due to the limited receptive fields. The context modeling modules learn more discriminative representations by reducing the intra-class variances. Compared to CNNs, the learned representations of ViTs are more coherent and even too close to distinguish different categories. The proposed methods separate the representations of different categories while maintaining the small intra-class variances.}
\label{fig1}
\end{figure}

Alternatively, self-attention mechanism dynamically establishes the global dependencies, which better captures the contextual information in semantic segmentation\cite{fu2019dual,yuan2021ocnet,yin2020disentangled}. Recently, some works introduce the transformer\cite{vaswani2017attention} only containing the self-attention modules and the MLP layers for vision tasks, which shows great potential. Dosovitskiy \emph{et al.}\cite{dosovitskiy2020image} transfer the intact encoder module to image classification and demonstrate the superiority of pure transformer architecture with a mass of training data. They first split the image into patches and generate token embeddings with a trainable linear projection. Position embeddings and a learning embedding are added to retain positional information and generate classification results. DeiT\cite{touvron2020training} improves the training strategy and achieves impressive accuracy on ImageNet validation only using ImageNet-1k for training. On top of this, some fundamental architecture improvements like pyramid features and shifted windows are proposed\cite{wang2021pyramid,liu2021swin} and meanwhile, some works transfer the powerful ViT backbones to semantic segmentation\cite{zheng2021rethinking,xie2021segformer,ranftl2021vision}. Since the impressive performances on image classification and intrinsic global modeling, ViTs outperform CNNs by a large margin without fancy decoders.

Although vision transformers have shown excellent transferability in semantic segmentation, there are essential differences between image classification and dense prediction. Following the routine, extra components coupled with multi-scale representations and attention variants are utilized to further boost the performance of transformer based semantic segmentation\cite{strudel2021segmenter,yan2022lawin,jain2021semask,lin2022structtoken,chen2022vision}. Given that vision transformers have possessed large enough receptive fields, we would like to rethink the segmentation transformers from a new perspective. Some existing works analyze the feature maps of transformers on image classification tasks and shed light on the low-pass characteristic of self-attention and the high similarity of patch embeddings\cite{park2021vision,gong2021vision}. Inspired by them, we empirically find that even the advanced semantic segmentation models with transformers suffer from the smooth segmentation compared to CNN based methods. The corresponding results can be found in Section \ref{A Review of Vision Transformers}.


In this paper, we explore the efficient solutions from the viewpoint of the representational discrimination. Through the above analysis, we argue that enhancing the representational discrepancy is particularly significant for semantic segmentation with ViTs due to the blurry feature maps and suppose that the discrepancy can be restored from the pre-trained representations only using the data of downstream tasks. It allows us to achieve the efficient utilization of pre-trained vision transformers, saving the pre-training cost through decoupling the transformer backbone design from downstream tasks.
We start from the general architecture of plain ViTs, providing three underlying reasons for the high similarity of patch embeddings: low-resolution feature maps after the first layer, completely global modeling, and low-pass spatially global filters. To this end, we introduce a representation separation framework consist of three novel components. To extract the local discrepancy of patch embeddings, we propose the decoupled two-pathway network which contains a local separation path to enhance and pass down the local-patch discrepancy. Specially, we build cascaded local separation blocks (LSBs) with the learnable high-pass filters (LHFs) to separate local patch embeddings and continually merge the local discrepancy into next semantically rich representations adaptively by attention-guided fusion (AF). The proposed two-pathway network allows for decoupling the local information from the pre-trained global representations and yields enhanced global representations with richer local details. Since the output of the last layer usually has the best transferability for plain ViTs, we further propose the spatially adaptive separation module (SASM) which guides the representation separation of the final output through local information of the decoupled two-pathway network. Specially, we convert the output of the local separation path to spatially adaptive filters applied on the last patch embeddings and expand the resolution by pixel shuffle. The discrepancy of patch embeddings is obviously enhanced due to the guidance of distinctly local information and the growing resolution of feature maps. In Segmenter\cite{strudel2021segmenter}, learnable queries are utilized to obtain dynamic mask embeddings through extra plain ViTs layers. The final segmentation masks can be acquired via a simple matrix multiplication of mask embeddings and patch embeddings. Inspired by it, we propose the discriminative cross-attention which enables each query pay more attention on the corresponding region of keys through two novel auxiliary supervisions and thus leads to more discriminatively aggregated region representations, namely mask embeddings. Inside it, the query-to-region matching loss lets each query vector directly match with the corresponding region embedding and the patch-to-region matching loss ensures the discrimination of keys. Finally, the proposed framework incorporates local representation separation of patch embeddings and global representation separation of mask embeddings, as shown in \cref{fig1}.

\begin{figure}
\centerline{\includegraphics[width=\columnwidth]{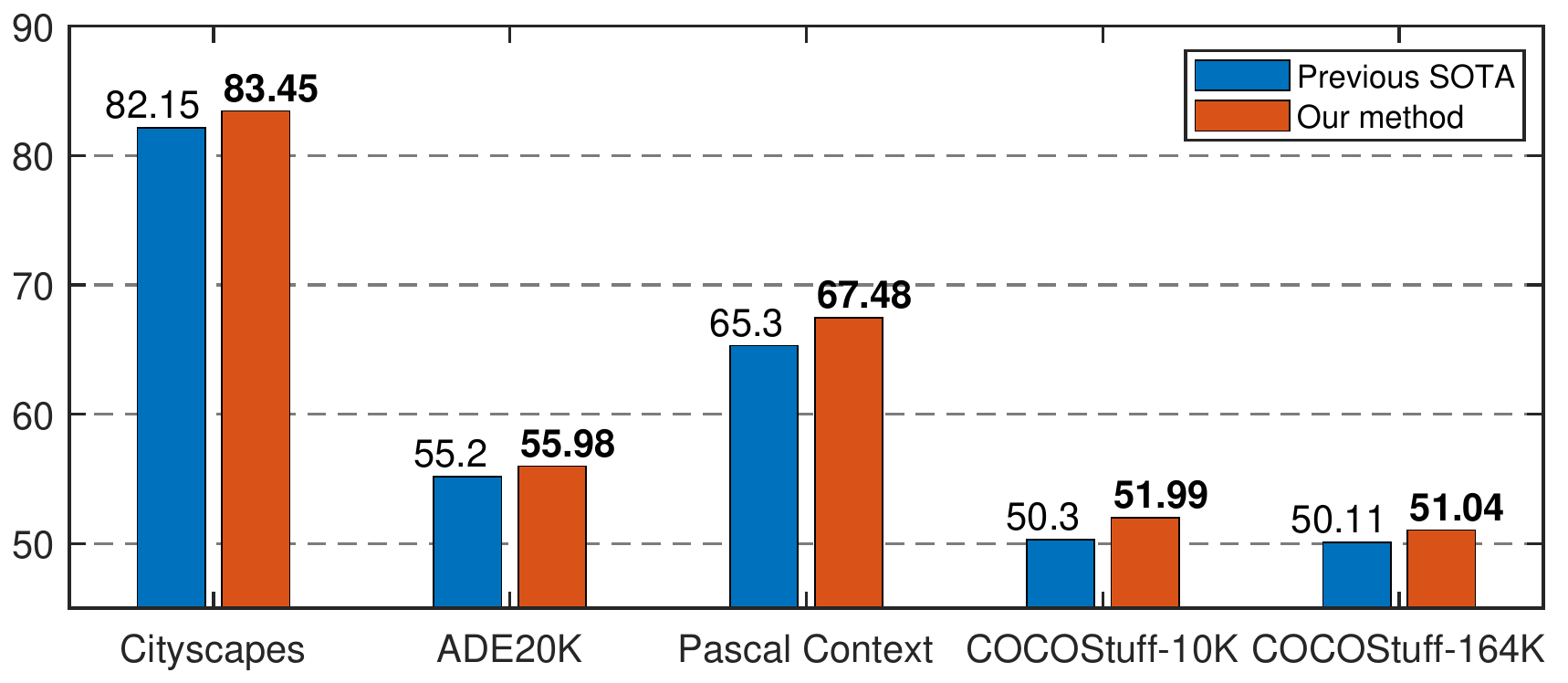}}
\caption{Comparisons with previous state-of-the-art methods on five benchmarks. All the results are obtained with ViT-L\cite{steiner2021train} and multi-scale test.}
\label{fig2}
\end{figure}

We integrate the proposed representation separation framework with large-scale plain ViTs (named RSSeg-ViT) and evaluate them on five widely used benchmarks, \emph{i.e.}, Cityscapes, ADE20K, Pascal Context, COCOStuff-10K, and COCOStuff-164K. The experimental results show that our method achieves new state-of-the-art accuracy on five datasets as show in Fig. \ref{fig2} and greatly improve the powerful self-attention baselines by yielding more distinct segmentation maps as shown in Section \ref{Visualized Results}. Using BEiT-L\cite{bao2021beit}, a large-scale masked pre-trained plain ViT, we achieve 68.9$\%$ mIoU on Pascal Context. To our best knowledge, it performs better than all the previous methods including the much more cumbersome models. We also apply the decoupled two-pathway network to advanced pyramid ViTs, outperforming the previous state-of-the-art results achieved by the well-designed high-resolution ViTs on Cityscapes. By evaluating the robustness to natural corruptions of the proposed method, we demonstrate that it can further improve such robustness upon pure attention networks.

Our contributions are summarized as follows:
\begin{itemize}
  \item We rethink semantic segmentation with ViTs from a different perspective and propose an efficient framework of representation separation in the local-patch level and global-region level. To our best knowledge, we are the first to systematically study representation separation for semantic segmentation with ViTs.
  \item The decoupled two-pathway network is proposed to enhance the local-patch discrepancy of the relatively smooth feature maps, which incorporates the novel learnable high-pass filters and attention-guided fusion mechanism.
  \item The spatially adaptive separation module is proposed to obtain more separate deep representations. It utilizes the local-patch discrepancy to guide the representation separation of the output of the last transformer layer.
  \item To obtain more discriminative mask embeddings, discriminative cross-attention is proposed to ensure the discriminative match between queries and keys by two novel auxiliary supervisions.
  \item Integrated with large-scale plain ViTs, the proposed method achieves new state-of-the-art accuracy on five benchmarks. With BEiT-L as the backbone, we achieve the best performance so far on Pascal Context with the mIoU of 68.9$\%$. Using advanced pyramid ViTs, our method even surpasses the well-designed high-resolution ViTs on Cityscapes which contains numerous small/thin objects.
  \item The robustness to natural corruptions of plain ViTs is explored and our representation separation framework can further improve such robustness upon pure attention networks.

\end{itemize}

\section{Related Work}

\subsection{Vision Transformers}

ViT\cite{dosovitskiy2020image} is a great success for applying original transformers to vision tasks, whereas it has huge model complexity and needs to be trained with hundreds of millions of images. DeiT\cite{touvron2020training} alleviates this problem by introducing an efficient training strategy and proposes a token-based distillation method. Wang \emph{et al.}\cite{wang2021pyramid} and Liu \emph{et al.} \cite{liu2021swin} propose the hierarchical vision transformers, the straight substitution for CNN backbones in various pixel-level dense prediction tasks.
Numerous architectural improvements have been proposed to introduce the locality for global attention\cite{chu2021conditional,xie2021segformer,chu2021twins,dai2021coatnet, dong2021cswin,yuan2022volo}.
Inspired by HRNet, Yuan \emph{et al.}\cite{yuan2021hrformer} propose high-resolution multi-branch transformers and Gu \emph{et al.}\cite{gu2021multi} deliver the more efficient variants which achieve the state-of-the-art performance on Cityscapes dataset. In the meanwhile, the potential of originally plain ViTs is constantly excavated by improving training recipes\cite{steiner2021train,touvron2022deit}. The masked transformers turn out to be scalable vision learners via valid self-supervised learning\cite{he2021masked,bao2021beit}. These improvements in training recipes or learning patterns give birth to superior plain ViTs with big transferability to semantic segmentation. For example, BEiT-L\cite{bao2021beit} achieves 57.0$\%$ mIoU on ADE20K without extra bells or whistles. Our method can directly employ these potent plain ViTs as backbones, achieving remarkable performance gains upon strong baselines. In addition, some designs like \cite{yuan2021hrformer,gu2021multi} inevitably increase the pre-training cost while our method may be a substitute for such methods without retraining the models on ImageNet.

\subsection{Semantic Segmentation with ViTs}
Zheng \emph{et al.}\cite{zheng2021rethinking} first explore plain ViTs for semantic segmentation and propose SETR which achieved state-of-the-art performance on challenging scene parsing tasks at that time. Ranftl \emph{et al.}\cite{ranftl2021vision} propose the U-shaped dense vision transformers of which encoders are ViTs (ViT-Hybrid) and decoders are made up of convolutional layers. In Segmenter \cite{strudel2021segmenter}, the proposed mask transformer takes as input the output of ViT and learnable class embeddings to predict segmentation masks. Xie \emph{et al.}\cite{xie2021segformer} improve plain ViTs through hierarchical feature maps, overlapped convolutional embeddings, efficient self-attention, and inserting 3$\times$3 depth-wise convolutions into the feed forward network (FFN). Benefitting by the potently representational capacity and long-range modeling process of vision transformers, most state-of-the-art methods use the UperNet decoder\cite{xiao2018unified} or directly concatenate the multi-resolution feature maps followed by a simple decoder \cite{liu2021swin,xie2021segformer,dong2021cswin}. Although several concurrent works\cite{strudel2021segmenter,yan2022lawin,jain2021semask,lin2022structtoken,chen2022vision} propose specific methods for semantic segmentation with ViTs, they still rely on the self-attention or cross-attention module design for extracting richer contextual information or utilize multi-scale features to improve the representational capacity. Inspired by DETR\cite{carion2020end}, the frameworks decoupling classification and mask prediction for unified semantic and instance segmentation are proposed\cite{cheng2021per,cheng2022masked}, and are utilized by following semantic segmentation methods\cite{jain2021semask,chen2022vision} to achieve state-of-the-art results. However, the framework becomes more and more complex, which possibly prevents us from finding out what semantic segmentation with ViTs really needs.
Other than the aforementioned works, we start from the over-smoothness of transformer backbones, exploring simple and effective methods from a different perspective, namely representation separation rather than context modeling.


\subsection{Semantic Segmentation with CNNs}

Ahead of vision transformers, the accuracy of semantic segmentation has been improved noticeably benefited from the advanced architectures and training settings of CNNs. From the early encoder-decoder structure\cite{long2015fully,ronneberger2015u} to the dilation backbone\cite{chen2017deeplab}, long-range context capturing modules are helpful supplements due to the locality of convolution operations.
Among them, Pyramid pooling\cite{zhao2017pyramid} and Atrous Spatial Pyramid Pooling (ASPP)\cite{chen2017rethinking} are two widely used tools. Large kernel convolution is first presented for semantic segmentation in \cite{peng2017large} but is rarely used due to the large complexity. Afterward, self-attention modules and their variants\cite{fu2019dual,yuan2021ocnet,9133304,9206128,9633155} are proposed to capture long-range context more flexibly. In addition to context modeling, blending high-resolution features and low-resolution features is a class of efficient methods for generating more fine-grained segmentation results\cite{long2015fully,ronneberger2015u,lin2017refinenet,xiao2018unified,kirillov2019panoptic,li2020semantic}.
Refining the segmentation boundary with the help of specific supervision is also very effective for fine segmentation scenarios \cite{takikawa2019gated,li2020improving,wang2021active,yuan2020segfix}. However, most of the aforementioned methods are unbefitting for the rising vision transformers. On the one hand, vision transformers with cascaded self-attention naturally embody powerful context modeling capacity. On the other hand, feature fusion methods or boundary refinement methods rely on valid high-resolution feature maps of CNN backbones which plain ViTs are not equipped with. Our method qualitatively differs from most methods in CNN era because we separate the highly coherent representations rather than enhance the consistency of existing high-resolution representations.

\subsection{Representation Separation}
As more and more properties of ViTs are explored, several recent works focus on reducing the high similarity of patch embeddings or introducing high-frequency information\cite{gong2021vision,bai2022improving}. However, these methods are mainly applied to image classification task and the improvements of semantic segmentation are achieved by transferring the pre-trained backbones. For example, Gong \emph{et al.}\cite{gong2021vision} reduce the similarity of all the patch embeddings, which possibly impairs the intra-class consistency when applied to semantic segmentation. More recently, Huang \emph{et al.}\cite{huang2022car} propose three loss functions to reduce intra-class variance and maximize inter-class separation of semantic segmentation. Our work differs from their work in terms of motivations, methods, and implementation technologies. Huang \emph{et al.} aim to use class-level information more effectively for semantic segmentation. However, starting from the over-smoothness of plain ViTs, we naturally argue for the significance of representation separation for semantic segmentation with ViTs. In addition, they ignore enhancing the discrepancy of local regions while our framework takes into account the discrepancy of local patch embeddings and global region embeddings jointly. At last, Huang \emph{et al.} explicitly reduce the inter-class similarity via direct supervision. However, it is very difficult to define the inter-class similarity precisely. On the contrary, we guide the network to learn more separate representations by effective architectures as well as auxiliary supervision.


\section{Method}
Since our methods are developed with ViT backbones, we first review the current vision transformers and analyze the immanent cause of over-smoothness. Next, we introduce the details of the unified framework of representation separation in the context of vision transformers. At last, we introduce three proposed components respectively.

\subsection{A Review of Vision Transformers}
\label{A Review of Vision Transformers}

The introduction of vision transformers of which the basic module is the multi-head self-attention (MSA) block followed by two MLP layers brings new opportunities to the field of Computer Vision. The widely used pre-activation transformer layer can be written in the following form:
\begin{equation}
\left\{
\begin{aligned}
&z = MSA(LN(x_{in})) + x_{in}, \\
&x_{out} = MLP(GELU(MLP(LN(z)))) + z,
\end{aligned}
\right.
\label{eq1}
\end{equation}
where LN denotes the Layer Normalization\cite{ba2016layer} and GELU is the abbreviation of the Gaussian Error Linear Unit\cite{hendrycks2016gaussian}.
Among them, the multi-head self-attention mechanism is the key point for global modeling. Given the 1D embedding sequence $X$ and position encodings as input (2D feature maps can be flattened into the 1D sequence), it is firstly projected to the embedding sequences $Q$, $K$, and $V$ $\in$ $\mathbb{R}^{B,N,C}$:
\begin{equation}
Q = XW_{Q}, K = XW_{K}, V = XW_{V},
\label{eq2}
\end{equation}
where $W_{Q}$, $W_{K}$, and $W_{V}$ are learnable parameters of linear projection layers. Multi-head self-attention is then computed as follows:
\begin{equation}
MSA(Q, K, V) = softmax(\frac{QK^T}{\sqrt{d_k}})V,
\label{eq3}
\end{equation}
where $d_k$ is the dimension of $K$.
Since the memory cost and computational complexity of this process is O($N^{2}$), plain ViTs directly deal with a sequence of patch embeddings generated by a single non-overlapping convolutional layer. We argue that the high patch similarity of plain ViTs is mainly caused by three architectural characteristics: drastic reduction of feature resolution at the first layer, completely global modeling, and low-pass attention weights. Note that the third point arises from the positive attention weights wrapped by the $Softmax$ function.

To decrease the memory cost and computational complexity, efficient self-attention variants are proposed by downsampling the $K, V$ or cutting the patch embeddings into multiple small sequences. Based on them, pyramid ViTs gradually aggregate spatial tokens between self-attention blocks and utilize convolutional operations to learn the location information. Due to the high-resolution feature maps in the early stage and the locality introduced by convolutions, pyramid ViTs have shown better performance than plain ViTs in the downstream tasks which rely more on fine-grained features. However, the over-smoothness possibly still exists because the locality introduced by convolutions is limited by the lowering resolution of feature maps and impaired by the posterior self-attention blocks.

\begin{table}[]\scriptsize
\caption{A comparison between CNNs and ViTs in segmentation results of small/thin objects on Cityscapes val set. $mIoU_{s}$ is the average accuracy of seven small objects. All the model weights can be obtained from the open source toolbox\cite{mmseg2020} except for SegFormer-B2 and CSWin-T which we train with a crop size of 768$\times$768}
\label{tab:17}
\begin{tabular}{p{60pt}|p{8pt}<{\centering}p{8pt}<{\centering}p{8pt}<{\centering}p{8pt}<{\centering}p{8pt}<{\centering}p{8pt}<{\centering}p{8pt}<{\centering}p{8pt}<{\centering}|p{8pt}<{\centering}}
\toprule
Model                &\rotatebox{90}{mIoU}        &\rotatebox{90}{pole} &\rotatebox{90}{tlight} &\rotatebox{90}{sign}  &\rotatebox{90}{person} &\rotatebox{90}{rider}  &\rotatebox{90}{motor} &\rotatebox{90}{bike}  &\rotatebox{90}{mIoU$_{s}$}   \\ \midrule
PSPNet(R50)\cite{zhao2017pyramid}& 79.6 &66.4 &73.3 &81.2 &83.0 &64.1 &68.5 &79.1 & \textbf{73.7}\\
SETR-PUP-L\cite{zheng2021rethinking}& 79.3      &61.4 &67.2 &76.2 &80.3 &60.7 &68.7 &75.9 &70.1 \\ \midrule
DLV3+(R101)\cite{chen2018encoder}&80.9     &69.8 &75.0 &82.5 &84.5 &68.1 &70.5 &80.4 &\textbf{75.8} \\
SegFormer-B2\cite{xie2021segformer}&80.7 &66.7 &72.4 &80.2 &83.3 &63.0 &71.6 &79.1 &73.8 \\
CSWin-T\cite{dong2021cswin}&81.5 &66.6 &72.5 &81.0 &83.9 &67.7 &71.2 &79.3 &74.6 \\ \bottomrule
\end{tabular}
\end{table}

\begin{table}[]\scriptsize
\caption{A comparison between CNNs and ViTs in segmentation results of small objects on ADE20K val set. All the model weights can be obtained from the open source toolbox\cite{mmseg2020}}
\label{tab:18}
\begin{tabular}{p{60pt}|p{8pt}<{\centering}p{8pt}<{\centering}p{8pt}<{\centering}p{8pt}<{\centering}p{8pt}<{\centering}p{8pt}<{\centering}p{8pt}<{\centering}p{8pt}<{\centering}|p{8pt}<{\centering}}
\toprule
Model                     &\rotatebox{90}{mIoU}  &\rotatebox{90}{vase} &\rotatebox{90}{clock} &\rotatebox{90}{shower}  &\rotatebox{90}{light} &\rotatebox{90}{streetlight}  &\rotatebox{90}{sconce} &\rotatebox{90}{ashcan}  &\rotatebox{90}{mIoU$_{s}$}   \\ \midrule
DLV3+(R101)\cite{chen2018encoder}   &44.6 &35.4 &22.6 &0.0   &49.6  &22.9 &35.2 &37.3 &\textbf{29.0} \\
SWin-T\cite{liu2021swin}       &44.4 &27.0 &20.1 &1.0 &52.2  &19.8 &36.2 &36.0 & 27.5 \\ \midrule
ConvNeXt-S\cite{liu2022convnet}   &48.6 &39.4 &37.4 &1.1 &54.6  &25.8 &46.3 &40.5 &\textbf{35.0} \\
SWin-B\cite{liu2021swin}   &48.0 &37.3 &40.0 &0.0 &56.5  &24.0 &44.5 &37.5 &34.3 \\
MAE\cite{he2021masked}   & 48.1 &29.3 &33.1 &1.3  &53.1 &25.7 &46.7 &39.3 &32.6\\
SETR-PUP-L\cite{zheng2021rethinking}& 48.2 &31.5 &31.9 &3.7  &41.9 &16.7 &36.9 &34.1 &28.1\\ \midrule
ConvNeXt-B\cite{liu2022convnet}   &52.1 &45.4 &44.2 &3.0 &59.1  &30.4 &53.3 &40.1 &\textbf{39.4} \\
Segmenter-L\cite{strudel2021segmenter} &51.7    &39.0 &26.2 &5.9 &43.9 &23.2 &42.3 &46.7 &32.5 \\ \midrule
ConvNeXt-XL\cite{liu2022convnet}   &53.6 &48.1 &44.5 &5.4 &61.8  &34.4 &56.3 &42.8 &\textbf{41.9} \\
BEiT-B\cite{bao2021beit} &53.1  &46.2 &43.1 &6.9 &57.9 &33.0 &54.1 &48.3 &41.4 \\\bottomrule
\end{tabular}
\end{table}

\begin{figure}
\centering
\begin{minipage}{0.65in}
\includegraphics[width=0.65in]{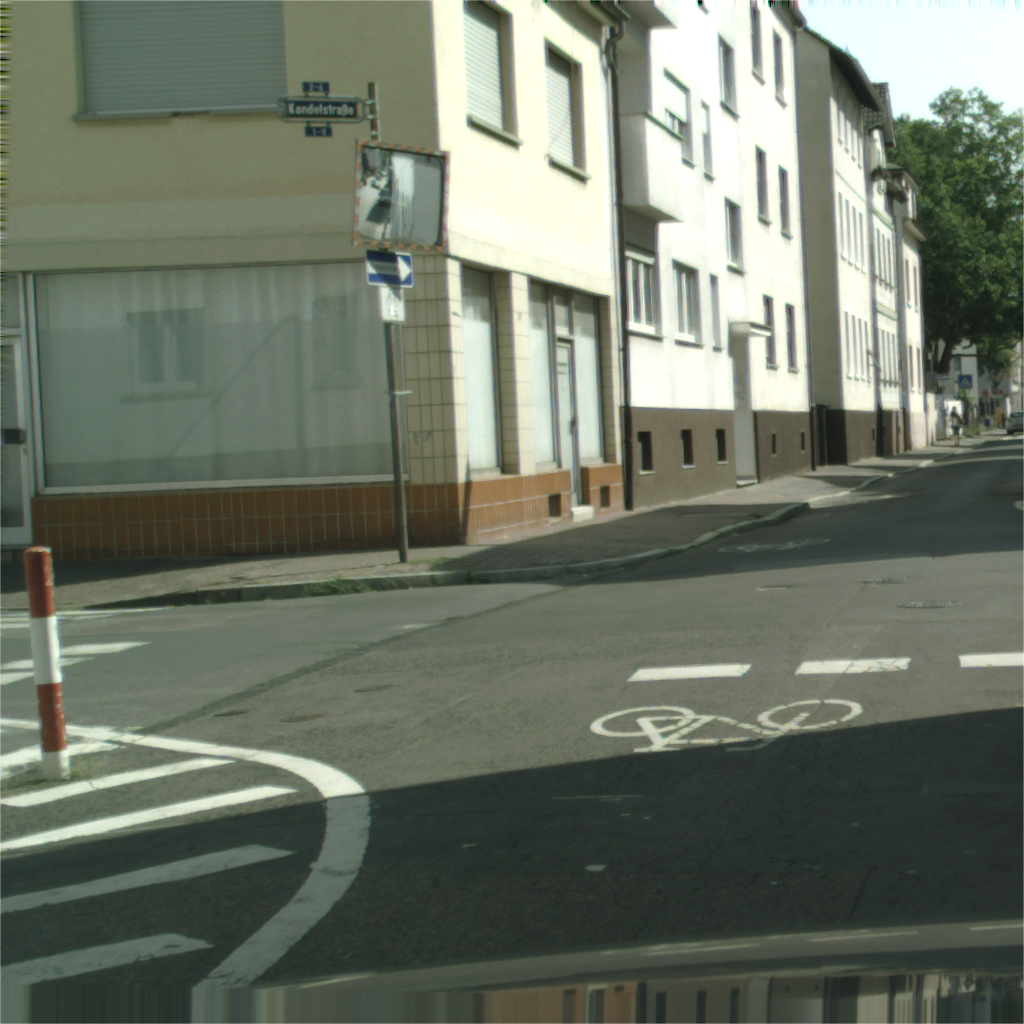}
\subcaption*{(a)}
\end{minipage}
\begin{minipage}{0.65in}
\includegraphics[width=0.65in]{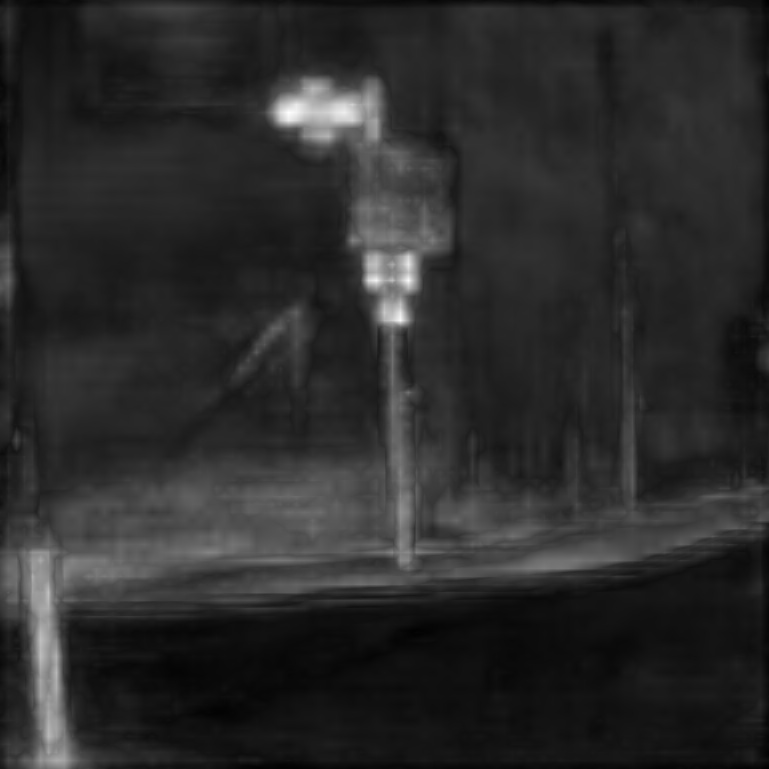}
\subcaption*{(b)}
\end{minipage}
\begin{minipage}{0.65in}
\includegraphics[width=0.65in]{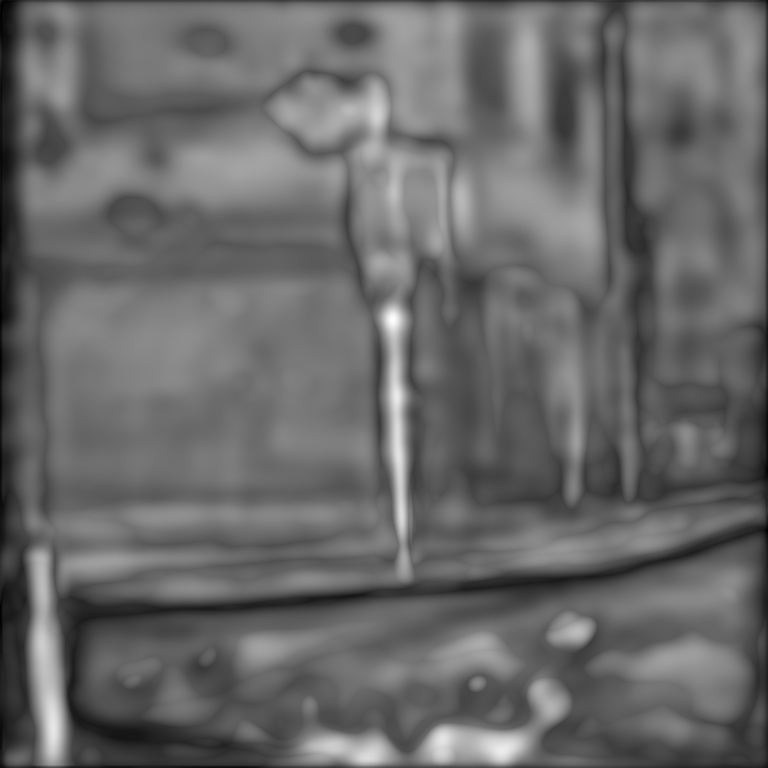}
\subcaption*{(c)}
\end{minipage}
\begin{minipage}{0.65in}
\includegraphics[width=0.65in]{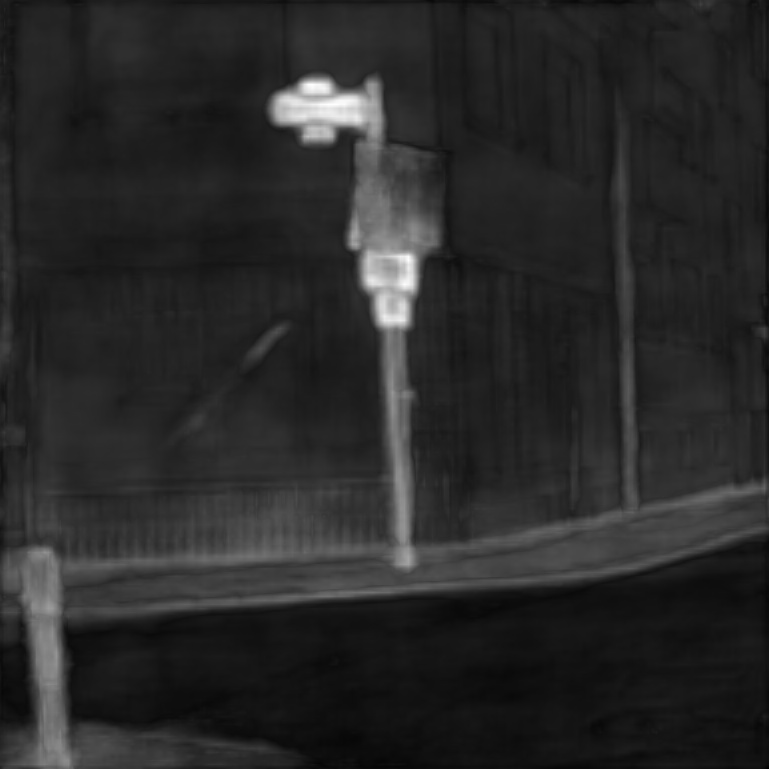}
\subcaption*{(d)}
\end{minipage}
\vspace{0.3em}
\begin{minipage}{0.65in}
\includegraphics[width=0.65in]{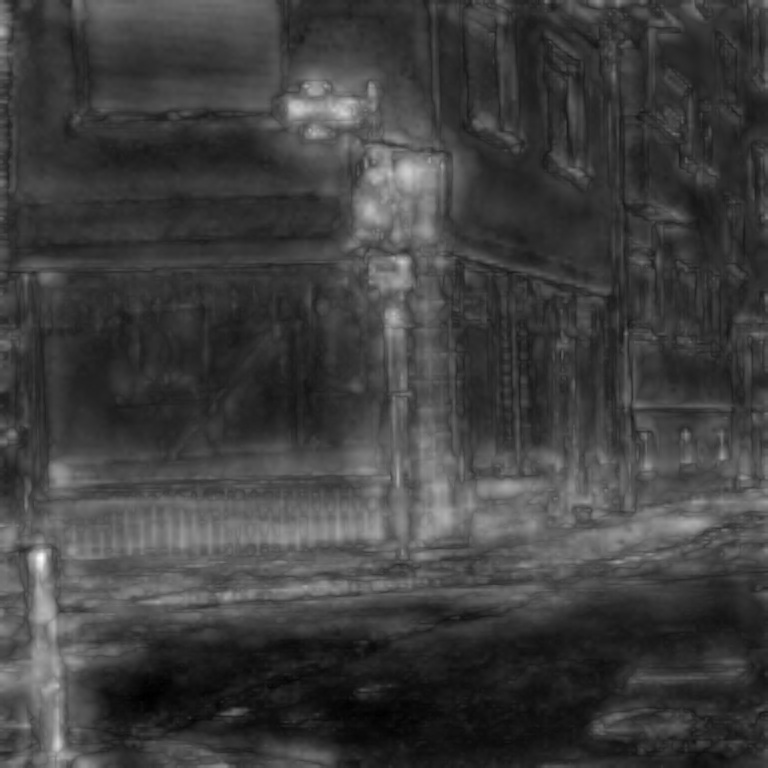}
\subcaption*{(e)}
\end{minipage}
\caption{Visualized feature maps before the last layer on Cityscapes. (a) Image, (b) PSPNet(R50), (c) SETR-PUP-L, (d) DLV3+(R101), and (e) SegFormer-B2. Zoom in for better details.}
\label{fig3}
\end{figure}

Given the smooth representations learned by ViT backbones, we further investigate the smoothness of existing semantic segmentation methods with ViTs (including plain ViTs and pyramid ViTs) by comparing them with CNN based methods. Specially, we investigate the segmentation accuracy of small or thin objects on two most commonly used datasets, $i.e.$, Cityscapes and ADE20K. It mediately reflects the degree of smoothness of the learned representations. To eliminate the influence of model capability and critical training settings, we follow the two principles, 1) the overall accuracy of models should be similar and 2) the training crop sizes should be consistent. For Cityscapes, we select the small or thin objects following previous works and for ADE20K, we select the seven smallest objects according to their average sizes of the entire dataset. As shown in Table \ref{tab:17} and \ref{tab:18}, compared to state-of-the-art CNNs based methods, ViTs based methods perform bad at small objects segmentation and plain ViTs based methods are worse. It indicates that ViT based methods tend to learn smoother representations. A visualized example is provided in Fig \ref{fig3}. In addition, an interesting discovery is that plain ViTs with masked pre-training\cite{he2021masked,bao2021beit} achieve better segmentation results of small objects compared to \cite{zheng2021rethinking,strudel2021segmenter}.

\subsection{Representation Separation Framework}

Since semantic segmentation methods with plain ViTs often suffer from over-smooth segmentation maps, we introduce a novel framework which incorporates local-patch representation separation and global-region representation separation, respectively counteracting the over-smoothness caused by self-attention and cross-attention.

\begin{figure*}
\centerline{\includegraphics[width=\textwidth]{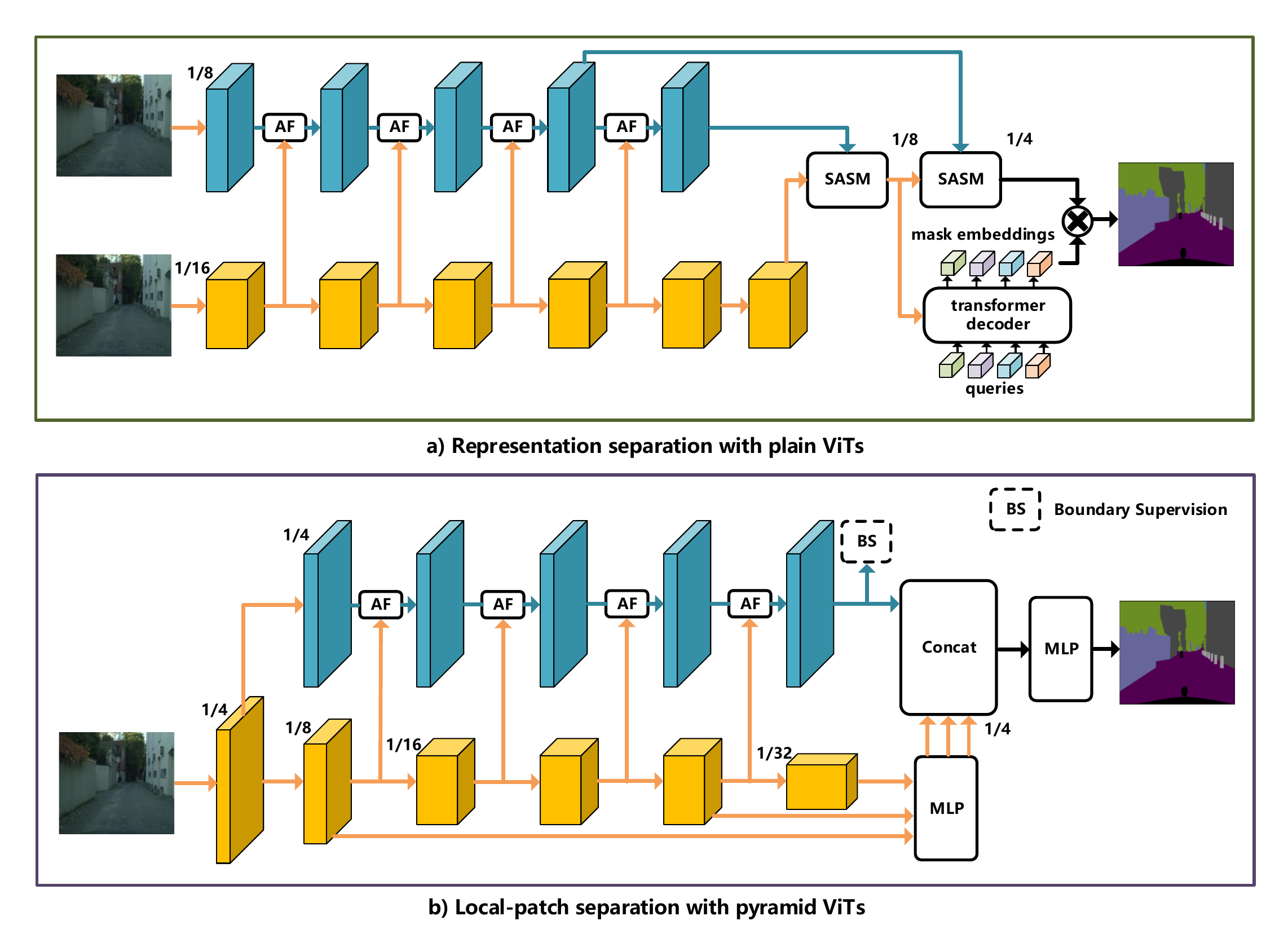}}
\caption{The overview of our methods with plain ViTs and pyramid ViTs. The blue blocks denote LSBs and the yellow blocks denote pre-trained transformer layers. The dotted box denotes the module is dropped during inference.}
\label{fig4}
\end{figure*}

\subsubsection{Local-patch representation separation}

We define patch-level representational discrepancy as the discrepancy among patch embeddings of a local region. It stems from the fact that pixels of a local region near the semantic boundary need to be classified into different categories. The patch-level representation separation should be achieved through operating the local patch embeddings and the source of local-patch discrepancy is the rich textures of shallow feature maps which usually contain a certain amount of noise. To obtain richer local information from smooth feature maps of transformers and suppress the noise, we propose the decoupled two-pathway network. As shown in \cref{fig4}, we construct an over-lapping patch embedding layer at first to generate a longer sequence of patch embeddings which is put into the first LSB. The over-lapping patch embedding layer has the same initialization as the vanilla patch embedding layer but outputs 4$\times$ longer sequence than it. Then, we upsample the outputs of transformer layers by bilinear interpolation and merge the outputs of LSBs into them through attention-guided fusion in a cascaded way. In addition to the top-down enhancement path, we also take into account how to effectively separate the deep representations in a bottom-up way. Different from pyramid CNNs or ViTs, plain ViTs keep the feature resolution constant to the last block, which means the last feature maps represent rich semantics and spatial cues at the same time. We therefore propose the spatially adaptive separation module which utilizes the shallow local discrepancy to guide the local-patch separation of the last feature maps. The SASM generates more separate patch embeddings in a local region with spatially adaptive weights formulated by the output of LSBs. In our framework, we apply two SASMs sequentially to obtain the feature maps with 1/4 image resolution.

In addition to plain ViTs, we also construct the decoupled two-pathway network with pyramid ViTs to demonstrate the extensibility as shown in the down of \cref{fig4}. We choose the output of the first stage as the input of the local separation path. Then, we upsample the output of the second stage and the evenly spaced feature maps of the third stage for the subsequent fusion. At last, four feature maps of pyramid resolutions are concatenated together and the final segmentation predictions are generated by two MLP layers. The binary boundary supervision is employed to better guide the learning of LSBs.

\subsubsection{Global-region representation separation}
\label{Region-level representation separation}

For the learnable queries paradigm, the final segmentation masks are acquired via a matrix multiplication of patch embeddings and mask embeddings which are usually obtained by making learnable queries interact with patch embeddings globally. Therefore, the mask embeddings can be regarded as globally region-level representations because each of them contains the global information of one semantic region. The discrepancy of region representations is also important because the final mask predictions can be seen as the similarity match between patch embeddings and mask embeddings. We suppose that this discrepancy is easily lost on account of the smooth feature maps and global match of cross-attention. To solve this problem, we propose the discriminative cross-attention to replace the vanilla cross-attention in the transformer decoder. It utilizes extra optimization objectives to achieve the targeted match during the phase of feature aggregation. As shown in \cref{fig4}, we enter the output of the first SASM into a transformer decoder to obtain discriminative mask embeddings.

\begin{table*}[]
\centering
\caption{The detailed configurations of architecture hyper-parameters. `Layers' denote the depth configuration of the backbone. The sampling indices start from 0. In general, we evenly sample the outputs of the transformer blocks for simplicity}
\label{tab:11}
\begin{tabular}{p{50pt}p{40pt}<{\centering}p{100pt}<{\centering}p{60pt}<{\centering}p{60pt}<{\centering}p{60pt}<{\centering}p{60pt}<{\centering}}
\toprule
Backbone                     &Layers               & Sampling indices    & Input dimension & Expand ratio  & Group number  & Group dim   \\\midrule
CSWin-T\cite{dong2021cswin}    & [1,2,21,1]      & [2,5,8,11,14,17,20,23]        & 64        & 4   &-&-          \\
CSWin-B\cite{dong2021cswin}    & [2,4,32,2]      & [5,9,13,17,21,25,29,33,37]    & 128       & 2    &-&-         \\
ViT-B\cite{steiner2021train}   & 12              & [1,3,5,7]      & 256          & 2  & 12          & 64  \\
ViT-L\cite{steiner2021train}   & 24              & [3,7,11,15]    & 384          & 2  &  16         & 64            \\
BeiT-B\cite{bao2021beit}       & 12              & [1,3,5,7]      & 256          & 2  & 12          & 64              \\
BeiT-L\cite{bao2021beit}       & 24              & [3,7,11,15]    & 384          & 2  &  16         & 64              \\  \bottomrule
\end{tabular}
\end{table*}

\subsubsection{Architectural hyper-parameters}

We provide the detailed configurations of architectural hyper-parameters in Table \ref{tab:11}, including the input dimension and expand ratio of LSBs, the group number and group dim of SASMs, and the sampling indices of intermediate layers. Note that we do not fine-tune most of the hyper-parameters and just use the common configurations.

\subsection{Decoupled Two-pathway Network}

\subsubsection{Local separation block}

We define the process of a local separation block following the inverse residual block (ignoring the normalization and activation layers) :
\begin{equation}
LMB(x) = Conv_{1\times1}(OP_{local}(Conv_{1\times1}(x))) + x,
\label{eq4}
\end{equation}
where $OP_{local}$ represents the depth-wise local operators in the patch embedding space.
As can be seen from the above formulas, $OP_{local}$ plays a significant role in enhancing the local-patch discrepancy.
Although vanilla depth-wise convolutions are widely used for local modeling, they still probably lead to
the local similarity. Inspired by handcrafted edge extraction filters, we propose the learnable high-pass filters to explicitly enhance the local discrepancy by constraining the sum of kernel weights to zero. Given a convolution kernel with arbitrary weights, we first apply the $softmax$ function in the kernel dimension to make these weights add up to 1. And then a small fixed value is subtracted to achieve the zero-sum weights:
\begin{equation}
kernel_{high-pass} = Softmax(kernel) - 1.0/kernel_{size},
\label{eq11}
\end{equation}
where $kernel_{size}$ denotes the size of the convolution kernel. In this paper, we use the 5$\times$5 convolution of which $kernel_{size}$ is 25.
A scalable BatchNorm or LayerNorm layer should be added behind the high-pass filters to offset the limitation of kernel weight values caused by $softmax$ function.

\subsubsection{Attention-guided feature fusion}
The enhancement of local discrepancy probably produces undesirable noise due to the limitation of receptive fields.
To suppress the noise, we propose the attention-guided feature fusion mechanism which selectively fuses the local discrepancy into the upsampled smooth feature maps:
\begin{equation}
Xl_{i+1} = LSB(Xl_{i}), i = 0,
\label{eq5}
\end{equation}
\begin{equation}
Xl_{i+1} = LSB(Xl_{i}^{'}), i > 0,
\label{eq6}
\end{equation}
\begin{equation}
Xl_{i}^{'} = Attn([GP(Xl_{i}), GP(Xg_{i})])*Xl_{i} + Up(Xg_{i}),
\label{eq7}
\end{equation}
\begin{equation}
Attn(\cdot) = Sigmoid(BN(Conv_{1\times1}(ReLU(BN(\cdot)))),
\label{eq8}
\end{equation}
where $Xl_{i}$ denotes the output feature map of the $i$-th LMB, $Xg_{i}$ denotes the $i$-th sampled smooth feature map from the transformer backbone, GP denotes the global average pooling, and [] denotes the feature concatenation.

\subsubsection{Relationship to existing two-pathway methods}

There have been some two-pathway methods in the CNN era\cite{yu2018bisenet,yu2021bisenet,hong2021deep,pang2019towards,huang2021alignseg,takikawa2019gated}. Some of them is appealing in real-time semantic segmentation\cite{yu2018bisenet,yu2021bisenet,hong2021deep}. However, the performances of these methods usually fall behind methods based on dilation backbones or HRNets. Different from \cite{pang2019towards} and \cite{huang2021alignseg}, our method centers on enhancing the local discrepancy and extracting the textural information. Gated-SCNN\cite{takikawa2019gated} is the most related work that utilizes the gated convolutions to enable another branch to focus only on boundary-related information. However, it does not fuse semantically rich feature maps in the middle of the shape stream and attempts to locate the semantic boundary by global operations. On the contrary, we continually merge feature maps from self-attention backbones by attention-guided mechanism and enhance the boundary-related information by local high-pass filters. Note that our method makes the local features preferably supplement the semantically rich features and therefore enhances the local-patch discrepancy of smooth representations, which is different from Gated-SCNN that attempts to make the local path better focus on itself.

\subsection{Spatially Adaptive Separation Module}

\begin{figure}
\centerline{\includegraphics[width=\columnwidth]{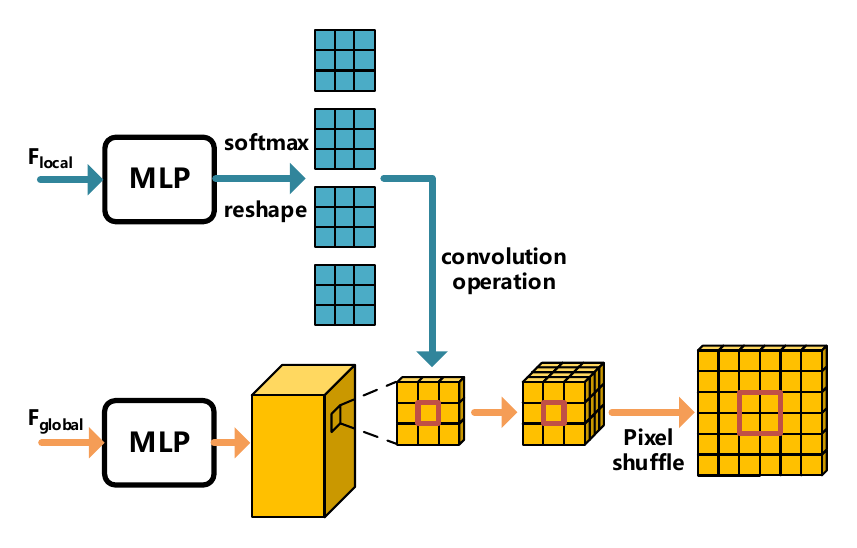}}
\caption{The details of spatially adaptive separation module.}
\label{fig5}
\end{figure}

The linear projection of the patch embedding layer actually encodes the structural information of image patches into the channel dimension of patch embeddings. Our local separation block further outputs the enhanced features of transformers which incorporate more local information and discrepancy. Since the local information is encoded in the channel dimension, we convert the patch embeddings of LSBs to spatial filters which are operated on the corresponding local patch embeddings of deep global representations. As shown in \cref{fig5},
we achieve this process by one MLP layer for matching the dimension and the reshape operation. Specially, we apply the $Softmax$ function to each spatial filter to constrain the weight values:
\begin{equation}
SAF^{(B,4,3,3,H,W)} = Softmax(Reshape(MLP(F_{local}))),
\label{eq16}
\end{equation}
where $SAF$ is the spatially adaptive filter and $F_{local}$ is the output of LSB.
Like the common interpolation methods generate new pixels by weighted summing surrounding pixels, we use four spatial filters applied on one 3$\times$3 local region to generate four new patch embeddings for the original patch embedding, the red box in \cref{fig5} and double the feature resolution through pixel shuffle:
\begin{equation}
F_{global 2\times}^{(B,C,2H,2W)} = Pixelshuffle(F_{global}\otimes SAF),
\label{eq17}
\end{equation}
where $\otimes$ denotes the convolution operation and $F_{global}$ $\in$ $\mathbb{R}^{B,C,H,W}$ is the guided global representation.
The new patch embeddings are more easily to be separated due to the guidance of local information and higher feature resolution.
To increase the representational capacity, we further group the guided features in channel dimension and learn the spatial filters respectively for each grouped feature. As shown in the up of \cref{fig4}, the first SASM utilizes the output of the last LSB to guide the upsampling of the final output of the transformer. Since the SASM requires that the guidance feature has the same resolution as the guided feature, we use a 2$\times$2 convolution with a stride of 2 to downsampling the output of the last LSB before putting it into the first SASM. Then, the output of the first SASM is passed into the second SASM and the output of the penultimate LSB is utilized as the guidance features. The output of the second SASM with resolution of 1/4 is used as the final patch embeddings to calculate the similarity with mask embeddings.

Zhou \emph{et al.} \cite{zhou2021decoupled} propose dynamic joint upsampling that utilizes high-resolution features as guidance to generate the spatial filters for upsampling the low-resolution features, which is related to our method. However, not like they use the high-resolution RGB images to upsample the low-resolution depth maps, we harvest the discriminative features with rich local information to upsample the smooth but semantically rich features of the same source. In addition, we only utilize the guidance features and apply $Softmax$ function to generate the positive filters but they obtain the spatial filters with a more complex decoupled method. Our method also differs from the guided upsample method in \cite{mazzini2018guided,li2020semantic,huang2021alignseg} which aligns low-resolution features and high-resolution features by learning where to be interpolated in the whole feature maps because we would like to separate the local representations.


\subsection{Discriminative Cross-attention}

\begin{figure}
\centerline{\includegraphics[width=\columnwidth]{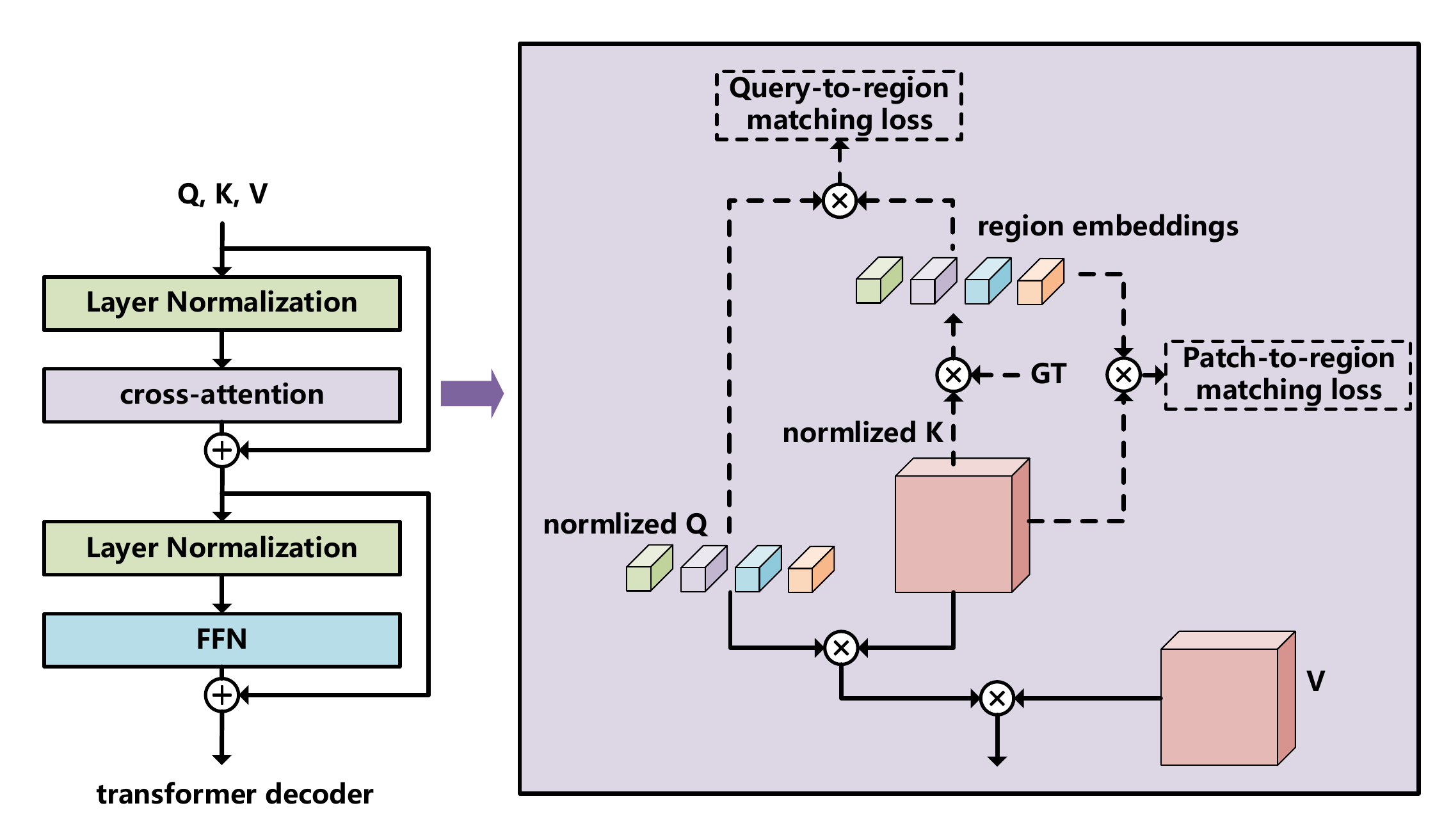}}
\caption{The details of discriminative cross-attention.}
\label{fig6}
\end{figure}
The vanilla cross-attention in transformer decoders can not ensure each mask embedding capture the region or instance oriented representations distinctly. We therefore propose the discriminative cross-attention which only utilizes the auxiliary supervisions to generate more discriminative mask embeddings without increasing the inference cost, as shown in Fig. \ref{fig6}. We first refer to the L2-normalized cross-attention:
\begin{equation}
MSA(Q, K, V) = Softmax(\frac{Q_{norm}K_{norm}^T}{s})V,
\label{eq12}
\end{equation}
where $Q_{norm}$ $\in$ $\mathbb{R}^{B,N_{class},C}$ and $K_{norm}$ $\in$ $\mathbb{R}^{B,N,C}$ denote the L2-normalized $Q$ and $K$ across the channel dimension, and $s$ denotes the learnable scale factor to restore the representational capacity. Inside the cross-attention, each query needs to calculate the similarity with all $H\times W$ patch embeddings, which increases the matching difficulty and probably results in the undiscriminating aggregation of patch embeddings, especially for the smooth feature maps. Since the purpose of each query is to search for all the embeddings of some specific regions from the whole feature maps, we directly enhance the matching capacity of each query to the corresponding semantic region in the region-level space. Specially, we construct a query-to-region matching task in which each query only matches the corresponding region embedding and keeps away from the other region embeddings. The region embedding ought to represent the whole information of the semantic region, which is defined as the average of all patch embeddings belonging to the same category. The query-to-region matching task reduces the number of keys from $H\times W$ to the number of categories ($N_{class}$), which contributes to more discriminative matching. We first utilize the downsampled one-hot ground truth to generate the region embedding (RE) upon the $K_{norm}$. Then, we calculate the similarity between the $Q_{norm}$ and the L2-normalized region embeddings like the formation of $attention$ matrix:
\begin{equation}
RE = Flatten(GT)K_{norm},
\label{eq13}
\end{equation}
\begin{equation}
Sim_{query-to-region} = Softmax(\frac{Q_{norm}RE_{norm}^T}{s}),
\label{eq14}
\end{equation}
where $Flatten()$ denotes flattening the dimensions of $HW$, $GT$ $\in$ $\mathbb{R}^{B,N_{class},H,W}$ denotes the one-hot ground truth, and $RE_{norm}$ $\in$ $\mathbb{R}^{B,N_{class},N_{class}}$ denotes the L2-normalized region embeddings. Since we expect each query closes to the corresponding region embedding and keeps away from the other region embeddings, the ground truth of $Sim_{query-to-region}$ ought to be a diagonal matrix of $N_{class}$$\times$$N_{class}$. However, there are always some zero vectors among the region embeddings due to the inexistence of some classes in one image. The corresponding queries can not get the matched region embeddings so we ignore the match of these queries when calculating the loss.
Moreover, to ensure the region embeddings are capable of representing the whole region, inspired by inter-class center-to-pixel loss in \cite{huang2022car}, we propose the patch-to-region matching loss to make $K_{norm}$ more compact and discriminative. The idea is to let the patch embeddings of the same category close to their own region embedding and keep away from the other region embeddings. We directly optimize the $softmax$ similarity between normalized patch embeddings and normalized region embeddings:
\begin{equation}
Sim_{patch-to-region} = softmax(\frac{K_{norm}RE_{norm}^T}{s}).
\label{eq15}
\end{equation}
The ground truth of $Sim_{patch-to-region}$ is the downsampling one-hot segmentation label. The query-to-region matching loss and patch-to-region matching loss are calculated with the widely used cross-entropy function. We initialize the learnable scale factor $1/s$ in Equation \eqref{eq14} and Equation \eqref{eq15} to 10.0 for accelerating convergence.

\subsection{Loss Function}

We upsample the final segmentation maps to the image size by bilinear interpolation and use the common cross-entropy function to measure the differences between the
segmentation results and the ground truths. For simplicity, we do not adopt the widely used deep supervision\cite{zhao2017pyramid}, Online Hard Example Mining, or class balance loss.
The overall loss $L$ consists of three parts:
\begin{equation}
L = l_{seg} + l_{query-to-region} + l_{patch-to-region}.
\label{eq18}
\end{equation}
For pyramid ViTs, we adopt the boundary loss proposed in \cite{fan2021rethinking} and set the weight of this auxiliary loss to 0.4 following previous works\cite{yuan2020object,9633155}. The total loss is:
\begin{equation}
L = l_{seg} + 0.4l_{boundary}.
\label{eq19}
\end{equation}

\section{Experiments}

\subsection{Datasets}

\textbf{Cityscapes}
\cite{cordts2016cityscapes} contains 5000 finely annotated images, among which contain 2975 training images, 500 validation images, and 1525 test images. The image resolution is 2048$\times$1024, which is challenging in computation and memory usage. In addition, it provides 20000 coarsely labeled images which we do not use.

\noindent\textbf{ADE20K}\cite{zhou2017scene} is a challenging scene parsing dataset composed of 150 semantic classes. It provides 20210, 2000, and 3352 images for training, validation, and testing.

\noindent\textbf{PASCAL Context}\cite{mottaghi2014role} provides 4998 images for training and 5105 images for validation. Following previous works, we evaluate the models with 59 semantic labels and 1 background label.

\noindent\textbf{COCO-Stuff-10K}\cite{caesar2018coco} is a subset of the COCO dataset for semantic segmentation, which consists of 9000 training images and 1000 testing images of 171 categories.

\noindent\textbf{COCO-Stuff-164K}\cite{caesar2018coco} includes 118k training images and 5k validation images of the COCO 2017. It is also labeled into 171 categories.

\subsection{Implementation Details}

We use the $\mathit{MMSegmentation}$ \cite{mmseg2020} toolbox for all the experiments. We select the advanced pre-trained transformers equipped with more powerful capacity of consistency modeling to better demonstrate the role of our representation separation methods. The encoders are CSWin\cite{dong2021cswin}, ViT-AugReg\cite{steiner2021train}, and BEiT\cite{bao2021beit} models pre-trained on ImageNet-1K\cite{deng2009imagenet} or ImageNet-22K. We adopt the default data augmentation including random scaling, random cropping, and random horizontal flipping. Since the proposed methods are incorporated with plain ViTs and pyramid ViTs, we follow their respective previous training settings in general for a fair comparison. For the models using BEiT backbones, we follow the training protocols of the original paper. In addition, we apply the 10$\times$ learning rate to the randomly initialized parameters except for the parameters of transformer decoders and the learnable queries. For the models using ViT-AugReg backbones, we employ the same training settings with our BEiT models but use the initial learning rate of 2e-5 and weight decay of 0.01 following StructToken\cite{lin2022structtoken}.  The batch size is set to 16 for all datasets. We set the crop size to 768$\times$768 for Cityscapes, 512$\times$512 for ADE20K (except 640$\times$640 for ViT-L), 480$\times$480 for PASCAL Context, 512$\times$512 for COCO-Stuff-10K, and 640$\times$640 for COCO-Stuff-164K. The corresponding training iterations are 80K, 80K, 30K, 30K, and 80K. For the models using CSWin backbones, we follow the protocols of SegFormer\cite{xie2021segformer} and HRViT\cite{gu2021multi} to train them on Cityscapes.

During the inference phase, we adopt the sliding window inference of which window size is identical to the training crop size for plain ViTs models. For pyramid ViTs models, we adopt the sliding window inference for Cityscapes following \cite{xie2021segformer}. For the multi-scale results, we perform the multi-scale test with scaling factors of (1.0, 1.25, 1.5, 1.75) on Cityscapes. For the other datasets, the default scaling factors of (0.5, 0.75, 1.0, 1.25, 1.5, 1.75) are used.

\subsection{Ablation Study}
The ablation experiments are conducted on various datasets (ADE20K, Pascal Context, and Cityscapes) using ViT-B, BEiT-B, and CSWin as backbones. If not expressly stated, we train the models with 512$\times$512 crop size, 80K iterations for ADE20K, 480$\times$480 crop size, 30K iterations for Pascal Context, and 768$\times$768 crop size, 80K iterations for Cityscapes. All the results including the baselines are obtained by our environment.

\subsubsection{Ablation study on representation separation}

Our baseline is the pre-trained ViT with a linear decoder, which achieves 49.3$\%$ mIoU on ADE20K with the single-scale test. There is an improvement of 0.7$\%$ mIoU (from 49.3$\%$ to 50.0$\%$) by adding a transformer decoder to generate dynamic mask embeddings. With the decoupled two-pathway network framework, we improve the performance from 50$\%$ to 51.4$\%$ while only increasing 4M parameters. The performance gain is 1$\%$ when adding two SASMs on the decoupled two-pathway network. By further employing the discriminative cross-attention, the accuracy reaches 53.1$\%$ mIoU without extra parameters or computation. For Pascal Context, the proposed components continually improve the baseline with BEiT-B. The discriminative cross-attention improves the performance from 64.0$\%$ to 65.0$\%$ without extra parameters or computation.

\begin{table}[]
\caption{Ablation study on representation separation}
\label{tab:7}
\begin{tabular}{p{130pt}p{25pt}<{\centering}p{25pt}<{\centering}p{25pt}<{\centering}}
\toprule
Model                                     & mIoU     &GFLOPs    &Params  \\ \midrule
\multicolumn{4}{c}{ADE20K} \\ \midrule
ViT-Linear-B                              & 49.3     &107.2     &86.6M          \\
+ transformer decoder                     & 50.0     &110.7     &95.5M     \\
+ decoupled two-pathway network           & 51.4     &129.5     &99.7M      \\
+ SASMs                                  & 52.4     &127.5     &99.8M        \\
+ discriminative cross-attention          & \textbf{53.1}   &127.5&99.8M              \\ \midrule
\multicolumn{4}{c}{Pascal Context} \\  \midrule
BEiT-Linear-B                             & 62.4     &92.1     &86.2M          \\
+ transformer decoder                     & 62.4     &94.6    & 95.1M     \\
+ decoupled two-pathway network           & 63.5     &110.8    & 99.4M          \\
+ SASMs                                  & 64.0     &109.0    & 99.5M         \\
+ discriminative cross-attention          & \textbf{65.0}     &109.0    & 99.5M         \\ \bottomrule
\end{tabular}
\end{table}

\subsubsection{Ablation study on SASM}

\begin{table}[]
\caption{Ablation study on SASM}
\label{tab:12}
\begin{tabular}{p{130pt}p{25pt}<{\centering}p{25pt}<{\centering}p{25pt}<{\centering}}
\toprule
Model                                   & mIoU     &GFLOPs    &Params  \\ \midrule
2 SASM                                 & 53.1     &127.5     &99.8M    \\
1 SASM                                 & 52.6     &127.0     &99.7M \\
Filter Norm\cite{zhou2021decoupled}     & 52.8     &127.5     &99.8M               \\
group number = 1                               & \textbf{53.2}     &126.9  &99.5M            \\\bottomrule
\end{tabular}
\end{table}

As shown in Table \ref{tab:12}, only using one SASM to generate the 1/8 resolution feature maps, the performance drops by 0.5$\%$. Replacing the $softmax$ with the Filter Norm proposed in \cite{zhou2021decoupled} leads to a performance drop of 0.3$\%$. It is possibly because the constrained weight values stabilize the training. The performance is a little bit better by reducing the group number to 1. However, we still use the default group number following multi-head attention since such a setting usually has better representational capacity. The performance may be further improved by fine-tuning these hyper-parameters.

\subsubsection{Ablation study on discriminative cross-attention}

\begin{table}[]
\caption{Ablation study on discriminative cross-attention}
\label{tab:13}
\begin{tabular}{p{180pt}p{48pt}<{\centering}}
\toprule
Method                                  & mIoU       \\ \midrule
discriminative cross-attention          & \textbf{53.1}              \\
query-to-region matching loss           & 52.6                    \\
patch-to-region matching loss           & 52.8               \\
w/o L2 normalization                    & 51.9                   \\\bottomrule
\end{tabular}
\end{table}

\begin{figure}
\centering
\begin{minipage}{0.8in}
\includegraphics[width=0.8in]{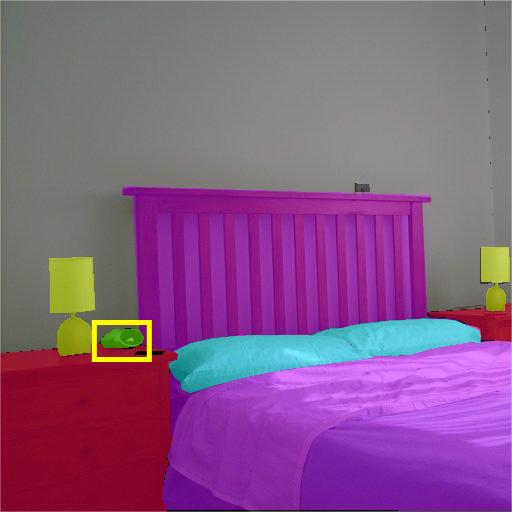}
\end{minipage}
\begin{minipage}{0.8in}
\includegraphics[width=0.8in]{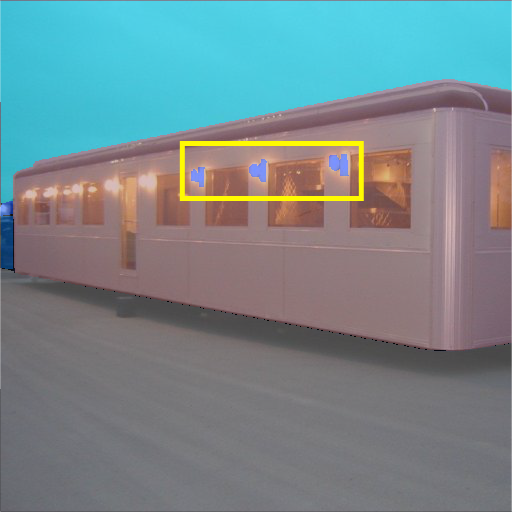}
\end{minipage}
\begin{minipage}{0.8in}
\includegraphics[width=0.8in]{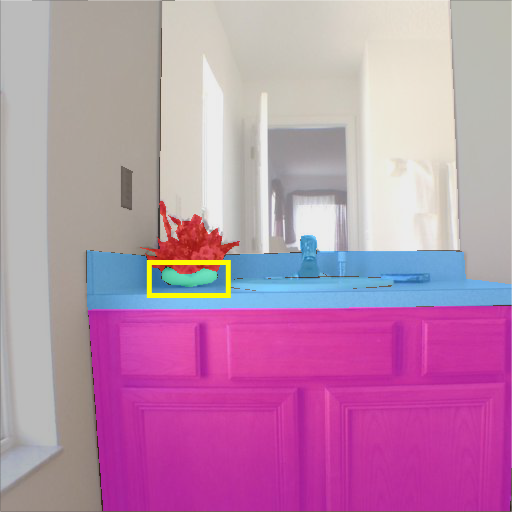}
\end{minipage}
\vspace{0.3em}
\begin{minipage}{0.8in}
\includegraphics[width=0.8in]{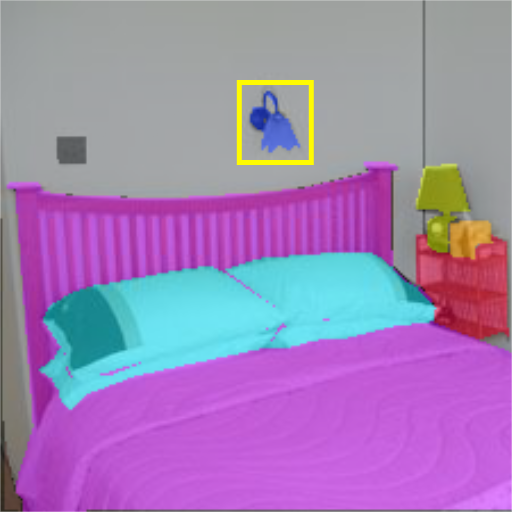}
\end{minipage}
\begin{minipage}{0.8in}
\includegraphics[width=0.8in]{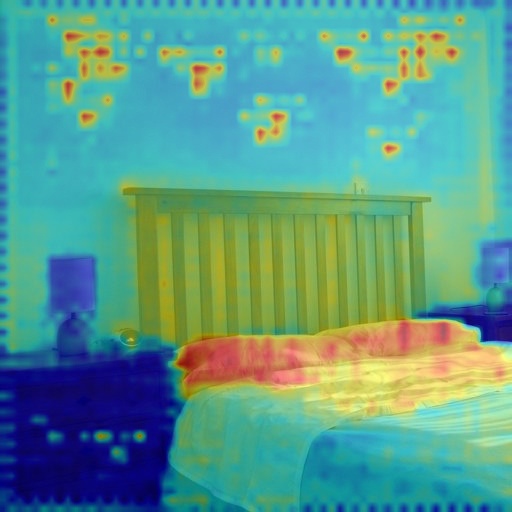}
\end{minipage}
\begin{minipage}{0.8in}
\includegraphics[width=0.8in]{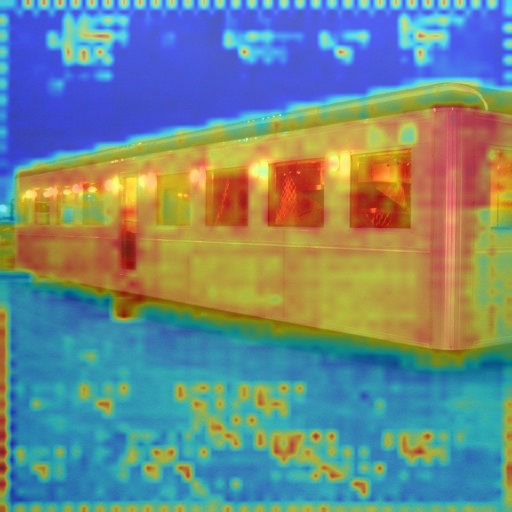}
\end{minipage}
\begin{minipage}{0.8in}
\includegraphics[width=0.8in]{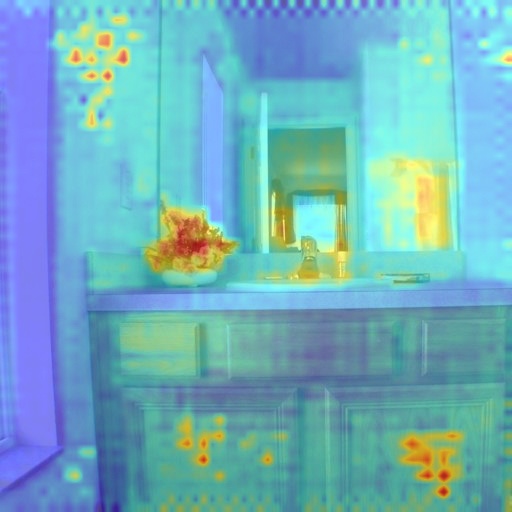}
\end{minipage}
\vspace{0.3em}
\begin{minipage}{0.8in}
\includegraphics[width=0.8in]{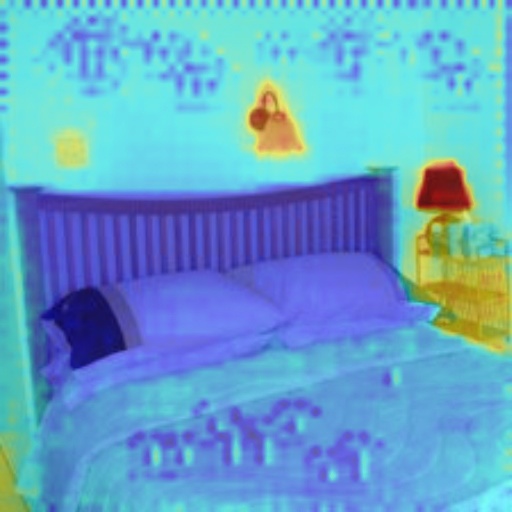}
\end{minipage}
\begin{minipage}{0.8in}
\includegraphics[width=0.8in]{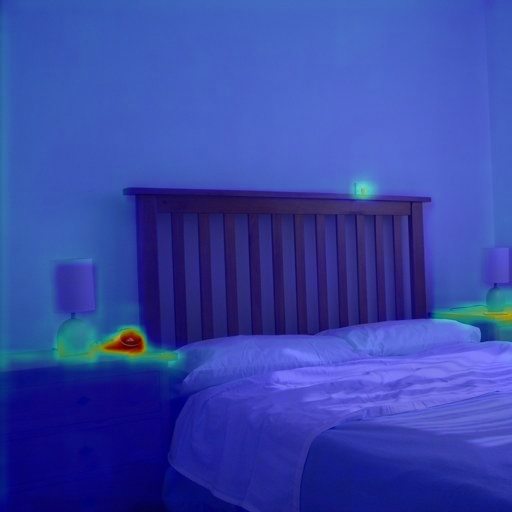}
\subcaption*{(a)}
\end{minipage}
\begin{minipage}{0.8in}
\includegraphics[width=0.8in]{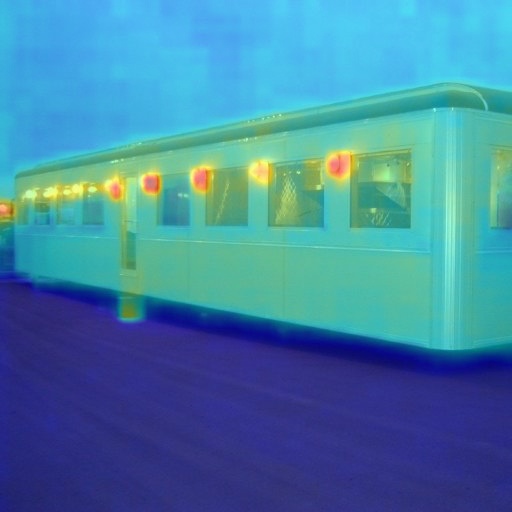}
\subcaption*{(b)}
\end{minipage}
\begin{minipage}{0.8in}
\includegraphics[width=0.8in]{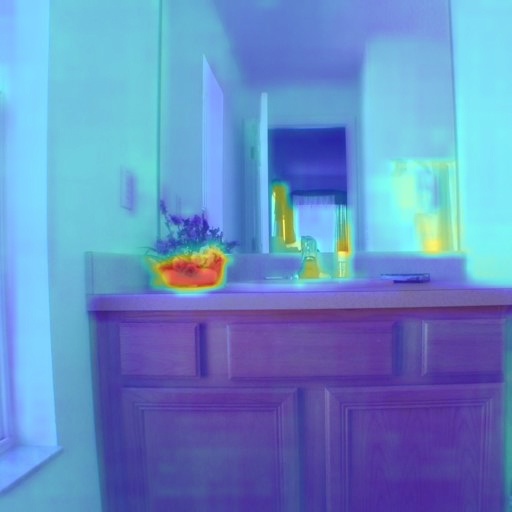}
\subcaption*{(c)}
\end{minipage}
\vspace{0.3em}
\begin{minipage}{0.8in}
\includegraphics[width=0.8in]{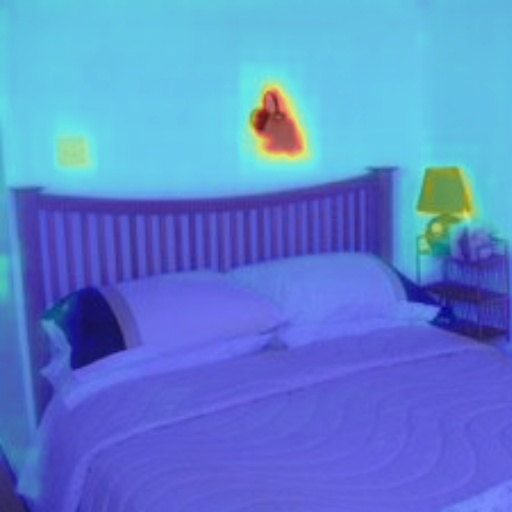}
\subcaption*{(d)}
\end{minipage}
\caption{Visualized attention maps of the transformer decoder on ADE20K val set. The figures from top to down are ground truths, attention maps of RSSeg-ViT-B without discriminative cross-attention, and attention maps of RSSeg-ViT-B. We select attention maps of typical small objects: (a) clock, (b) streetlight, (c) vase, and (d) sconce.}
\label{fig7}
\end{figure}

We further conduct ablation experiments on discriminative cross-attention, as shown in Table \ref{tab:13}. Only utilizing query-to-region matching loss or patch-to-region matching loss will decrease the performance. If not normalizing the embeddings before calculating the similarity, there is a dramatic drop in performance. It may be caused by the imbalance optimization of different region embeddings. As shown in Fig. \ref{fig7}, it is very difficult for vanilla cross-attention to locate the very small object in the whole image but our discriminative cross-attention is capable of pinpointing these small objects. In Fig. \ref{fig7} (a) and (b), the vanilla cross-attention pays more attention to the irrelevant background or more extensive object while the proposed discriminative cross-attention precisely finds the correct objects. In Fig. \ref{fig7} (c) and (d), the vanilla cross-attention focuses on the adjacent object or similar object mistakenly while our method centers on the correct targets in the existence of mighty interference. Note that we do not directly supervise attention maps with per-pixel segmentation labels.

\subsubsection{Ablation study on decoupled two-pathway network}

\begin{figure}
\centering
\begin{minipage}{1.1in}
\includegraphics[width=1.1in]{image.png}
\subcaption*{(a)}
\end{minipage}
\begin{minipage}{1.1in}
\includegraphics[width=1.1in]{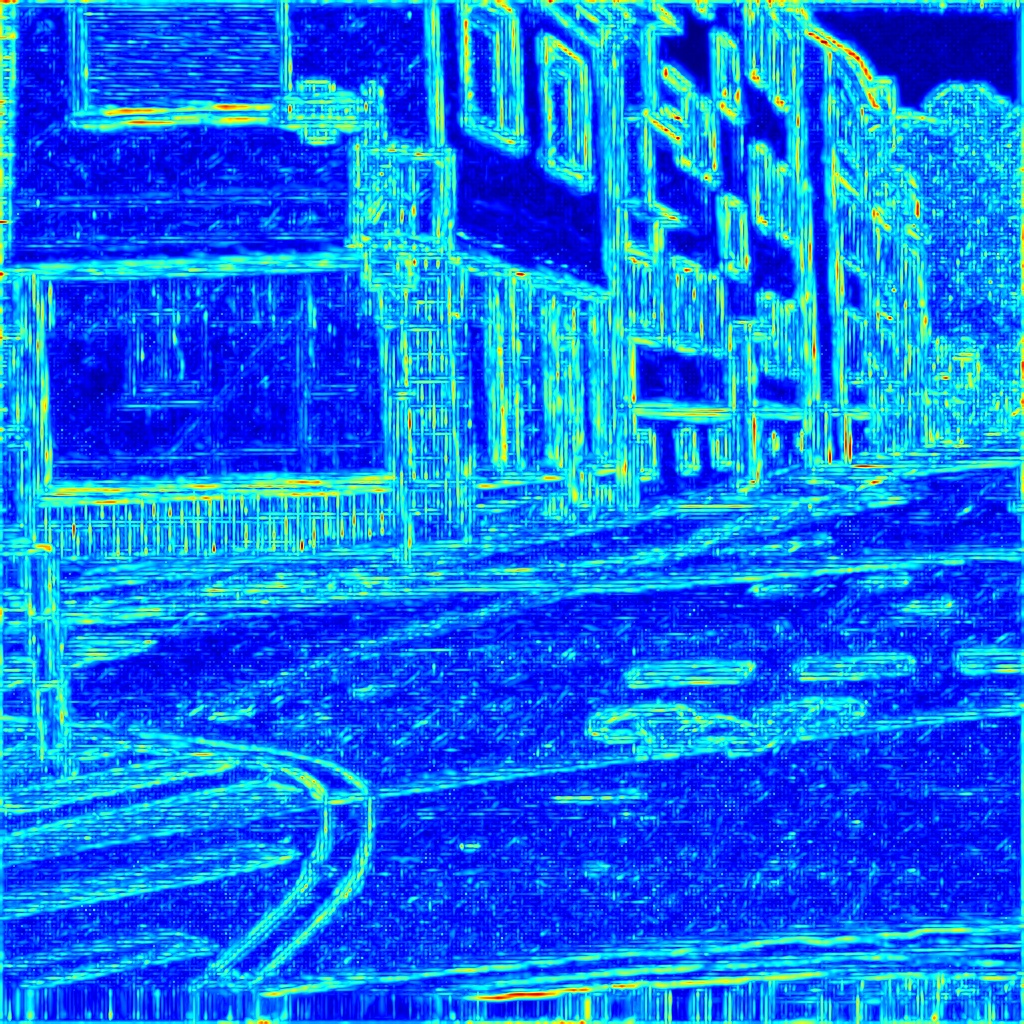}
\subcaption*{(b)}
\end{minipage}
\vspace{0.3em}
\begin{minipage}{1.1in}
\includegraphics[width=1.1in]{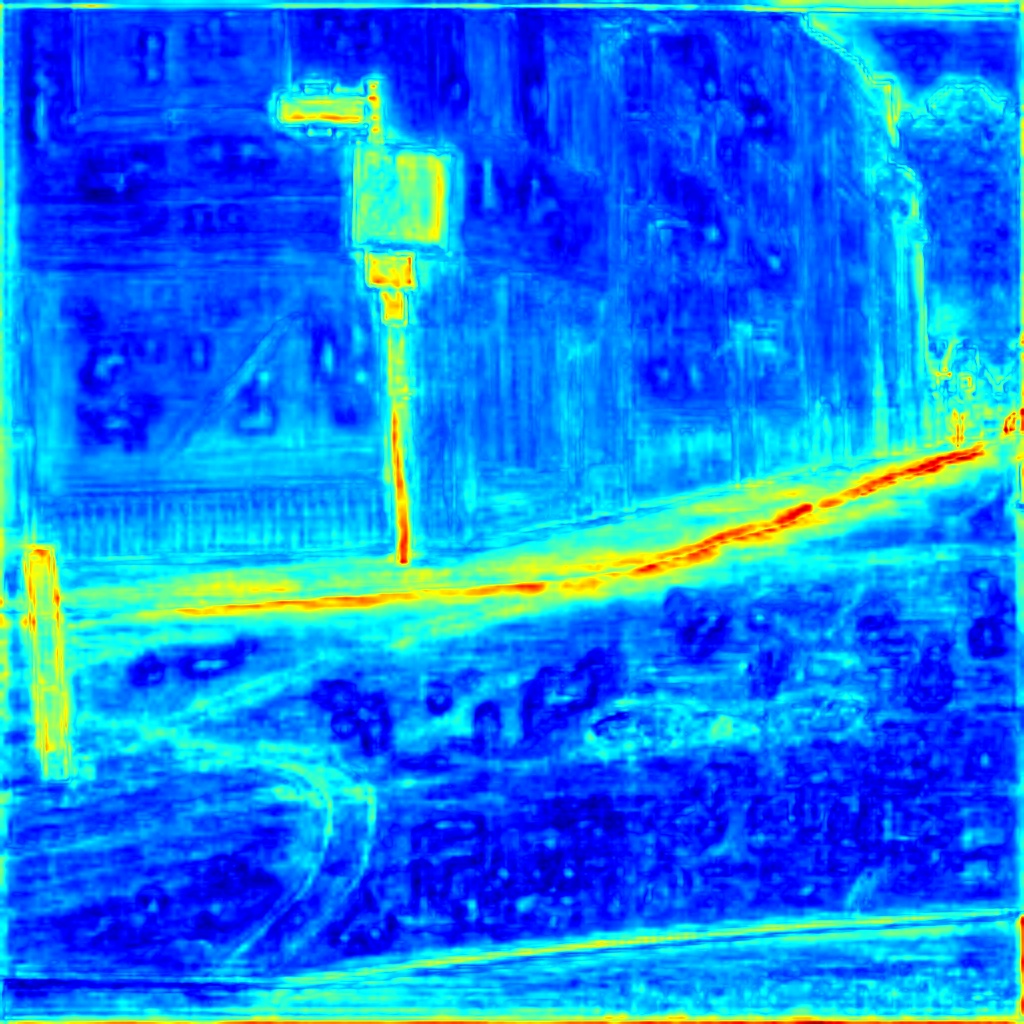}
\subcaption*{(c)}
\end{minipage}
\begin{minipage}{1.1in}
\includegraphics[width=1.1in]{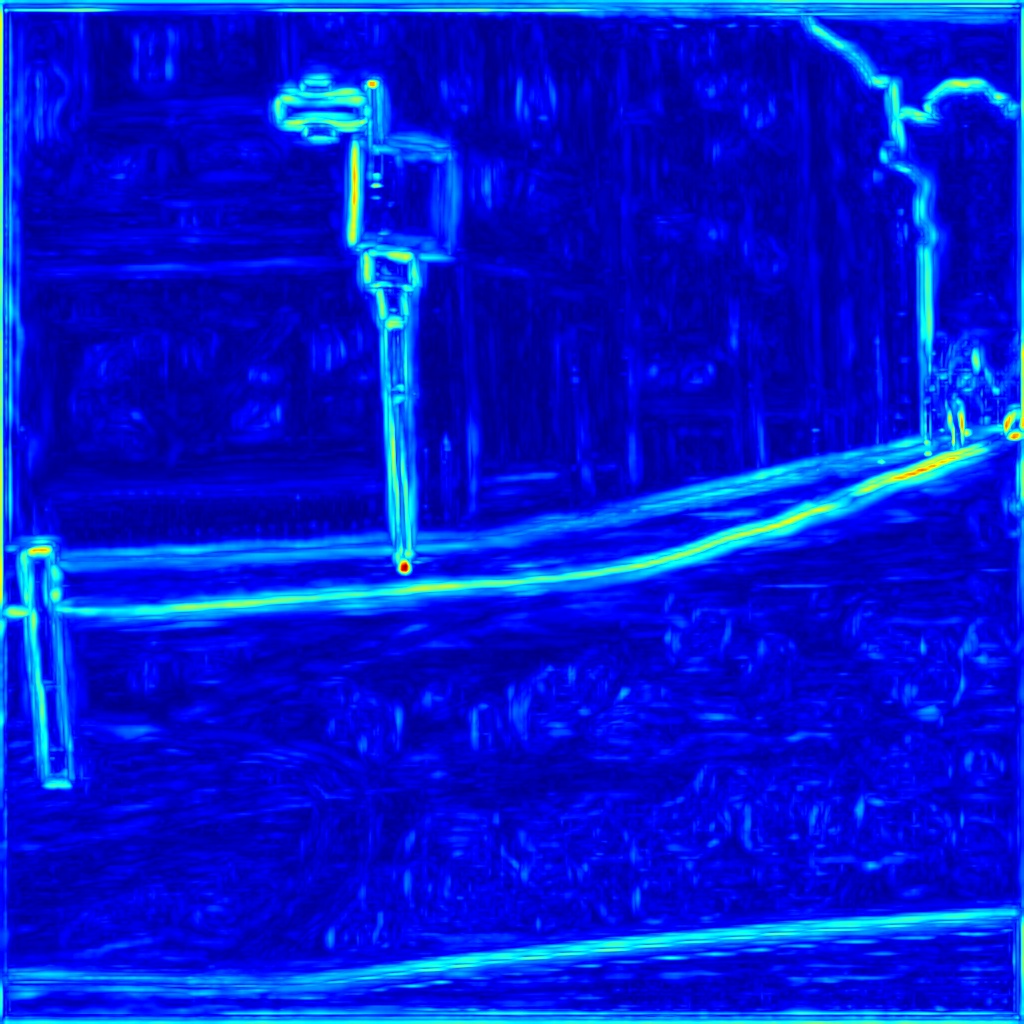}
\subcaption*{(d)}
\end{minipage}
\begin{minipage}{1.1in}
\includegraphics[width=1.1in]{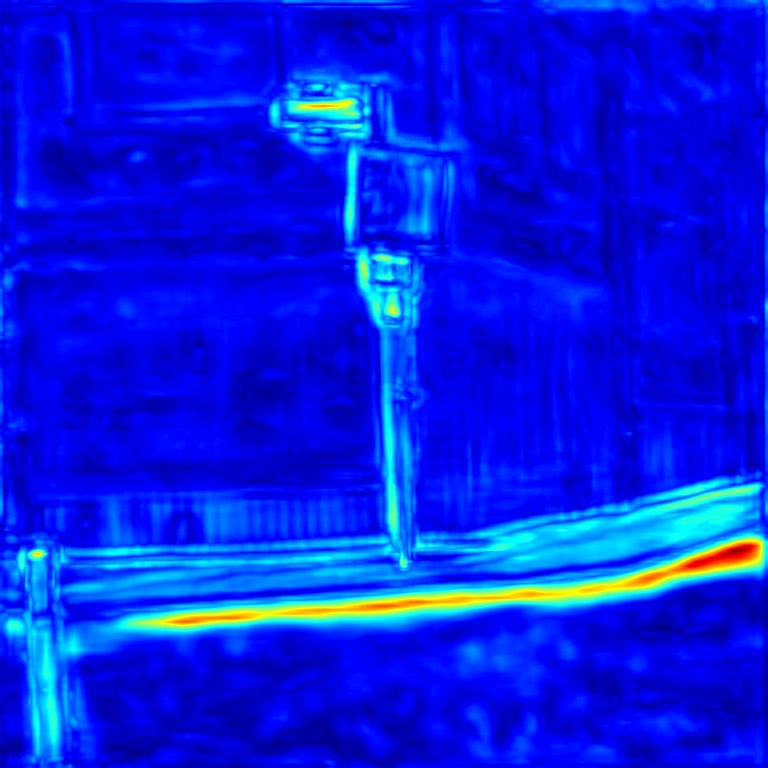}
\subcaption*{(e)}
\end{minipage}
\begin{minipage}{1.1in}
\includegraphics[width=1.1in]{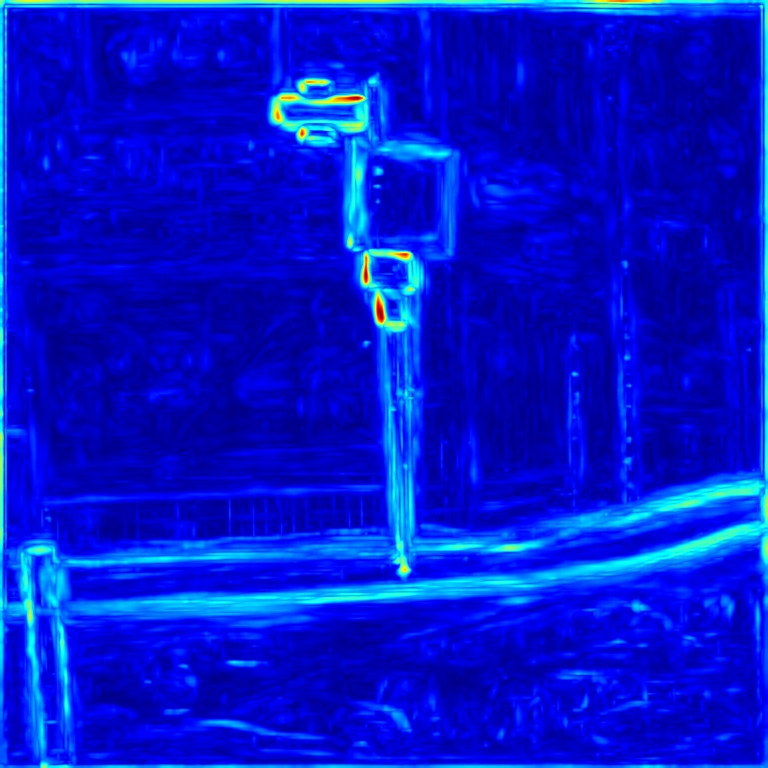}
\subcaption*{(f)}
\end{minipage}
\caption{Visualized feature maps on Cityscapes val set. (a) input image and segmentation map, (b) visualized result of the output of LHFs in the first LMB, (c) visualized result of the input of LHFs in the last LMB, (d) visualized result of the output of LHFs in the last LMB, (e) visualized result of the output of vanilla convolutions in the last LMB, (f) visualized result by replacing vanilla convolutions in (e) with LHFs.}
\label{fig8}
\end{figure}

The ablation study for the decoupled two-pathway network is conducted on Cityscapes using CSWin-T and CSWin-S as the backbone. We take the MLP decoders of SegFormer as the baseline. As shown in Table \ref{tab:8}, increasing the number of LSBs brings more performance gain, and adding 9 blocks improves the mIoU from 81.5$\%$ to 82.3$\%$. Enlarging the learning rate enables more sufficient training of LSBs, leading to 0.3$\%$ performance gain. The attention-guided feature fusion improves the performance slightly. At last, using the proposed LHFs together with edge supervision improves the mIoU from 82.7$\%$ to 83.0$\%$. In conclusion, the local-patch separation improves the performances of CSWin-T and CSWin-S from 81.5$\%$ to 83.0$\%$ and 81.9$\%$ to 83.4$\%$. We also compare the LHFs to different convolutional variants in Table \ref{tab:9}. Using large-kernel convolutions or dilation convolutions does not improve the performance, which demonstrates the local operations are adequate. To better demonstrate the local separation capacity of LHFs, we visualize the feature maps of RS-CSWin-T. As shown in Fig. \ref{fig8} (c) and Fig. \ref{fig8} (d), our LHFs can extract clear boundary from relatively smooth feature maps. Fig. \ref{fig8} (b) and Fig. \ref{fig8} (d) show that feature maps of shallow layers contain more semantically agnostic edge information while our method reduces these noises and keeps the distinct semantic boundary. Fig. \ref{fig8} (e) and Fig. \ref{fig8} (f) show the comparison results of vanilla convolutions and LHFs, indicating that LHFs capture thinner boundary than vanilla convolutions.

\begin{table}[]
\caption{Ablation results of decoupled two-pathway network. 10$\times$lr denotes that we impose the 10$\times$ learning rate on LSBs. We increase the number of LSBs by reducing the sampling interval of the third stage. For CSWin-T, `3 LSBs', `5 LSBs', and `9 LSBs' mean the sampling intervals are 21, 7, and 3. For CSWin-S, `10 LSBs' means the sampling interval is 4}
\label{tab:8}
\begin{tabular}{p{155pt}p{30pt}<{\centering}p{30pt}<{\centering}}
\toprule
Model                              &GFLOPs                & mIoU                                                  \\ \midrule
CSWin-T                          & 146.3                     & 81.5                   \\
+3 LSBs                          & 153.2                     & 81.8                   \\
+5 LSBs                          & 158.0                     & 82.0                   \\
+9 LSBs                          & 167.6                     & 82.3                   \\
+9 LSBs, 10$\times$lr            & 167.6                     & 82.6                   \\
+9 LSBs, 10$\times$lr, AF        & 167.6                     & 82.7                   \\
+9 LSBs, 10$\times$lr, AF, LHFs, edge loss   &170.0           & \textbf{83.0}                   \\  \midrule
CSWin-S                            &211.5                      & 81.9                   \\
+10 LSBs, 10$\times$lr, AF         &259.1                      & 83.1                    \\
+10 LSBs, 10$\times$lr, AF, LHFs, edge loss& 261.8       & \textbf{83.4}                    \\\bottomrule
\end{tabular}
\end{table}

\begin{table}[]
\caption{Comparisons between different convolutional variants}
\label{tab:9}
\begin{tabular}{p{180pt}p{48pt}<{\centering}}
\toprule
Method                                     & mIoU           \\ \midrule
3$\times$3 convolutions                   & 82.7               \\
depth-wise large-kernel convolutions\cite{peng2017large}  & 82.6                   \\
multi-dilation convolutions\cite{pang2019towards}    & 82.5              \\
3$\times$3 convolutions + boundary loss                   & 82.8                   \\
LHFs + boundary loss                         & \textbf{83.0}               \\\bottomrule
\end{tabular}
\end{table}

\subsection{Comparisons with State-of-the-art Methods}
In this section, we demonstrate the efficiency of our method by comparing the accuracy, parameters, and FLOPs with previous start-of-the-art methods on five benchmarks.
For the sake of fairness, we mainly compared the methods based on per-pixel classification paradigm and mark with grey the methods  decoupling classification and mask prediction\cite{cheng2021per,cheng2022masked}. Note that we recalculate the GFLOPs of most compared methods except for methods that are not open source or too cumbersome by using the tools in \cite{mmseg2020}.

\subsubsection{Cityscapes}

\begin{table}[!h]
\center
\caption{Performance comparisons on Cityscapes val set. The GFLOPs of CNNs and pyramid ViTs are calculated under 1024$\times$1024 resolution while the GFLOPs of plain ViTs are calculated under 768$\times$768 resolution. For a fair comparison, reported data in the second line are obtained from $\mathit{MMSegmentation}$ model zoo, which are higher than those of original papers in most cases. $\dagger$ denotes the pre-train weights are from ViT-AugReg\cite{steiner2021train}.}
\label{tab:2}
\begin{tabular}{p{90pt}p{50pt}<{\centering}p{30pt}<{\centering}p{30pt}<{\centering}}
\toprule
Model                                     & mIoU(SS/MS)          & GFLOPs            & Params.     \\ \midrule
FCN\cite{long2015fully}                   & 75.5/76.6            & 1102            & 68.6M       \\
EncNet\cite{zhang2018context}             & 78.6/79.5            & 874             & 54.9M       \\
CCNet\cite{9133304}                & 79.5/80.7            & 1113            & 68.8M       \\
PSPNet\cite{zhao2017pyramid}              & 79.8/81.1            & 1025            & 68.0M       \\
HRNet\cite{wang2020deep}                   & 80.7/81.9           & 375             & 65.9M       \\
DANet\cite{fu2019dual}                    & 80.5/82.0            & 1108            & 68.8M       \\
DeepLabV3+\cite{chen2018encoder}          & 81.0/82.2            & 1016            & 62.6M       \\
OCRNet\cite{yuan2020object}               & 81.4/82.7            & 649             & 70.4M       \\ \midrule
SegFormer-B2\cite{xie2021segformer}       & 81.0/82.2            & 290             & 27.5M       \\
HRFormer-B + OCR\cite{yuan2021hrformer}   & 81.9/82.6            & 1120            & 56.2M           \\
SegFormer-B5\cite{xie2021segformer}       & 82.4/84.0            & 562             & 84.7M       \\
HRViT-b2\cite{gu2021hrvit}                & 82.81/-              & 138   & 20.8M           \\
HRViT-b3\cite{gu2021hrvit}                & 83.16/-              & 307            & 28.6M           \\
lawin-L\cite{yan2022lawin}                & -/84.4                & 899               & 201.2M            \\
\textcolor[RGB]{123,123,123}{Semask-L-Mask2\cite{jain2021semask}}& \textcolor[RGB]{123,123,123}{83.97/84.98} & \textcolor[RGB]{123,123,123}{900+}                 & \textcolor[RGB]{123,123,123}{222M}            \\
\textbf{RSSeg-CSWin-T}                      & 83.41/\textbf{84.58}           & 170& 23.2M    \\
\textbf{RSSeg-CSWin-S}                      & \textbf{83.64}/84.56  & 262             & 36.3M       \\ \midrule
Segmenter-B\cite{strudel2021segmenter}    & -/80.6                 & 350                    & 103.4M \\
\textbf{RSSeg-ViT-B}$\dagger$                         & \textbf{81.20/83.26} & 338   & 100.7M \\\midrule
Segmenter-L$\dagger$\cite{strudel2021segmenter}      & 79.1/81.3            & 1045                 & 333.2M \\
StructToken-PWE-L$\dagger$\cite{lin2022structtoken}  & 80.05/82.07          & 1050+              & 364M \\
SETR-PUP-L\cite{zheng2021rethinking}                & 79.34/82.15          & 1082                 & 318.3M      \\
\textbf{RSSeg-ViT-L}$\dagger$                         & \textbf{81.30/83.45}   & 1039   & 330.1M                   \\ \bottomrule
\end{tabular}
\end{table}

\begin{figure}
\centerline{\includegraphics[width=\columnwidth]{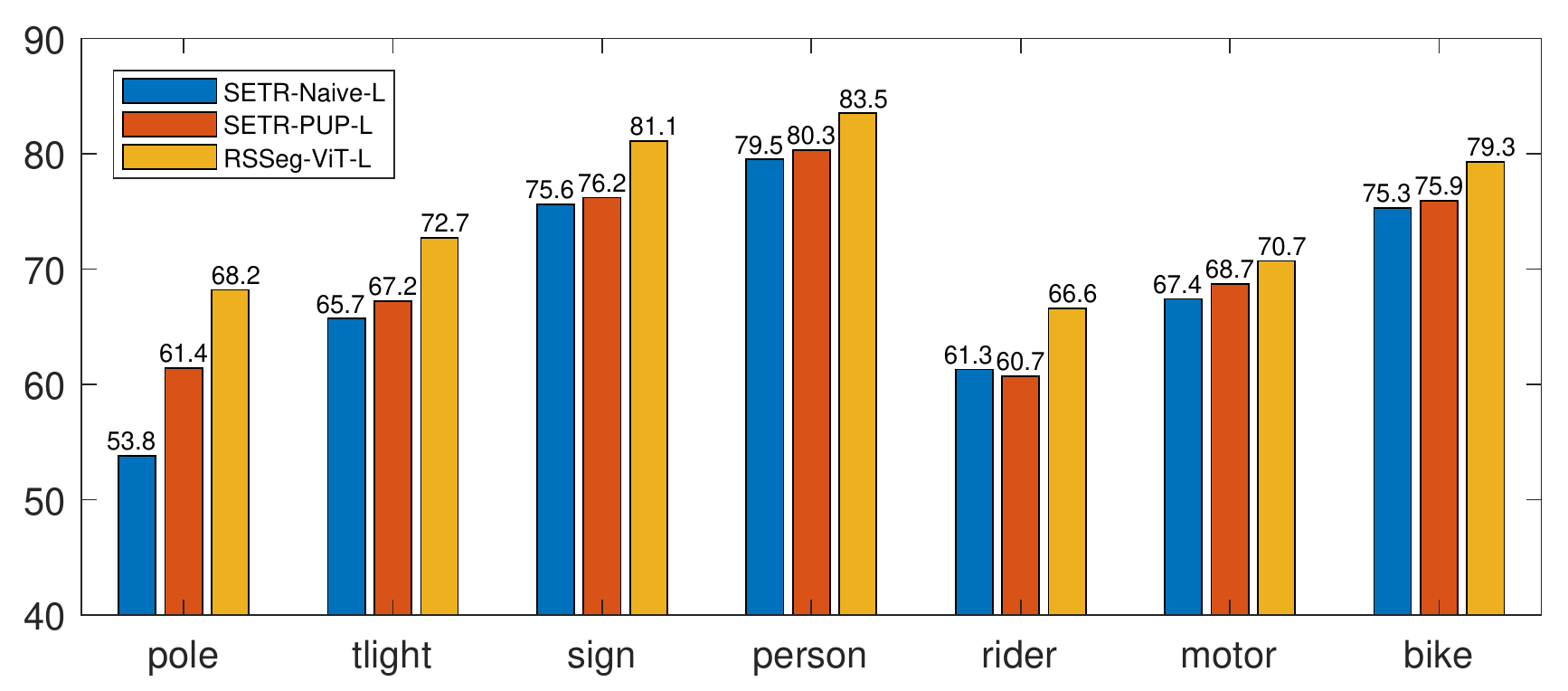}}
\caption{Singe-scale segmentation results of small/thin objects on Cityscapes val set.}
\label{fig9}
\end{figure}

As shown in Table \ref{tab:2}, our RSSeg-ViT-L achieves 83.45$\%$ mIoU and outperforms the previous state-of-the-art method by 1.3$\%$. It can be also seen that plain ViTs usually perform inferior to CNNs and pyramid ViTs and previous methods with extra attention such as Segmenter and StructToken do not improve the performance.
We investigate the segmentation results of small or thin objects, as shown in Fig. \ref{fig9}. It can be seen that SETR-PUP improves the average accuracy of small objects by cascaded upsampling and refining while our method achieves substantial improvements on all small objects.
In addition to the worse plain ViTs, pyramid ViTs based models still possibly generate smoother prediction results compared to the state-of-the-art CNNs models. Our method can also generate sharper results for them. As can be seen from Table \ref{tab:2}, CSwins integrated with our decoupled two-pathway network achieve state-of-the-art performances with much fewer parameters and computation. For example, RS-CSWin-T achieves similar performance with lawin-L with only 1/5 computation and 1/8 parameters of the latter. The best model Semask-L-Mask2 uses the Swin-L as the backbone and utilizes the computationally intensive framework of Mask2Former\cite{cheng2022masked}. We do not train bigger models since the performance gain from CSWin-T to CSWin-S is little. It is noticeable that RS-CSWin-S outperforms the HRViT-b3 by 0.5$\%$ mIoU with lower GFLOPs, whereas the latter contains a series of improvements on the original cross-shaped self-attention of CSWin and the well-designed network architecture, which needs to be retrained on ImageNet-1K. It demonstrates that existing advanced pyramid ViTs plus our local-patch separation are enough for the fine segmentation scenarios.

\noindent\textbf{Scales of the multi-scale test} Under normal circumstances, smaller test resolution leads to better intra-class consistency and larger test resolution leads to sharper semantic boundary. Therefore, the previous methods combine all the predictions to obtain the best results. Since the ViTs are capable of capturing rich context and have very large receptive fields, reducing the test resolution possibly does not improve the intra-class consistency and leads to smoother predictions, which hampers the ensemble performance. As shown in Table \ref{tab:10}, we do not need to shrink the images for multi-scale test on Cityscapes which contains numerous small/thin objects. And once again, it reflects the difference between semantic segmentation with ViTs and CNNs.


\begin{table}[]
\caption{Multi-scale test results with different scale factors. The stride is 0.25. CNN based models of which weights can be obtained from the open source toolbox\cite{mmseg2020} are listed in the second line}
\label{tab:10}
\begin{tabular}{p{63pt}p{40pt}<{\centering}p{50pt}<{\centering}p{50pt}<{\centering}}
\toprule
Model                                     & SS (1.0)     &MS (0.5$\sim$1.75)    &MS (1.0$\sim$1.75)  \\ \midrule
PSPNet(R50)                               & 79.61        &\textbf{80.36}                 &80.15(-0.21)               \\
DeepLabV3+(R101)                          & 80.92        &\textbf{81.88}                 &81.55(-0.33)               \\
OCRNet(HRW48)                             & 81.35        &\textbf{82.70}                 &82.32(-0.38)               \\\midrule
RSSeg-CSWin-T                             & 83.41        &84.39                 &\textbf{84.58}(+0.19)                   \\
RSSeg-CSWin-S                             & 83.64        &\textbf{84.64}              &84.56(-0.08)               \\
RSSeg-ViT-B                               & 81.20        &82.95                 &\textbf{83.26}(+0.31)               \\
RSSeg-ViT-L                               & 81.30        &83.20                  &\textbf{83.45}(+0.25)            \\\bottomrule
\end{tabular}
\end{table}

\subsubsection{ADE20K}

\begin{figure}
\centerline{\includegraphics[width=\columnwidth]{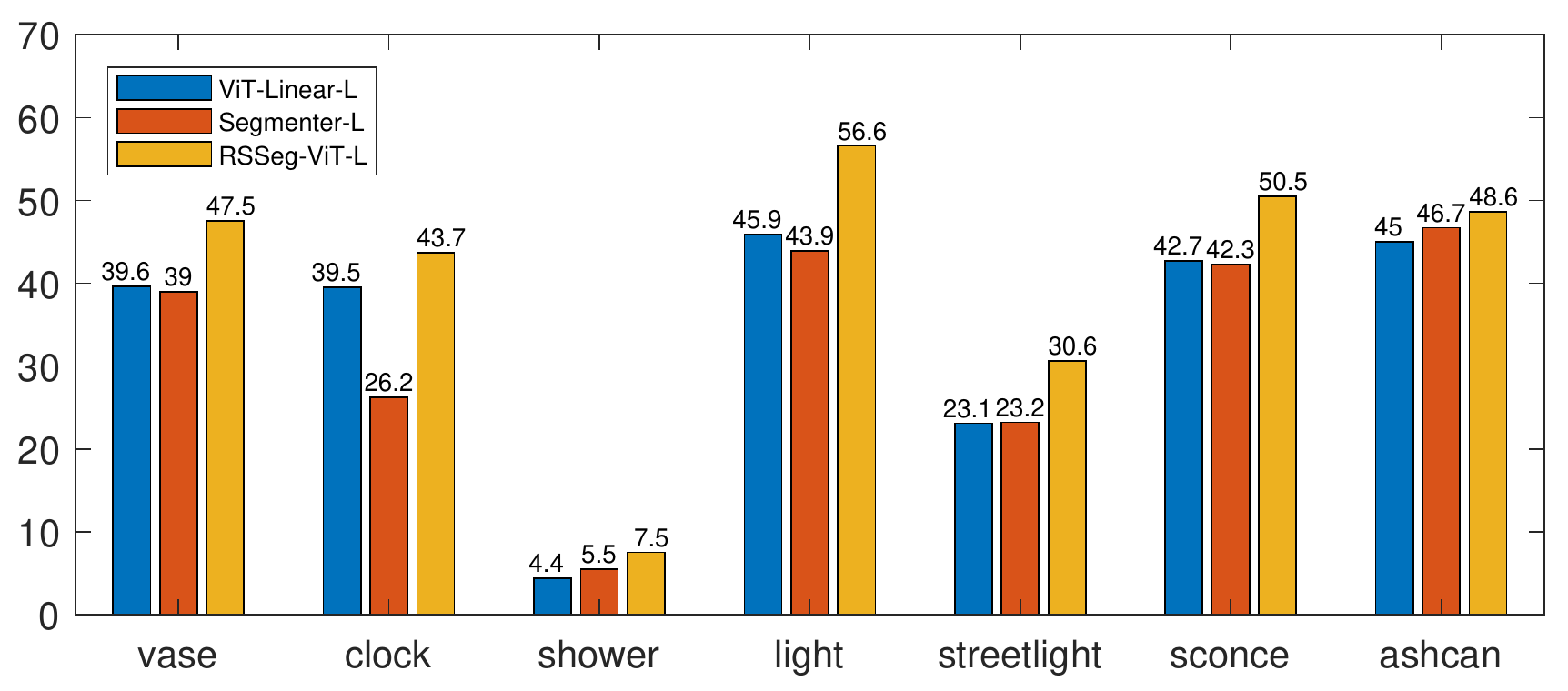}}
\caption{Singe-scale segmentation results of small objects on ADE20K val set.}
\label{fig10}
\end{figure}

\begin{table}[]
\center
\caption{Performance comparisons on ADE20K dataset. We mainly list the state-of-the-art methods of CNNs and ViTs. The GFLOPs are calculated under 512$\times$512 or 640$\times$640 resolution according to the test crop size. $\ast$ denotes the result is reproduced by ourselves. $\dagger$ denotes the pre-train weights are from ViT-AugReg\cite{steiner2021train}.}
\label{tab:3}
\begin{tabular}{p{95pt}p{35pt}<{\centering}p{50pt}<{\centering}p{25pt}<{\centering}}
\toprule
Model                                   &Encoder    & mIoU(SS/MS)      & GFLOPs                 \\ \midrule
PSPNet\cite{zhao2017pyramid}            &R101    & 41.96/43.29         &  -             \\
PSANet\cite{zhao2018psanet}             &R101    & 42.75/43.77         & -              \\
OCRNet\cite{yuan2020object}             &R101    & -/45.28         & -       \\
CCNet\cite{9133304}                      &R101    & -/45.76            & -       \\
CPNet\cite{yu2020context}               &R101    & 45.39/46.27         &-           \\
STLNet\cite{zhu2021learning}            &R101    & -/46.5              &-         \\
UperNet\cite{liu2022convnet} &CNext-L  & -/53.7     &-         \\ \midrule
SegFormer-B5\cite{xie2021segformer}     &MiT-B5  & 51.0/51.8                    &183            \\
UperNet\cite{liu2021swin}        &SWin-L        & -/53.5                 &647            \\
SenFormer\cite{bousselham2021efficient} &Swin-L  & 53.1/54.2                    &546  \\
Lawin\cite{yan2022lawin}                &SWin-L  & 54.7/55.2                    &351           \\
\textcolor[RGB]{123,123,123}{MaskFormer\cite{cheng2021per}}      &\textcolor[RGB]{123,123,123}{SWin-L}  & \textcolor[RGB]{123,123,123}{54.1/55.6}      &\textcolor[RGB]{123,123,123}{375}           \\
UperNet\cite{dong2021cswin}     &CSWin-L       & 53.4/55.7              &619            \\
\textcolor[RGB]{123,123,123}{Mask2Former\cite{cheng2022masked}}       &\textcolor[RGB]{123,123,123}{SWin-L}  & \textcolor[RGB]{123,123,123}{56.1/57.3}      &\textcolor[RGB]{123,123,123}{403}           \\
\textcolor[RGB]{123,123,123}{SeMask-Mask2\cite{jain2021semask}}&\textcolor[RGB]{123,123,123}{SWin-L}   &\textcolor[RGB]{123,123,123}{56.41/57.52}  &\textcolor[RGB]{123,123,123}{400+} \\\midrule
Segmenter\cite{strudel2021segmenter}       &ViT-B & 48.5/50.0                     &130         \\
StructToken-SSE\cite{lin2022structtoken}   &ViT-B & 50.72/51.85                   &150+ \\
\textcolor[RGB]{123,123,123}{SegViT\cite{zhang2022segvit}}   &\textcolor[RGB]{123,123,123}{ViT-B}  & \textcolor[RGB]{123,123,123}{51.3/53.0} & \textcolor[RGB]{123,123,123}{121} \\\midrule
UperNet$\ast$\cite{steiner2021train}     &ViT-B$\dagger$ & 49.82/50.98                    & 340 \\
ViT-Adapter-UperNet\cite{chen2022vision}     &ViT-B$\dagger$ & 51.9/52.5                    & 350+ \\
\textbf{RSSeg-ViT}                      &ViT-B$\dagger$         & \textbf{53.11/54.09} &128            \\ \midrule
ViT-Linear$\ast$\cite{strudel2021segmenter}    &ViT-L$\dagger$ &51.04/52.30          &612\\
Segmenter\cite{strudel2021segmenter}       &ViT-L$\dagger$ & 51.8/53.6             &672         \\
StructToken-CSE\cite{lin2022structtoken}   &ViT-L$\dagger$ & 52.84/54.18           &700+ \\
Segmenter+RankSeg\cite{he2022rankseg}       &ViT-L$\dagger$ & 52.6/54.4            &672  \\
ViT-Adapter-UperNet\cite{chen2022vision}     &ViT-L$\dagger$ & 53.4/54.4                    &610+  \\
\textcolor[RGB]{123,123,123}{SegViT\cite{zhang2022segvit}}         &\textcolor[RGB]{123,123,123}{ViT-L$\dagger$}  & \textcolor[RGB]{123,123,123}{54.6/55.2} &\textcolor[RGB]{123,123,123}{638}  \\
\textbf{RSSeg-ViT}                      &ViT-L$\dagger$         & \textbf{54.61/55.98} &670          \\ \midrule
BEiT-Linear$\ast$\cite{strudel2021segmenter}    &BEiT-L &54.86/55.49                     & 613\\
UperNet\cite{bao2021beit}     &BEiT-L            &56.7/57.0                     & 1993\\
UperNet-RR\cite{cui2022region}         &BEiT-L       &57.2/57.7        & 1993\\
UperNet+RankSeg\cite{he2022rankseg}    &BEiT-L       &57.0/57.8      & 1998\\
ViT-Adapter-UperNet\cite{chen2022vision}     &BEiT-L       & \textbf{58.0/58.4}                    & 2000+ \\
\textcolor[RGB]{123,123,123}{ViT-Adapter-Mask2\cite{chen2022vision}}&\textcolor[RGB]{123,123,123}{BEiT-L} & \textcolor[RGB]{123,123,123}{58.3/59.0}  & \textcolor[RGB]{123,123,123}{1400+} \\
\textbf{RSSeg-ViT}                      &BEiT-L            & 57.89/\textbf{58.37}       &670            \\ \bottomrule
\end{tabular}
\end{table}
On the more challenging ADE20K dataset, ViT based models have outperformed CNN based models by a wide margin. As shown in Table \ref{tab:3}, our method achieves 54.1$\%$ mIoU using ViT-B, even outperforming the Segmenter with ViT-L by 0.5$\%$ mIoU but only costing 1/5 computation. When using ViT-L, the multi-scale test accuracy reaches 56.0$\%$ mIoU, surpassing the previous state-of-the-art method SegViT by 0.8$\%$ mIoU. Note that it utilizes the decoupled supervision of Mask2Former.
Using BEiT-L as the backbone, we achieve 58.4$\%$ mIoU, the same result as ViT-Adapter-UperNet while our method only needs 1/3 computation requirements of it. For a more intuitive representation of the validness of the proposed method, we train the ViT-L and BEiT-L with a linear decoder using the identical training settings, which is a simple and effective baseline used in our method and Segmenter\cite{strudel2021segmenter}. Our method improves the performance by 3.7$\%$ mIoU and 2.9$\%$ mIoU upon the very strong baselines while only increasing about 9$\%$ FLOPs. We also select the seven smallest objects and show their segmentation results in Fig. \ref{fig10}.
Using two extra transformer layers, there is even a slight drop in the segmentation accuracy of small objects for Segmenter. As a contrast, our method still achieves remarkable performance gains under more challenging scenarios, demonstrating it effectively separates the smooth representations and brings out the small objects.


\subsubsection{PASCAL Context}

Since the original paper of Segmenter only reports the results of 60 classes, we retrain the models with 40K iterations using AdamW optimizers. As shown in Table \ref{tab:4}, our method achieves 67.5$\%$ mIoU with ViT-L, 2.5$\%$ mIoU higher than Segmenter and 2.2$\%$ mIoU higher than SegViT using the same backbone. Our method even surpasses the ViT-Adapter-Mask2 by 1$\%$ mIoU with BEiT-B. We achieve 68.9$\%$ mIoU with BEiT-L, which is the best result for Pascal Context to our best knowledge. Note that ViT-Adapter-UperNet and ViT-Adapter-Mask2 are much more cumbersome than our methods.

\begin{table}[]
\centering
\caption{State-of-the-art results on PASCAL Context dataset. We report the mIoU of 59 classes which is more widely adopted. $\ast$ denotes the result reproduced by ourselves. $\dagger$ denotes the pre-train weights are from ViT-AugReg\cite{steiner2021train}.}
\label{tab:4}
\begin{tabular}{p{95pt}p{60pt}<{\centering}p{60pt}<{\centering}}
\toprule
Model                               & Encoder                            & mIoU(SS/MS)                      \\ \midrule
PSPNet\cite{zhao2017pyramid}        & R101                               & -/47.8                       \\
EncNet\cite{zhang2018context}       & R101                               & -/51.7                    \\
DANet\cite{fu2019dual}              & R101                               & -/52.6                 \\
CPNet\cite{yu2020context}           & R101                               & -/53.9                 \\
OCRNet\cite{yuan2020object}         & R101                               & -/54.8                 \\
CAA\cite{huang2021channelized}      & R101                               & -/55.0                 \\
STLNet\cite{zhu2021learning}        & R101                               & -/55.8                \\
OCRNet\cite{yuan2020object}         & HRW48                              & -/56.2                 \\
CAA\cite{huang2021channelized}      & EfficientNet-B7                    & -/60.5                 \\
CAR\cite{huang2022car}              & ConvNext-L                         & 62.97/64.12           \\\midrule
CAR\cite{huang2022car}              &Swin-L                              & 60.68/62.21           \\
SenFormer\cite{bousselham2021efficient}&Swin-L                         & 63.1/64.5             \\ \midrule
DPT\cite{ranftl2021vision} & ViT-Hybrid-L               & -/60.5                  \\
Segmenter$\ast$\cite{strudel2021segmenter}& ViT-L$\dagger$                & 63.98/65.01                  \\
\textcolor[RGB]{123,123,123}{SegViT\cite{zhang2022segvit}}     &\textcolor[RGB]{123,123,123}{ViT-L$\dagger$}                 & \textcolor[RGB]{123,123,123}{-/65.3}     \\
\textbf{RSSeg-ViT}                  & ViT-L$\dagger$                     & \textbf{66.85/67.48}                \\ \midrule
\textcolor[RGB]{123,123,123}{ViT-Adapter-Mask2\cite{chen2022vision}} & \textcolor[RGB]{123,123,123}{BEiT-B}                   & \textcolor[RGB]{123,123,123}{64.0/64.4}               \\
\textbf{RSSeg-ViT}                  & BEiT-B                        & \textbf{64.95/65.39}                 \\\midrule
ViT-Adapter-UperNet\cite{chen2022vision}    & BEiT-L                   & 67.0/67.5               \\
\textcolor[RGB]{123,123,123}{ViT-Adapter-Mask2\cite{chen2022vision}} & \textcolor[RGB]{123,123,123}{BEiT-L}                   & \textcolor[RGB]{123,123,123}{67.8/68.2}               \\
\textbf{RSSeg-ViT}                  & BEiT-L                        & \textbf{68.48/68.87}  \\\bottomrule
\end{tabular}
\end{table}

\subsubsection{COCOStuff-10K}

\begin{table}[]
\centering
\caption{State-of-the-art results on COCOStuff-10K dataset. $\dagger$ denotes the pre-train weights are from ViT-AugReg\cite{steiner2021train}. $\ddagger$ denotes the result is reported in \cite{he2022rankseg}.}
\label{tab:5}
\begin{tabular}{p{95pt}p{60pt}<{\centering}p{60pt}<{\centering}}
\toprule
Model                                    & Encoder                            & mIoU(SS/MS)                      \\ \midrule
SeMask\cite{jain2021semask}              &Swin-L                             & 47.47/48.54  \\
CAR\cite{huang2022car}                   &ConvNext-L                          & 49.03/50.01           \\
SenFormer\cite{bousselham2021efficient}  &Swin-L                             & 49.8/51.5          \\\midrule
Segmenter$\ddagger$\cite{strudel2021segmenter}& ViT-L$\dagger$                & 45.5/47.1                  \\
Segmenter+RankSeg\cite{he2022rankseg}& ViT-L$\dagger$                & 46.6/47.9                  \\
StructToken-SSE\cite{lin2022structtoken} & ViT-L$\dagger$                         & -/49.1         \\
\textcolor[RGB]{123,123,123}{SegViT\cite{zhang2022segvit}}             &\textcolor[RGB]{123,123,123}{ViT-L$\dagger$}                 & \textcolor[RGB]{123,123,123}{-/50.3}     \\
\textbf{RSSeg-ViT}                       & ViT-L$\dagger$                   &\textbf{50.42/51.99}                 \\ \midrule
UperNet$\ddagger$\cite{bao2021beit}     &BEiT-L            &49.7/49.9                     \\
UperNet+RankSeg\cite{he2022rankseg}     &BEiT-L            &49.9/50.3                     \\
ViT-Adapter-UperNet\cite{chen2022vision}   & BEiT-L                   & 51.0/51.4               \\
\textbf{RSSeg-ViT}                       & BEiT-L                      &\textbf{51.91/52.63}      \\  \bottomrule
\end{tabular}
\end{table}

Table \ref{tab:5} shows the results on COCO-Stuff-10K dataset. With ViT-L as the backbone, our RSSeg-ViT achieves 52.0$\%$ mIoU, outperforming the previous state-of-the-art method by 1.7$\%$ mIoU and ViT-Adapter-UperNet with BEiT-L by 0.6$\%$ mIoU. With BEiT-L, RSSeg-ViT outperforms the ViT-Adapter-UperNet by 1.2$\%$ mIoU.

\subsubsection{COCOStuff-164K}

\begin{table}[]
\centering
\caption{State-of-the-art results on COCOStuff-164K dataset. Models with pyramid ViTs are trained with a crop size of 512$\times$512 while models with  plain ViTs are trained with a crop size of 640$\times$640. $\dagger$ denotes the pre-train weights use the ImageNet 22K. $\ddagger$ denotes the model is trained with 320K iterations (the default is 80K). $\ast$ denotes the result is reproduced by ourselves.}
\label{tab:14}
\begin{tabular}{p{95pt}p{60pt}<{\centering}p{60pt}<{\centering}}
\toprule
Model                                    & Encoder                          & mIoU(SS/MS)                      \\ \midrule
SegFormer\cite{xie2021segformer}         & MiT-B5                           & 46.7/-           \\
Lawin\cite{yan2022lawin}                 & MiT-B5                           & 47.5/-          \\
UperNet-RR$\ddagger$\cite{cui2022region} & Swin-B$\dagger$                  & 48.21/49.20  \\\midrule
ViT-Linear$\ast$                        & ViT-L$\dagger$                   & 48.81/49.69   \\
Segmenter$\ast$\cite{strudel2021segmenter}& ViT-L$\dagger$                   & 49.13/50.11                  \\
\textbf{RSSeg-ViT}                          & ViT-L$\dagger$                   &\textbf{50.47/51.04}                 \\ \midrule
ViT-Adapter-UperNet\cite{chen2022vision} & BEiT-L$\dagger$                  & 50.5/50.7               \\
\textbf{RSSeg-ViT}                          & BEiT-L$\dagger$                  &\textbf{51.40/51.70}      \\  \bottomrule
\end{tabular}
\end{table}

We benchmark several segmentation models with plain ViTs on COCOstuff-164K, a much larger semantic segmentation dataset. As shown in Table \ref{tab:14}, our method opens up a lead over ViT-Linear, Segmenter, and ViT-Adapter-UperNet with different backbones. Our method using ViT-L also outperforms ViT-Adapter-UperNet with BEiT-L. As far as we know, we achieve the best results under the training crop size of 640$\times$640.

\subsection{Robustness Evaluation}

\begin{table}[]\scriptsize
\centering
\caption{Robustness evaluation on Cityscapes and ADE20K. We report the average results of `Blur', `Noise', `Digital', and `Weather'. Each of them contains four types of corruptions. `Corr.' denotes the average result of all 16 types of corruptions. `Rete.' denotes the retention which is defined as the ratio of `Corr.' to `Clean'. $\dag$ denotes the result is reported in \cite{kamann2020benchmarking}}
\label{tab:19}
\begin{tabular}{p{60pt}|p{13pt}<{\centering}p{13pt}<{\centering}p{13pt}<{\centering}|p{13pt}<{\centering}p{13pt}<{\centering}p{13pt}<{\centering}p{13pt}<{\centering}}
\toprule
Model                     &Clean  &Corr.  &Rete. &Blur  &Noise &Digit.  &Weath.    \\ \midrule
\multicolumn{8}{c}{Cityscapes}      \\ \midrule
GSCNN$\dag$\cite{takikawa2019gated}   &80.9 &40.0 &49.4 &54.8 &9.9 &55.1   & 40.1   \\
DLV3+(R101)\cite{chen2018encoder}   &80.9 &43.9 &54.2 &53.4 &22.8 &58.3   &41.1   \\
SegFormer-B5\cite{xie2021segformer}   &82.4 &65.8 &79.9 &67.9 &61.6 &74.4   &59.4    \\
FAN-L-Hybrid\cite{zhou2022understanding}   &82.3 &68.7 &83.5 &- &- &-   &-    \\\midrule
CSWin-S\cite{dong2021cswin}   &81.9 &66.4 &\textbf{81.1} &68.7 &61.5 &74.2   &61.2    \\
RSSeg-CSWin-S  &83.6 &66.8 &79.9 &69.5 &59.0 & 75.3  &63.4    \\\midrule
ViT-Linear-B\cite{steiner2021train}       &79.5 &65.9 &82.9 &69.1 &63.2 &71.8  &59.6   \\
RSSeg-ViT-B   &81.2 &67.9 &\textbf{83.6} &70.2  &65.4 &74.4  &61.4  \\\midrule
SETR-PUP-L\cite{zheng2021rethinking}       &79.3 &66.1 &83.3 &70.5 &63.4 &70.3  &60.1   \\
RSSeg-ViT-L   &81.3 &70.1 &\textbf{86.3} &72.1  &70.3 &75.7  &62.5  \\\midrule
\multicolumn{8}{c}{ADE20K}  \\ \midrule
DLV3+(R101)\cite{chen2018encoder}   &44.6 &26.5 &59.5 &24.9 &20.5 &35.9 &24.8    \\
SegFormer-B5\cite{xie2021segformer}   &51.0 &39.2  &76.8  &34.1 &39.6 &45.8   &37.3    \\\midrule
ViT-Linear-B\cite{steiner2021train}   &49.3 &40.1 &81.3&37.5 &40.7 &44.8  &37.3   \\
RSSeg-ViT-B   &53.1 &43.8 &\textbf{82.5} &40.9  &44.3 &48.5  &41.6  \\\bottomrule
\end{tabular}
\end{table}

Model robustness is crucial for many safety-critical tasks such as autonomous driving\cite{kamann2020benchmarking}. Recent works have shown that ViTs learn more robust representations than CNNs\cite{bhojanapalli2021understanding,mahmood2021robustness,paul2022vision,xie2021segformer} in image classification and semantic segmentation and some works propose more robust ViTs\cite{zhou2022understanding,mao2022towards}, which open up the vast perspective of ViTs in improving the robustness of deep learning. However, the robustness of plain ViTs in semantic segmentation has not been explored. Since the strong robustness comes from the learned representations of ViTs and we separate such coherent representations, we evaluate the robustness of our method and see its effect on robustness on Cityscapes and ADE20K following precedent works\cite{kamann2020benchmarking,xie2021segformer,zhou2022understanding}. The results are presented in Table \ref{tab:19} and we have several observations. First, ViT-Linear-B has the similar encoder size with SegFormer-B5 but has higher retention. It indicates that plain ViTs made up of only self-attention and MLPs are very robust learners to natural corruptions in semantic segmentation, and the improvements in robustness really come from self-attention. Second, the performance gain of the decoupled two-pathway network does not transfer from clean images to corrupted images ideally. The similar phenomenon also happens to GSCNN. This may be due to the fact that the high-resolution representations with rich local information learned by convolutions is not robust compared to self-attention. However, our method does not lead to severe degradation like GSCNN, indicating that our method is more robust. Third, plain ViTs with our representation separation perform more robust than the pure attention baselines. On Cityscapes, RSSeg-ViT-B even surpasses the FAN-L-Hybrid\cite{zhou2022understanding}, a well-designed fully attentional network for stronger robustness with a similar encoder size. With greater model capacity, our RSSeg-ViT-L achieves the unprecedented robustness to natural corruptions on Cityscapes. This indicates the interesting properties of our representation separation framework, which maintain or even enhance the robustness by increasing the representational discrepancy.

\subsection{Visualized Results}
\label{Visualized Results}

\begin{figure}
\centering
\begin{minipage}{1.1in}
\includegraphics[width=1.1in]{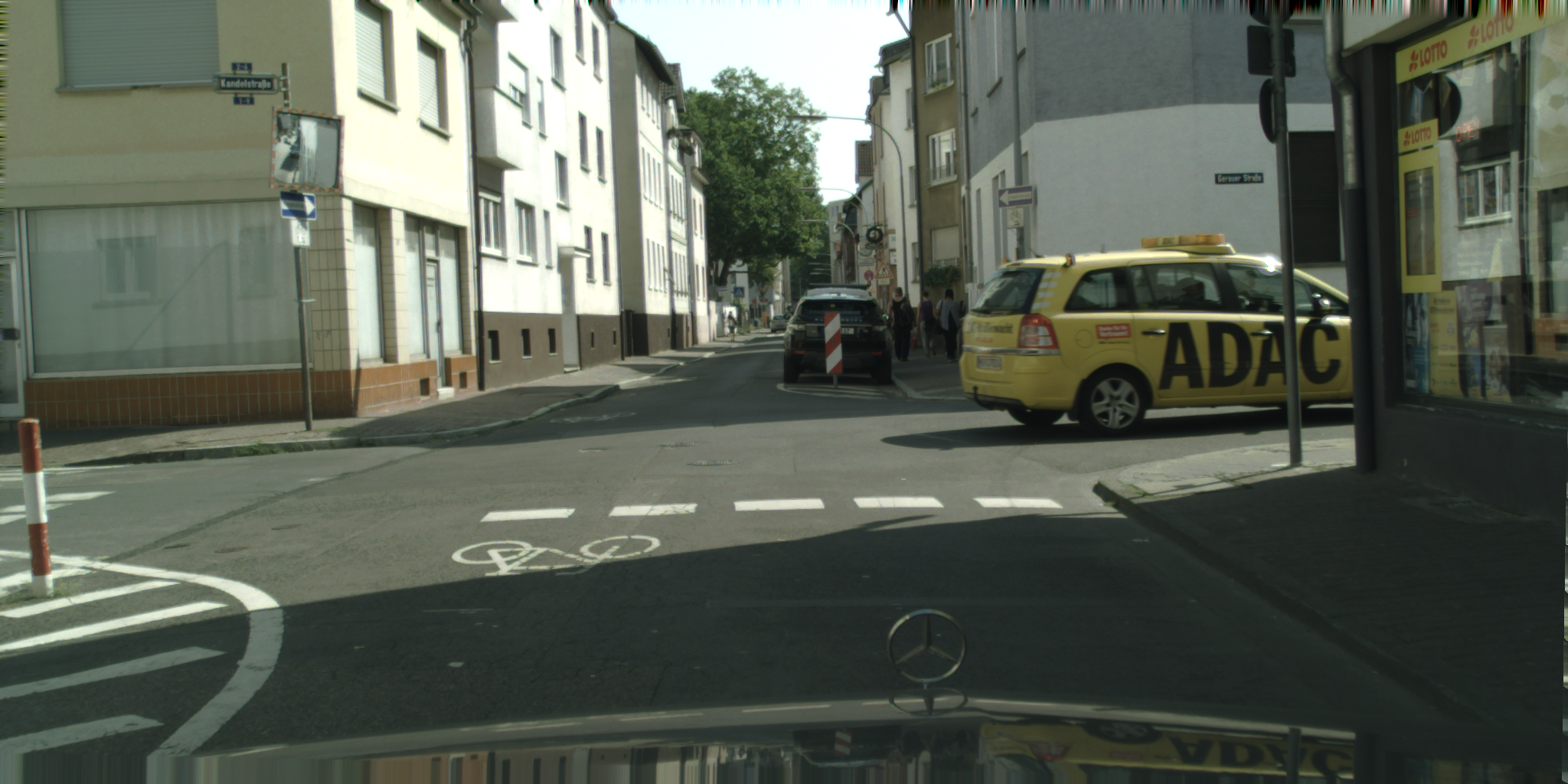}
\end{minipage}
\begin{minipage}{1.1in}
\includegraphics[width=1.1in]{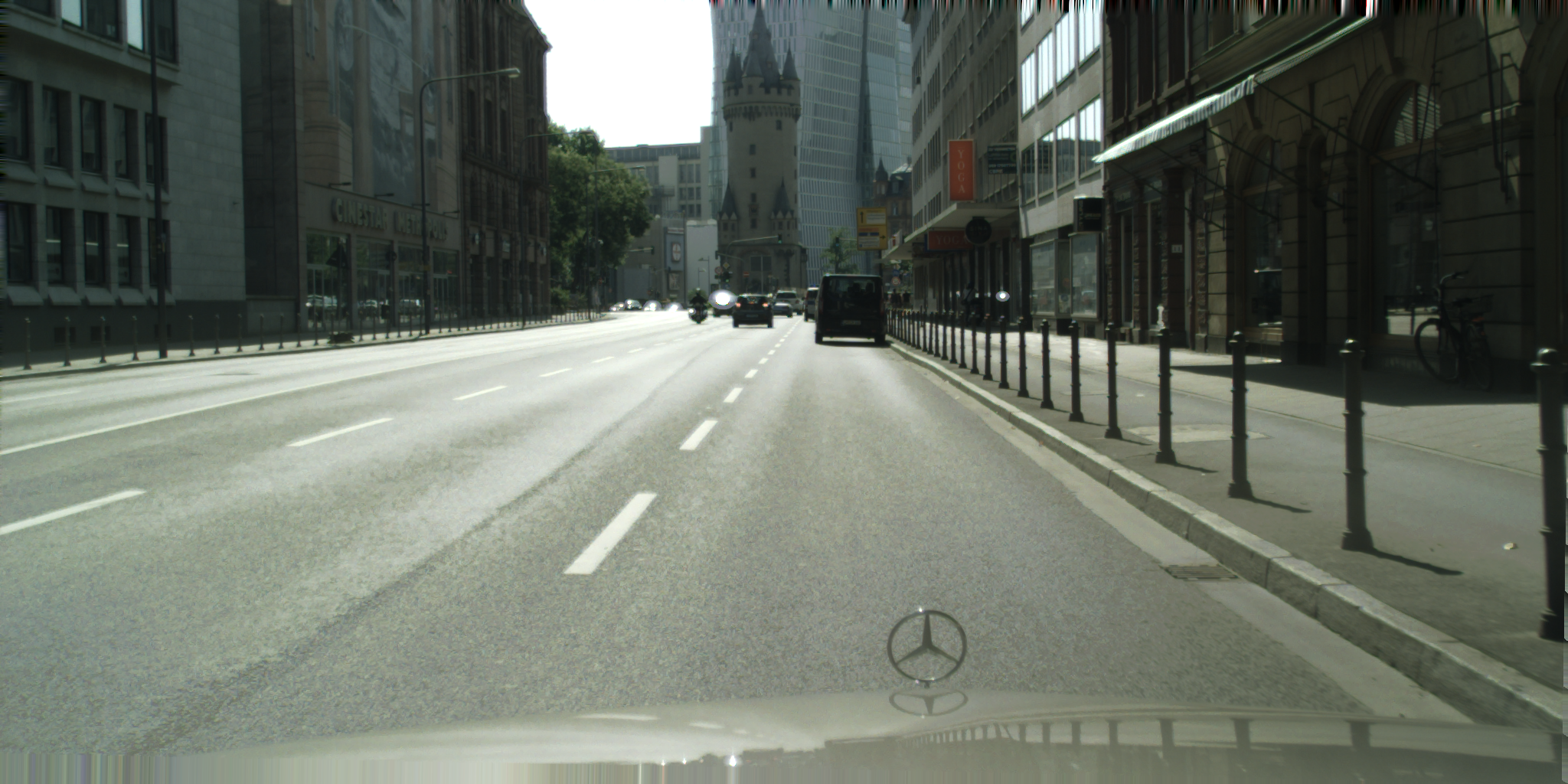}
\end{minipage}
\vspace{0.3em}
\begin{minipage}{1.1in}
\includegraphics[width=1.1in]{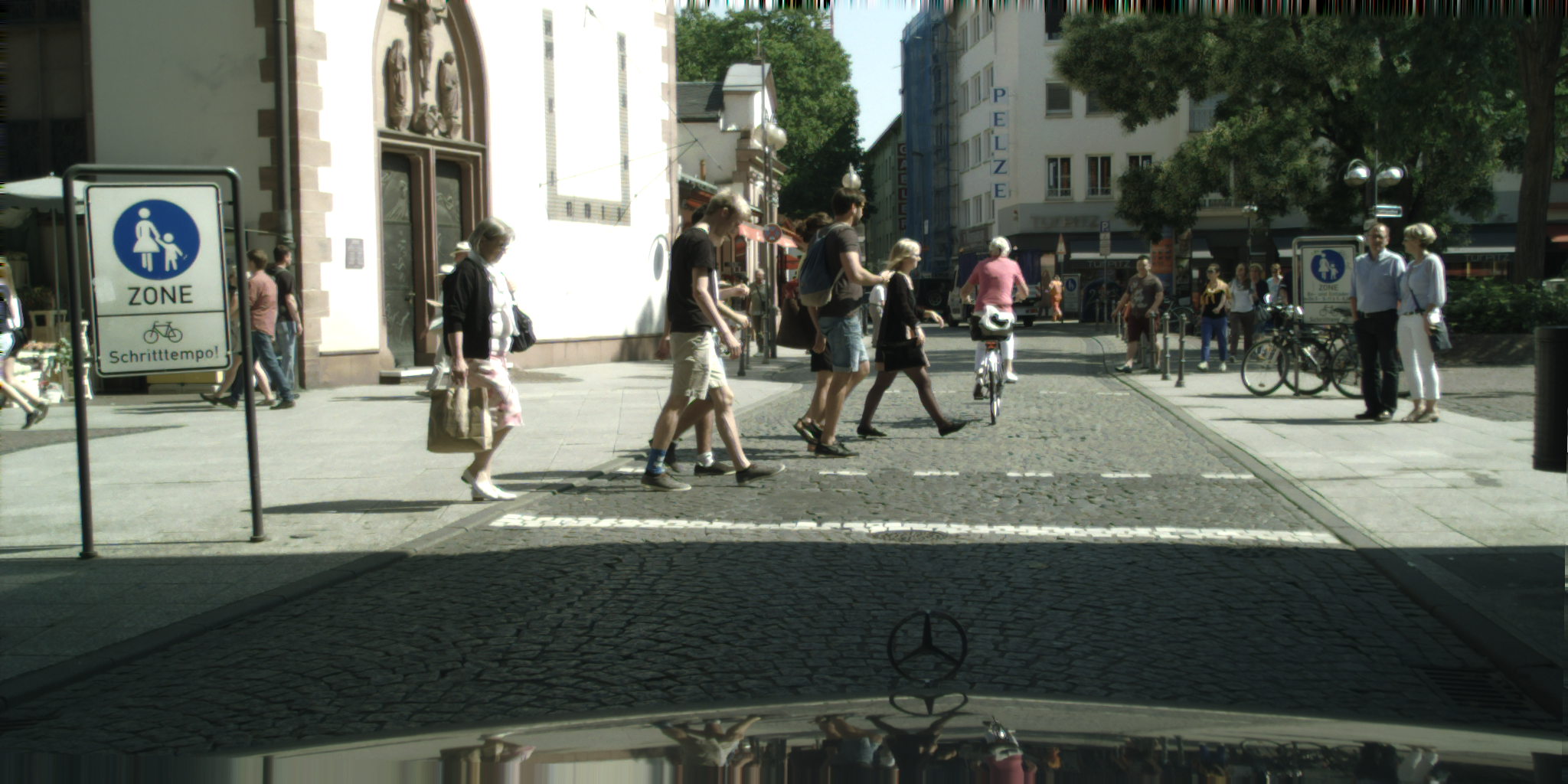}
\end{minipage}
\begin{minipage}{1.1in}
\includegraphics[width=1.1in]{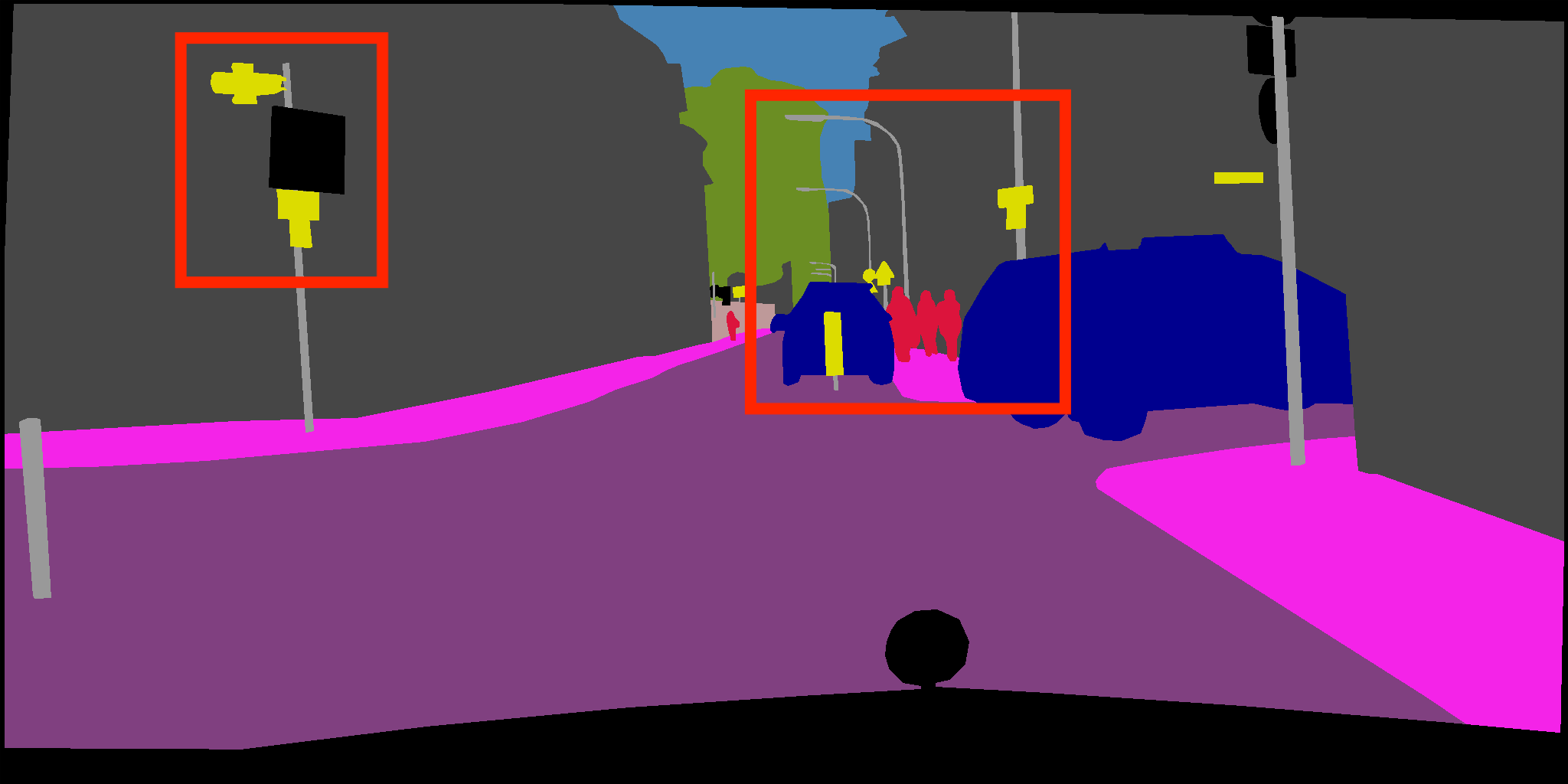}
\end{minipage}
\begin{minipage}{1.1in}
\includegraphics[width=1.1in]{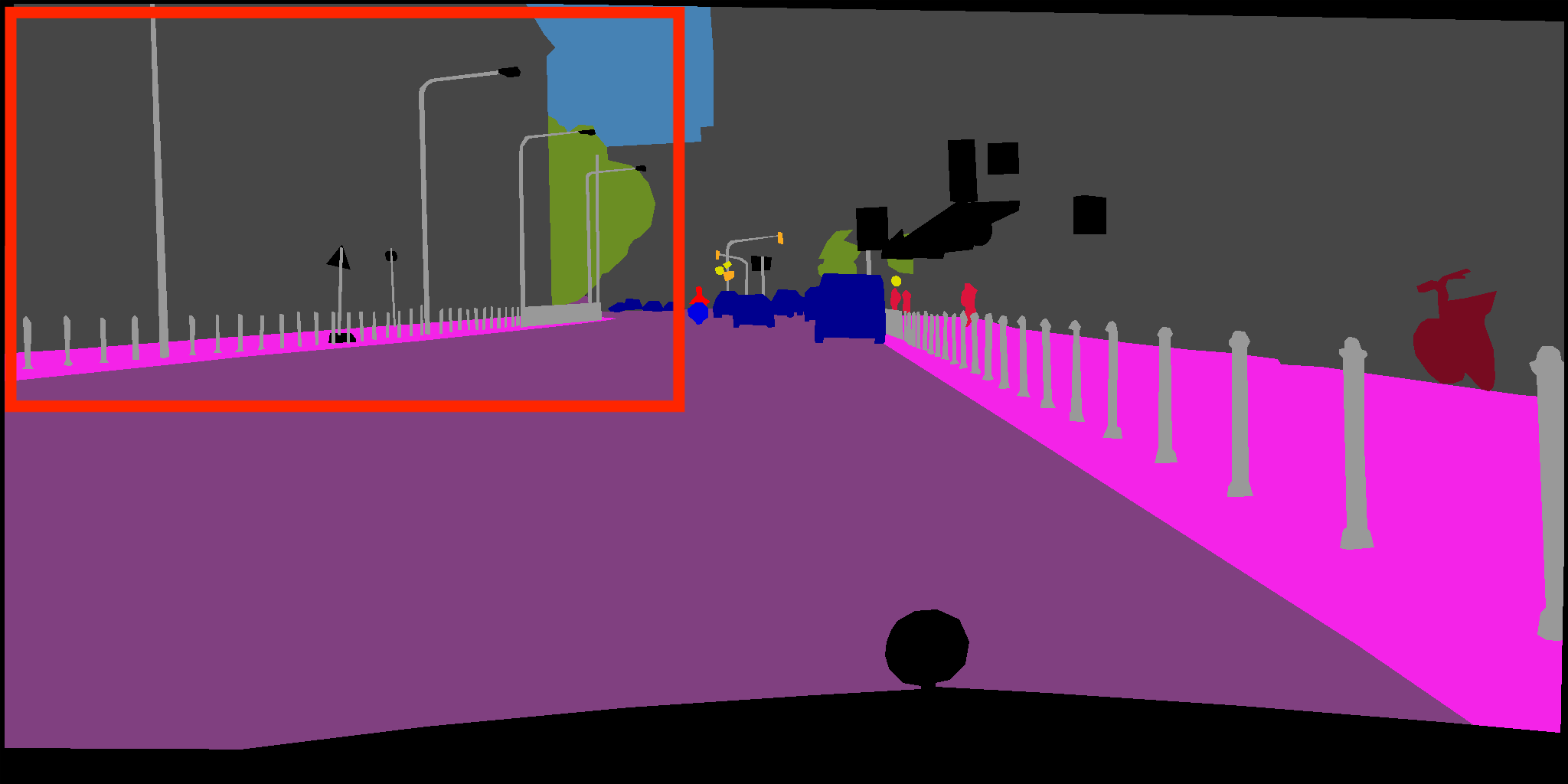}
\end{minipage}
\vspace{0.3em}
\begin{minipage}{1.1in}
\includegraphics[width=1.1in]{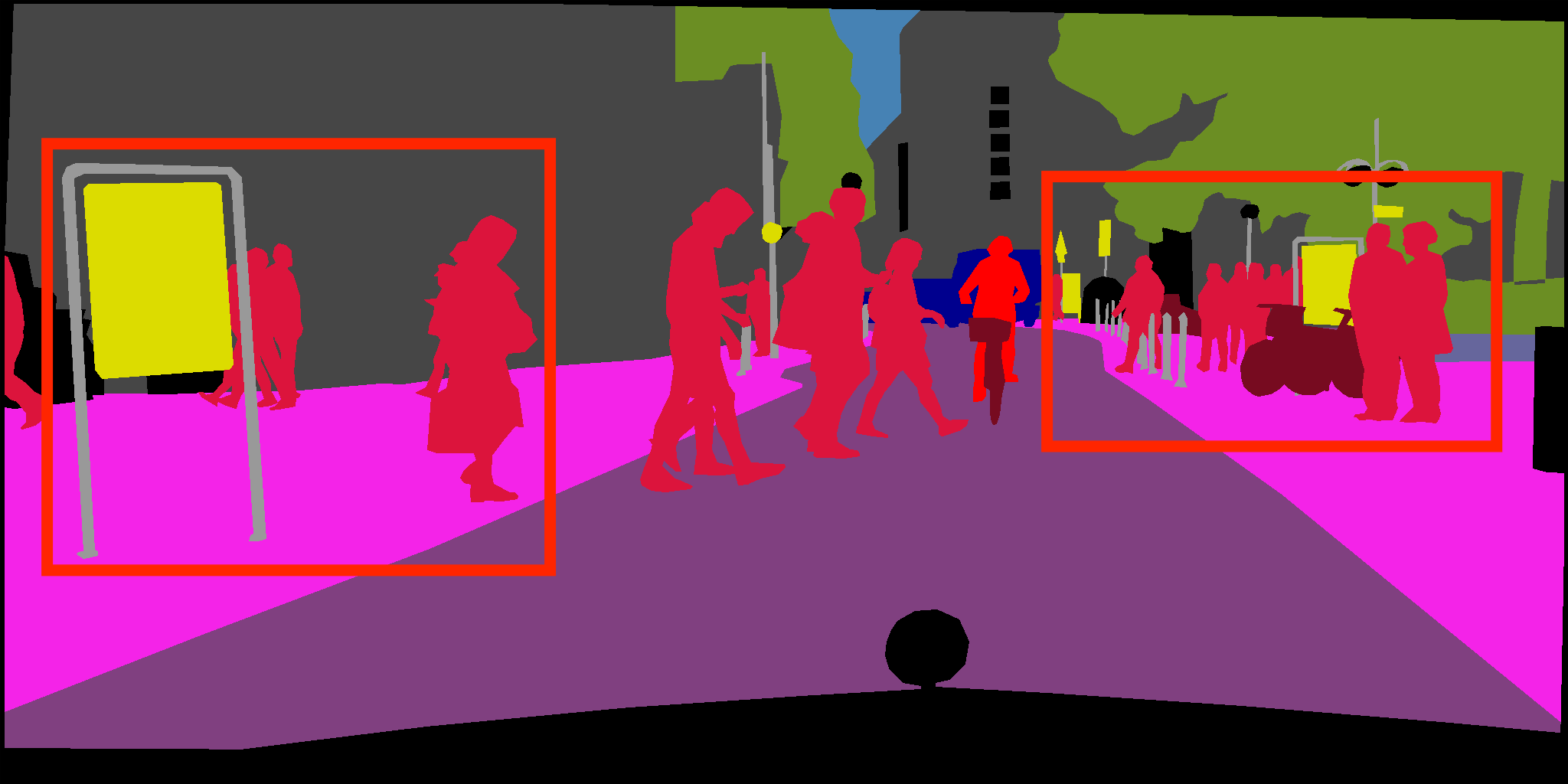}
\end{minipage}
\begin{minipage}{1.1in}
\includegraphics[width=1.1in]{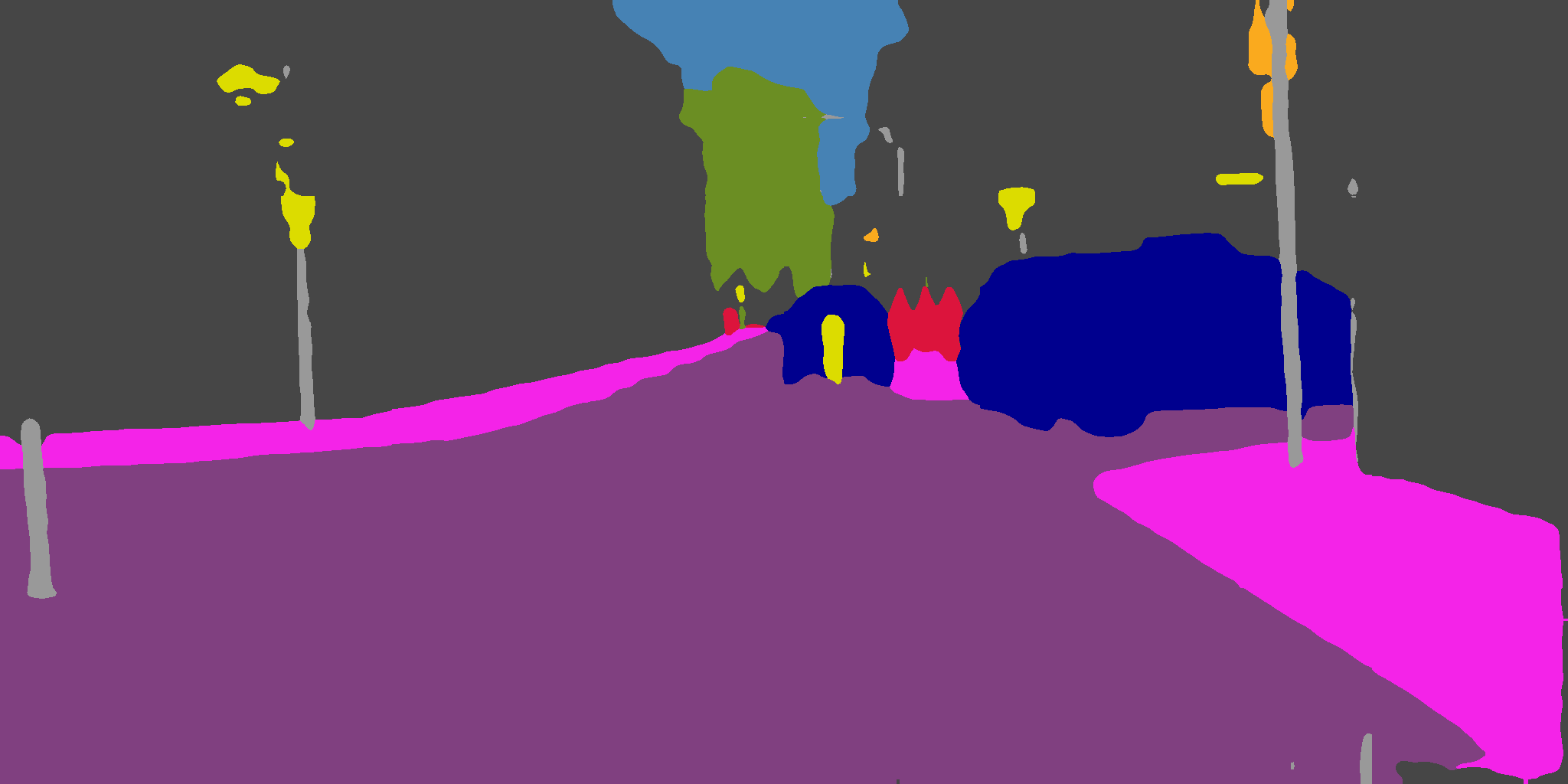}
\end{minipage}
\begin{minipage}{1.1in}
\includegraphics[width=1.1in]{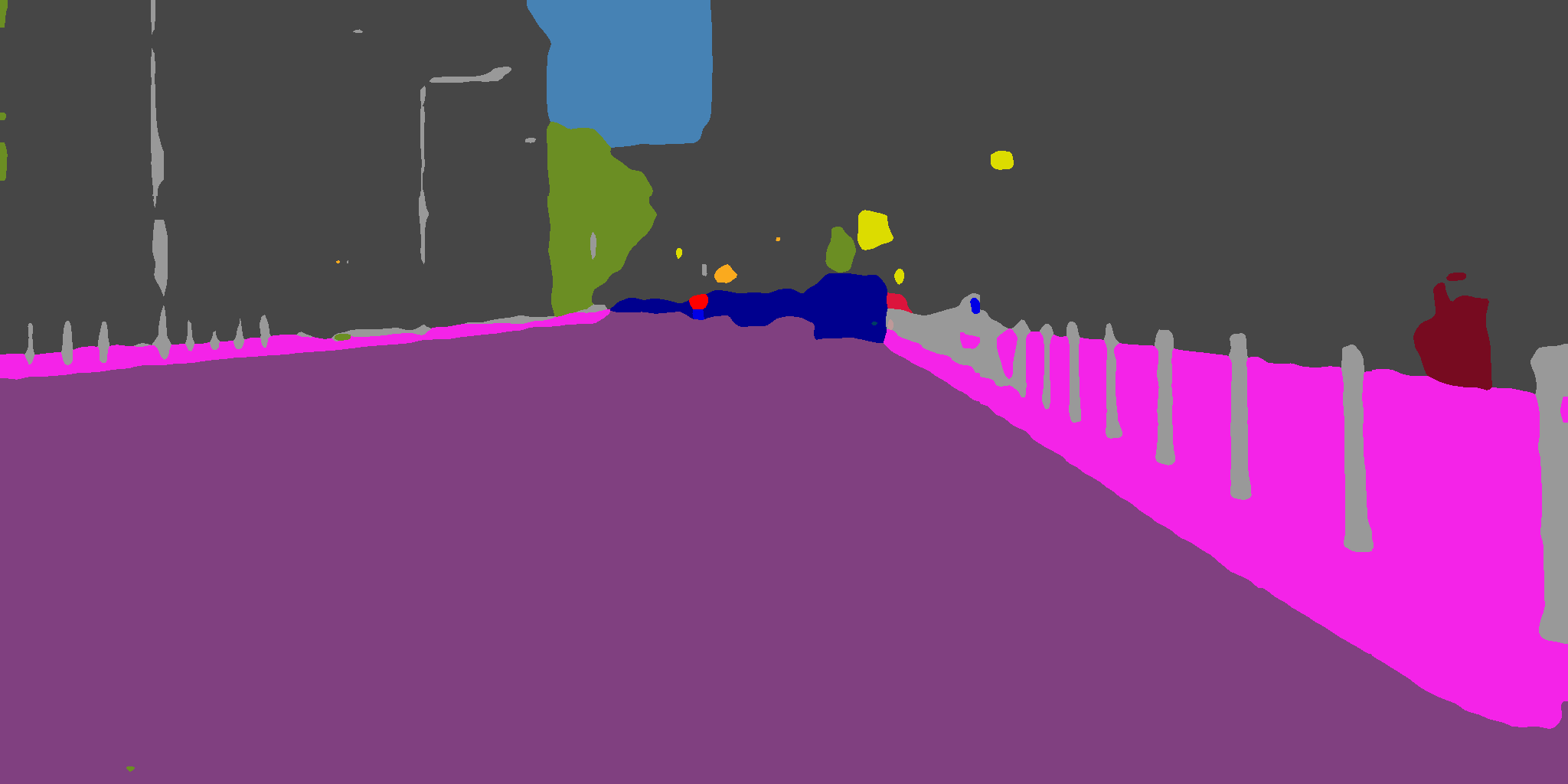}
\end{minipage}
\vspace{0.3em}
\begin{minipage}{1.1in}
\includegraphics[width=1.1in]{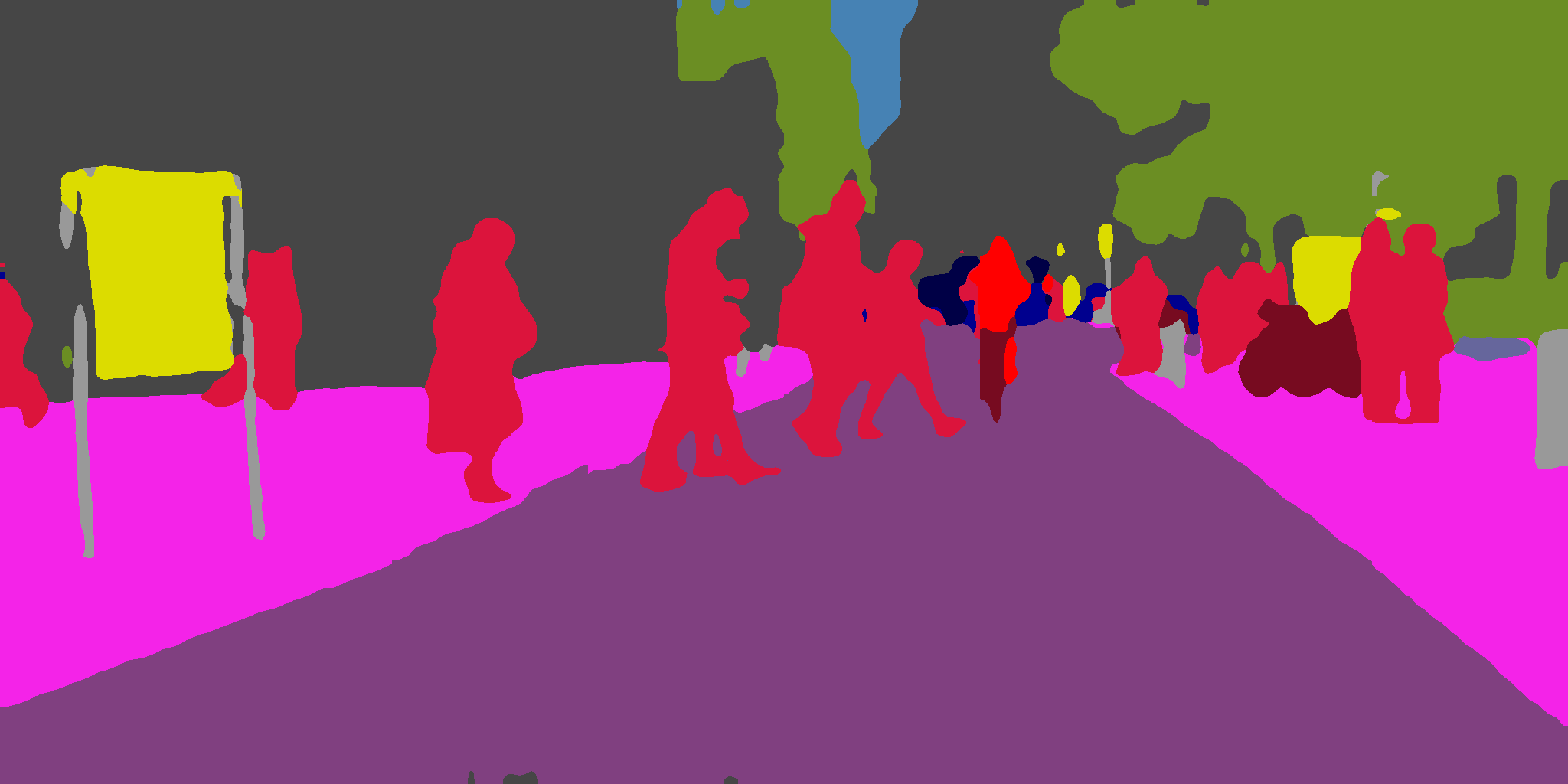}
\end{minipage}
\begin{minipage}{1.1in}
\includegraphics[width=1.1in]{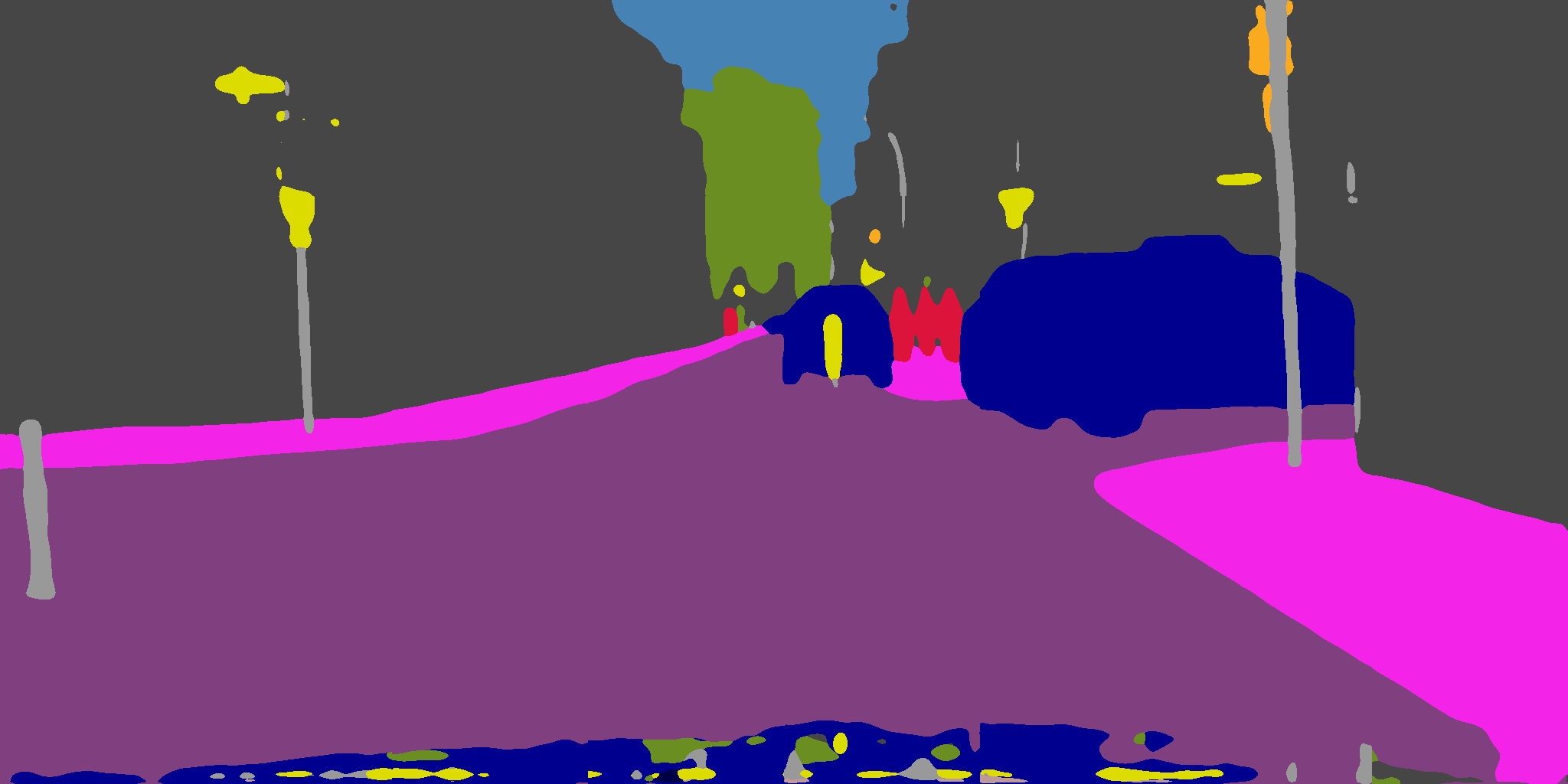}
\end{minipage}
\begin{minipage}{1.1in}
\includegraphics[width=1.1in]{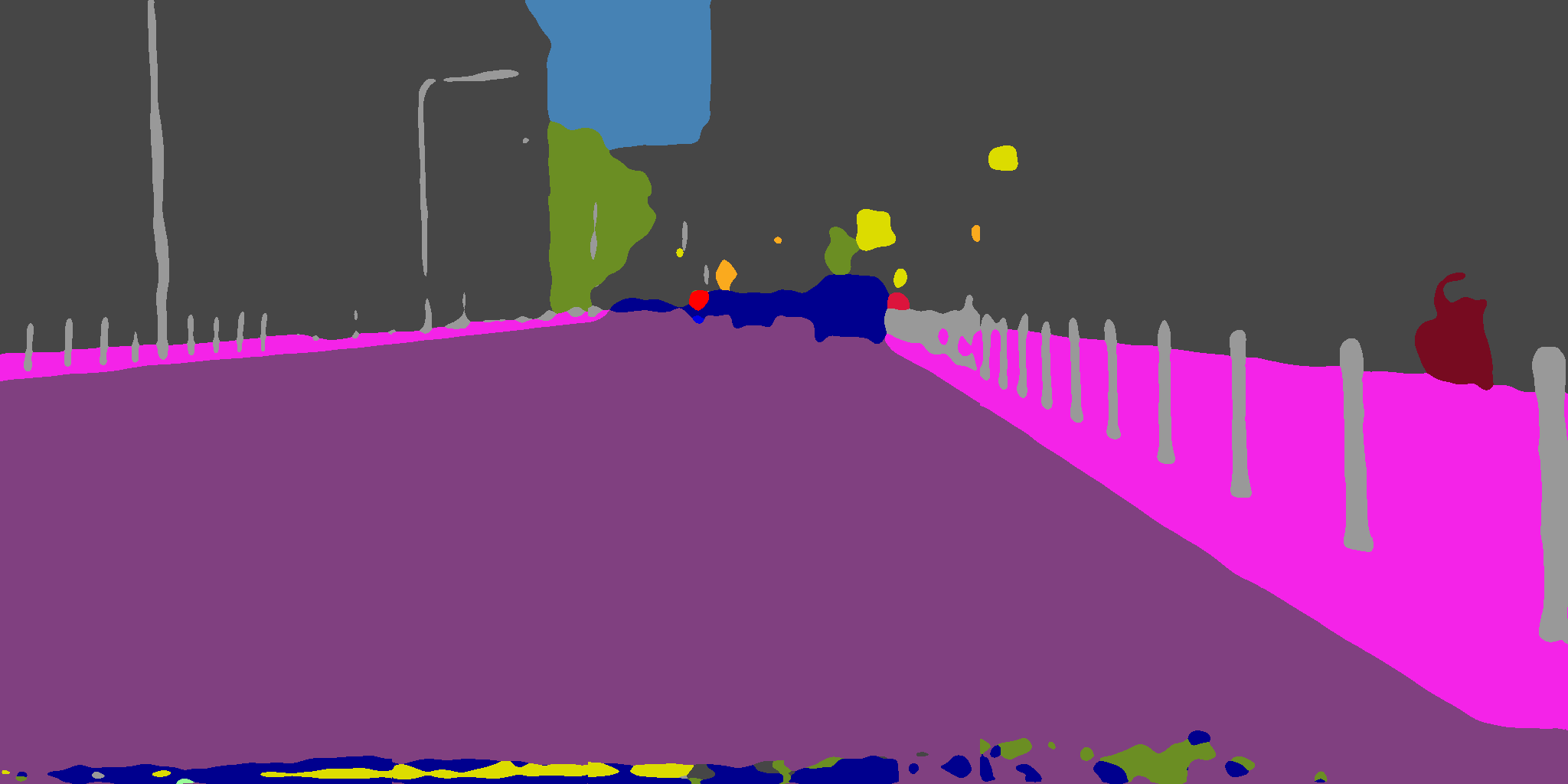}
\end{minipage}
\vspace{0.3em}
\begin{minipage}{1.1in}
\includegraphics[width=1.1in]{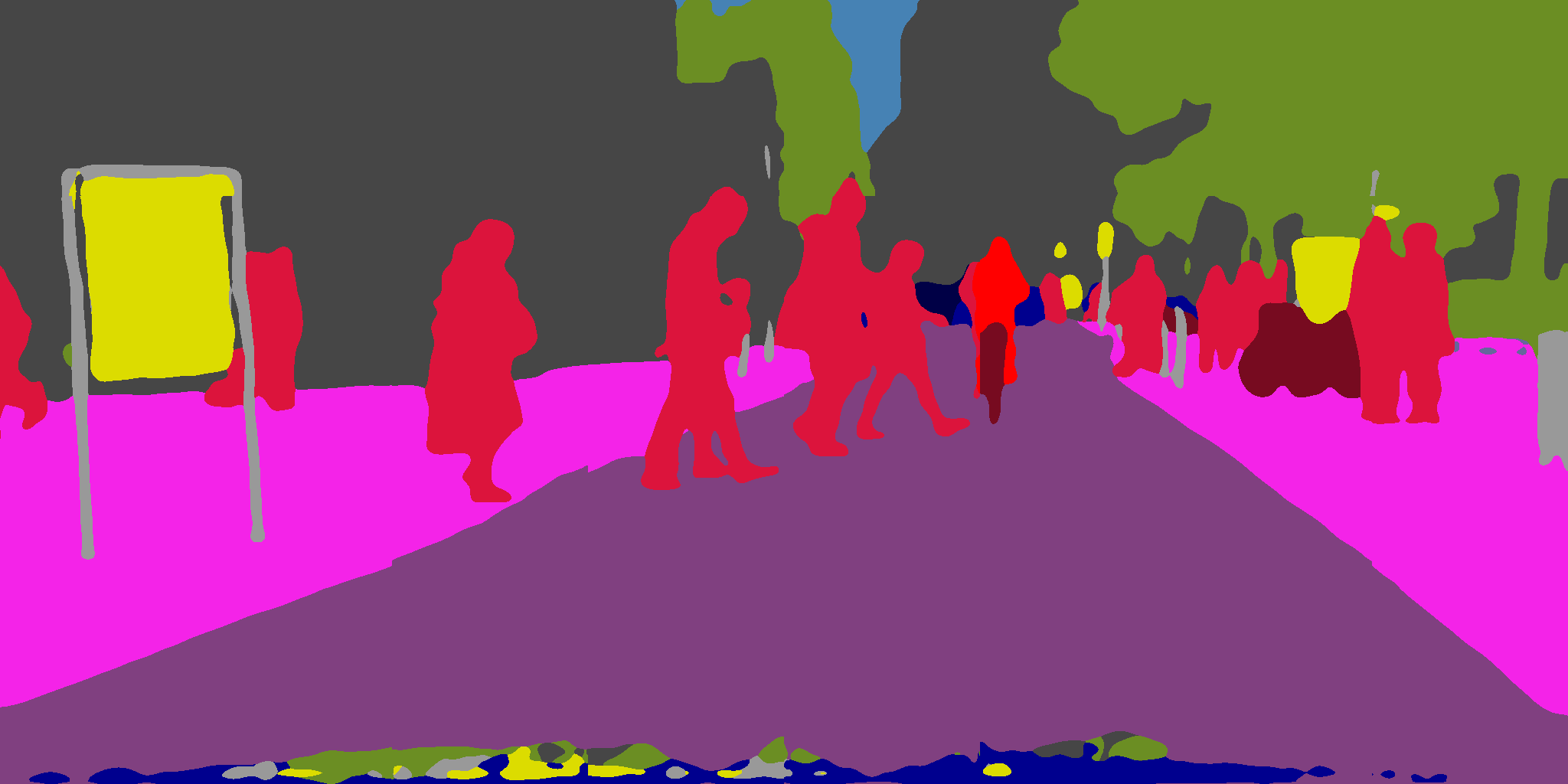}
\end{minipage}
\begin{minipage}{1.1in}
\includegraphics[width=1.1in]{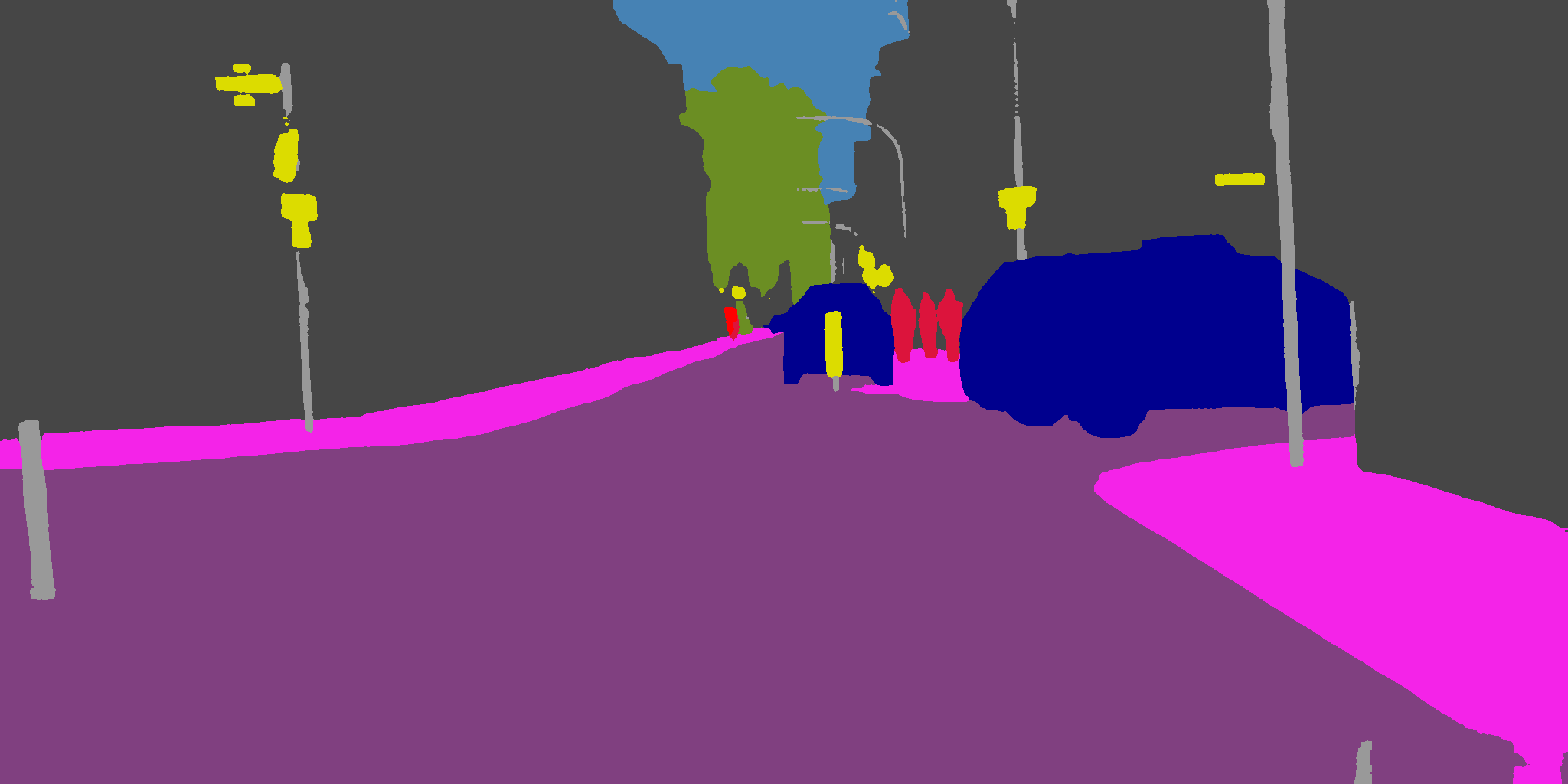}
\subcaption*{(a)}
\end{minipage}
\begin{minipage}{1.1in}
\includegraphics[width=1.1in]{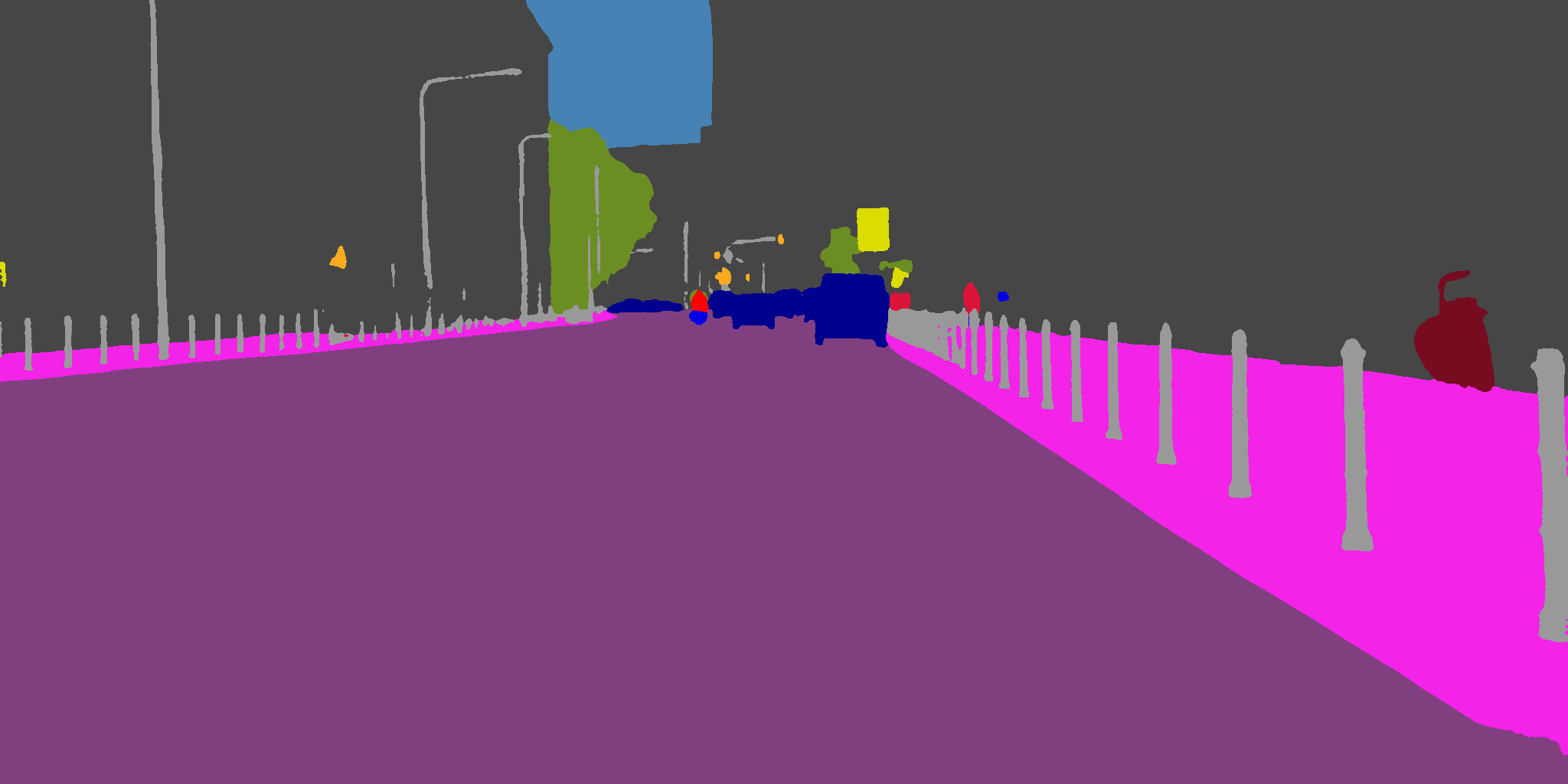}
\subcaption*{(b)}
\end{minipage}
\begin{minipage}{1.1in}
\includegraphics[width=1.1in]{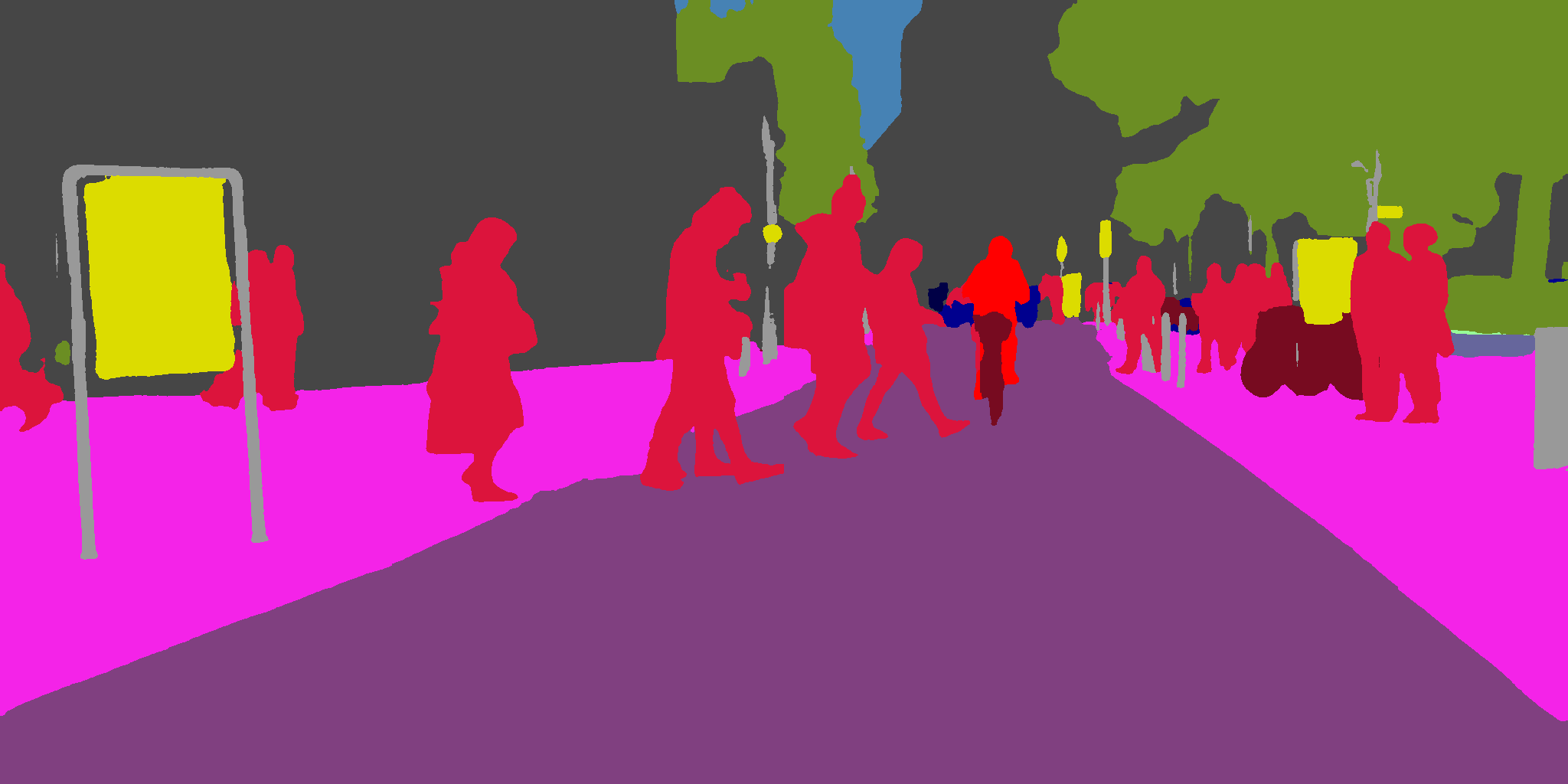}
\subcaption*{(c)}
\end{minipage}
\caption{Visualized segmentation maps on Cityscapes val set. The figures from top to down are input images, ground truths, segmentation maps of SETR-Naive\cite{zheng2021rethinking}, segmentation maps of SETR-PUP\cite{zheng2021rethinking}, and segmentation maps of our method. All the methods use the ViT-L as the backbone. We frame the most improved parts in the figures of ground truths.}
\label{fig11}
\end{figure}

\begin{figure}
\centering
\begin{minipage}{1.1in}
\includegraphics[width=1.1in, height=0.8in]{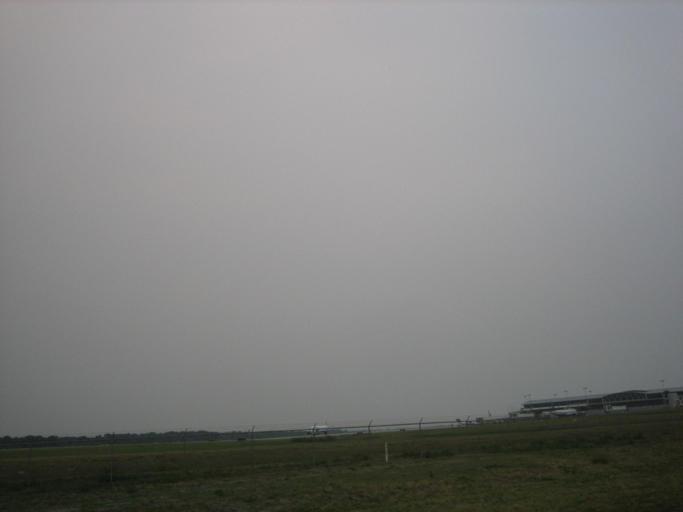}
\end{minipage}
\begin{minipage}{1.1in}
\includegraphics[width=1.1in, height=0.8in]{ADE_val_00000177_img.jpg}
\end{minipage}
\vspace{0.3em}
\begin{minipage}{1.1in}
\includegraphics[width=1.1in, height=0.8in]{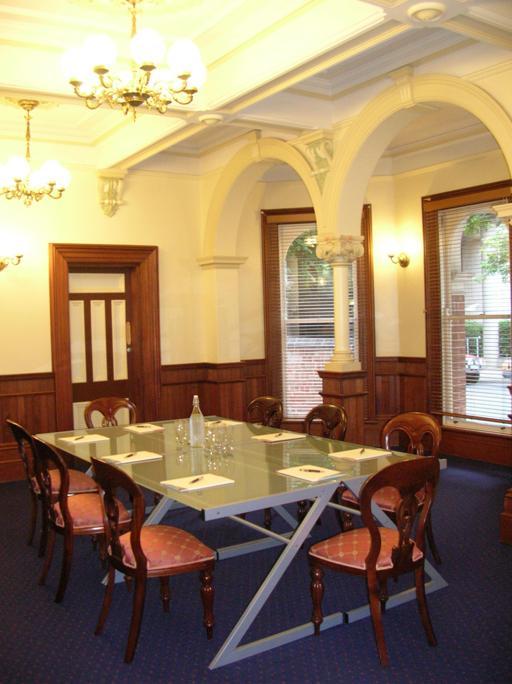}
\end{minipage}
\begin{minipage}{1.1in}
\includegraphics[width=1.1in, height=0.8in]{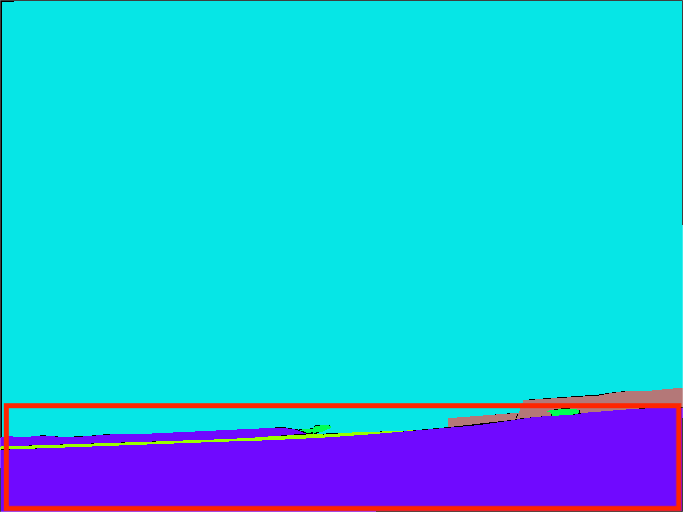}
\end{minipage}
\begin{minipage}{1.1in}
\includegraphics[width=1.1in, height=0.8in]{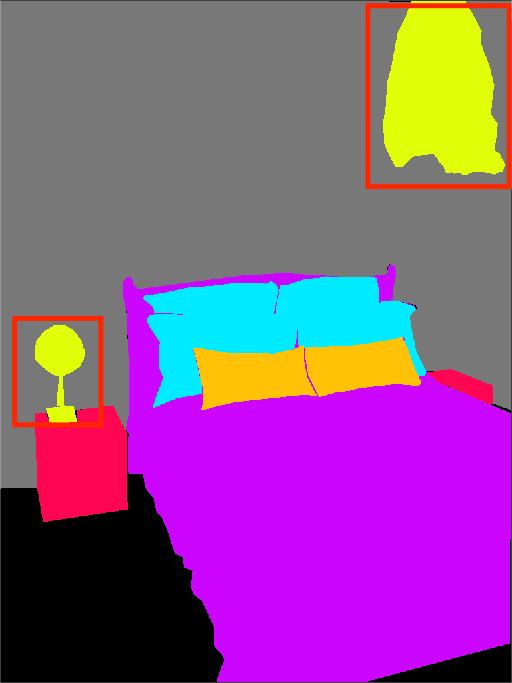}
\end{minipage}
\vspace{0.3em}
\begin{minipage}{1.1in}
\includegraphics[width=1.1in, height=0.8in]{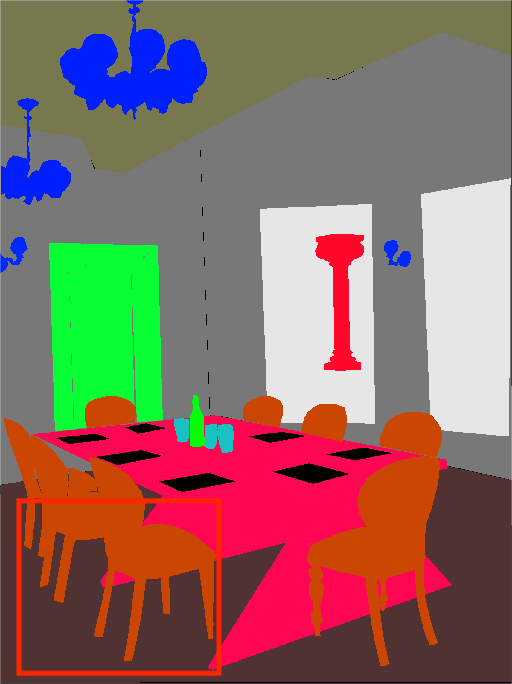}
\end{minipage}
\begin{minipage}{1.1in}
\includegraphics[width=1.1in, height=0.8in]{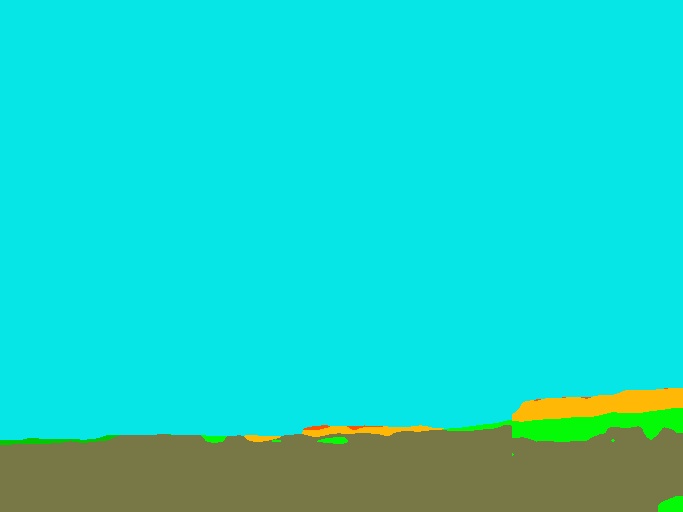}
\end{minipage}
\begin{minipage}{1.1in}
\includegraphics[width=1.1in, height=0.8in]{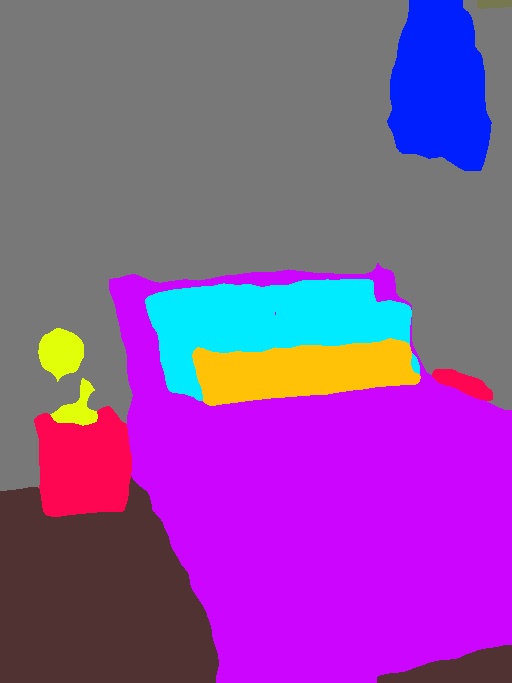}
\end{minipage}
\vspace{0.3em}
\begin{minipage}{1.1in}
\includegraphics[width=1.1in, height=0.8in]{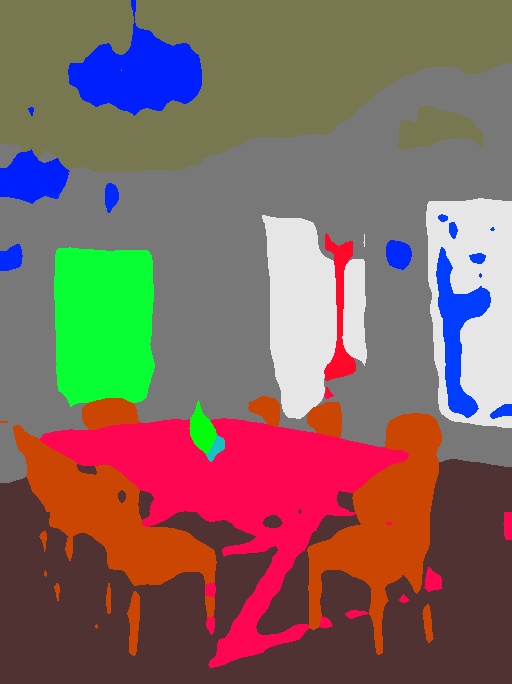}
\end{minipage}
\begin{minipage}{1.1in}
\includegraphics[width=1.1in, height=0.8in]{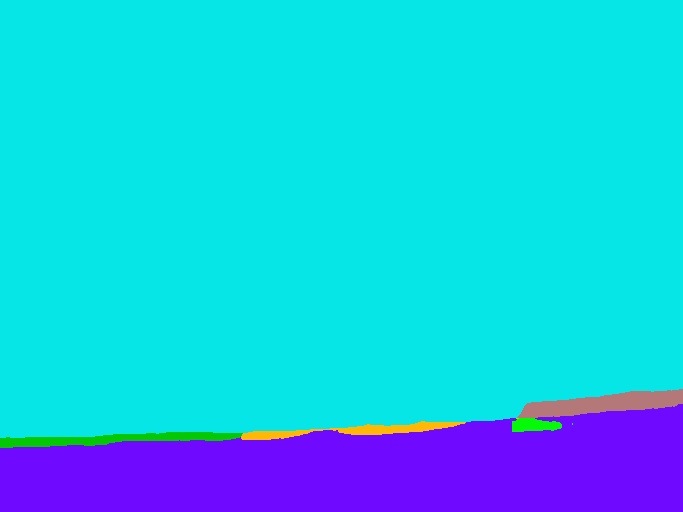}
\subcaption*{(a)}
\end{minipage}
\begin{minipage}{1.1in}
\includegraphics[width=1.1in, height=0.8in]{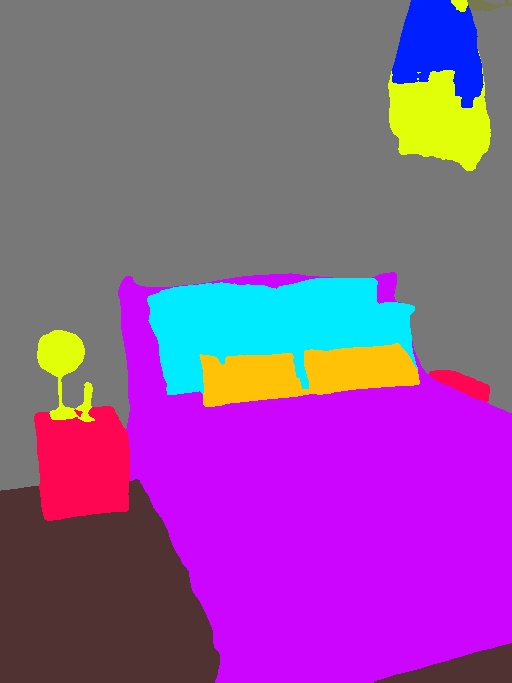}
\subcaption*{(b)}
\end{minipage}
\begin{minipage}{1.1in}
\includegraphics[width=1.1in, height=0.8in]{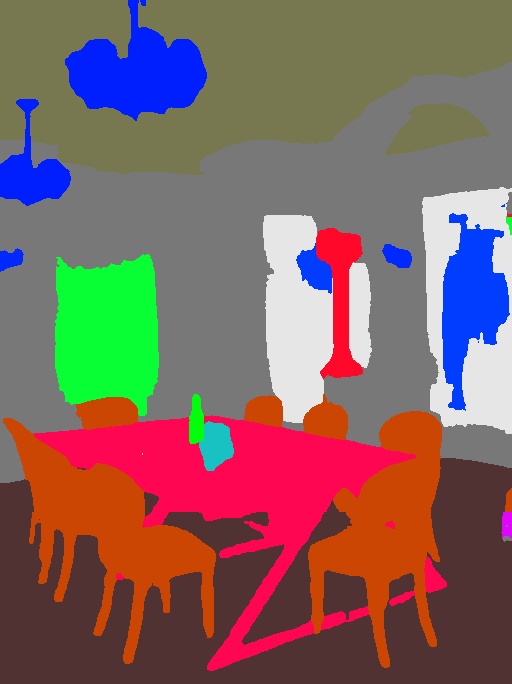}
\subcaption*{(c)}
\end{minipage}
\caption{Visualized segmentation maps on ADE20K val set. The figures from top to down are input images, ground truths, segmentation maps of Segmenter\cite{strudel2021segmenter}, and segmentation maps of our method. All the methods use the ViT-L as the backbone. We frame the most improved parts in the figures of ground truths.}
\label{fig12}
\end{figure}

\begin{figure}
\centering
\begin{minipage}{1.1in}
\includegraphics[width=1.1in, height=0.8in]{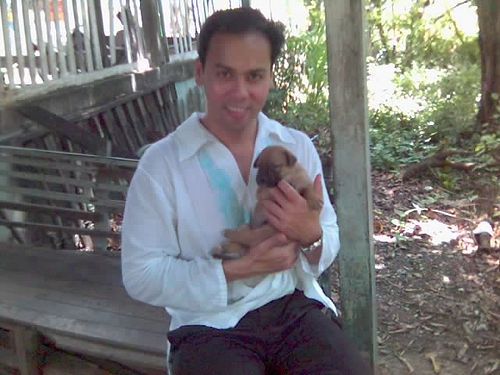}
\end{minipage}
\begin{minipage}{1.1in}
\includegraphics[width=1.1in, height=0.8in]{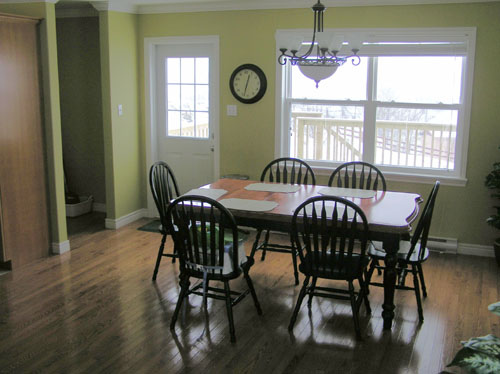}
\end{minipage}
\vspace{0.3em}
\begin{minipage}{1.1in}
\includegraphics[width=1.1in, height=0.8in]{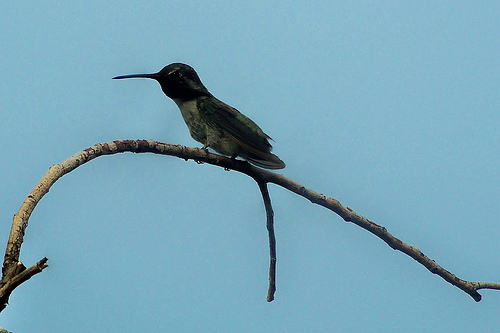}
\end{minipage}
\begin{minipage}{1.1in}
\includegraphics[width=1.1in, height=0.8in]{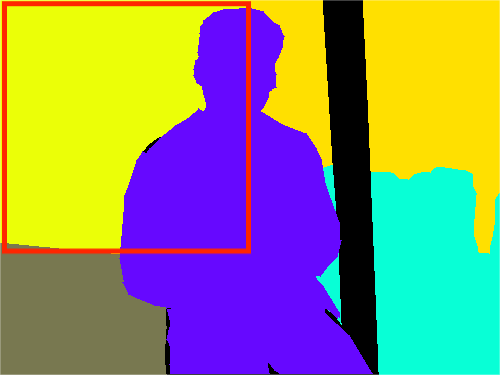}
\end{minipage}
\begin{minipage}{1.1in}
\includegraphics[width=1.1in, height=0.8in]{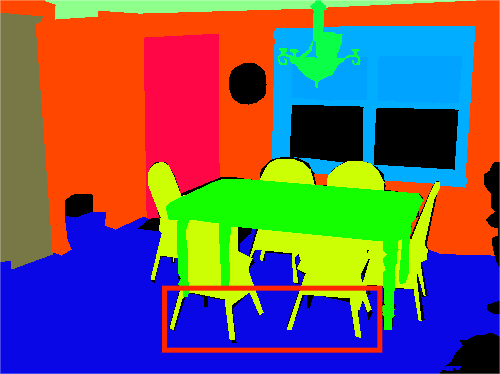}
\end{minipage}
\vspace{0.3em}
\begin{minipage}{1.1in}
\includegraphics[width=1.1in, height=0.8in]{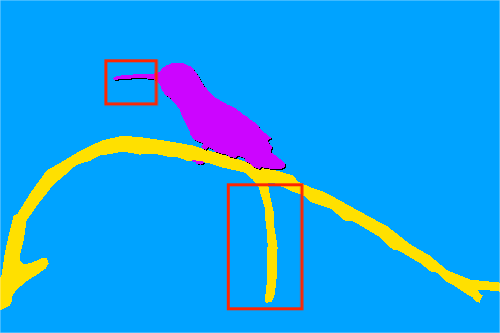}
\end{minipage}
\begin{minipage}{1.1in}
\includegraphics[width=1.1in, height=0.8in]{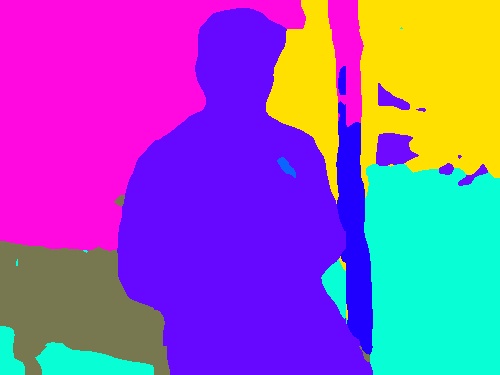}
\end{minipage}
\begin{minipage}{1.1in}
\includegraphics[width=1.1in, height=0.8in]{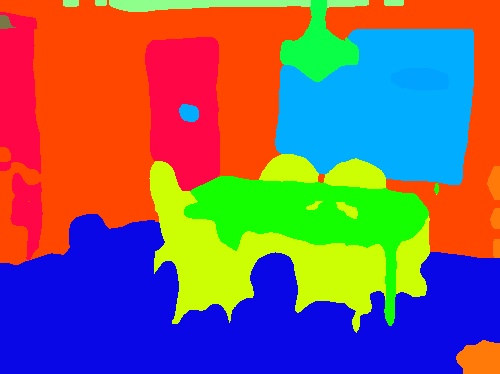}
\end{minipage}
\vspace{0.3em}
\begin{minipage}{1.1in}
\includegraphics[width=1.1in, height=0.8in]{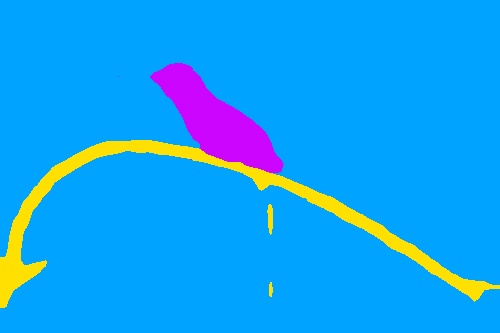}
\end{minipage}
\begin{minipage}{1.1in}
\includegraphics[width=1.1in, height=0.8in]{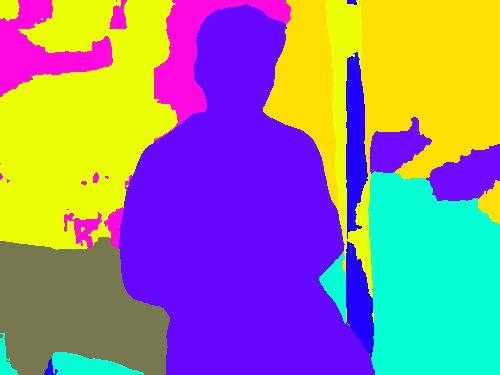}
\subcaption*{(a)}
\end{minipage}
\begin{minipage}{1.1in}
\includegraphics[width=1.1in, height=0.8in]{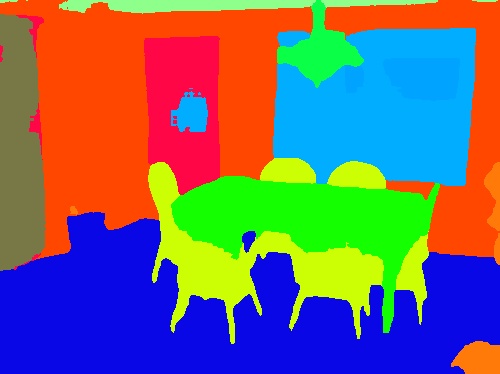}
\subcaption*{(b)}
\end{minipage}
\begin{minipage}{1.1in}
\includegraphics[width=1.1in, height=0.8in]{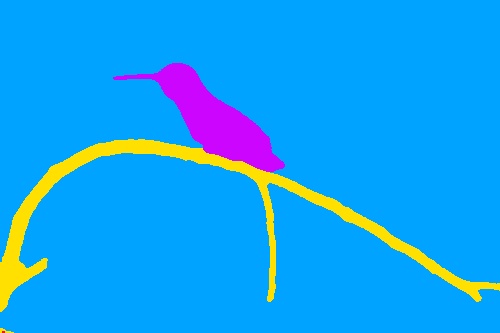}
\subcaption*{(c)}
\end{minipage}
\caption{Visualized segmentation maps on Pascal Context val set. The figures from top to down are input images, ground truths, segmentation maps of Segmenter\cite{strudel2021segmenter}, and segmentation maps of our method. All the methods use the ViT-L as the backbone. We frame the most improved parts in the figures of ground truths.}
\label{fig13}
\end{figure}

We provide the visualized results on Cityscapes, ADE20K, and Pascal Context val set. As shown in Fig. \ref{fig11}, Fig. \ref{fig12}, and Fig. \ref{fig13}, our method not only generates sharper segmentation maps but also recognizes the regions more precisely compared to different methods. The substantial improvements in multiple challenging scenarios demonstrate the generality and robustness of our method.

\section{Conclusions}

In this paper, we introduce an efficient framework of representation separation for semantic segmentation with transformers and propose three novel components for it. Specially, the decoupled two-pathway network keeps the original local discrepancy which the spatially adaptive separation module utilizes to separate the deepest representations. And the discriminative cross-attention generates more discriminative mask embeddings to classify the final patch embeddings.
Integrated with large-scale plain ViTs, our method achieves state-of-the-art accuracy on five popular benchmarks with decently computational efficiency. The visualized results demonstrate that our method learns more discriminative representations and yields sharper segmentation maps. When applied to pyramid ViTs, our method even surpasses the well-designed high-resolution transformers on fine segmentation scenarios.

For a long time, capturing contextual information and enhancing intra-class consistency are the focus of semantic segmentation. We hope that our methods and results can inspire future works to pay more attention to representation separation for taming the highly coherent representations of ViTs in semantic segmentation or other dense prediction tasks. In addition, our method is a feasible solution that generates sharper segmentation maps, and at the same time maintains strong robustness. In the future, we intend to make effort on powerful and practical semantic segmentation models with better segmentation of boundary and small objects, and stronger robustness.


%
\bibliographystyle{ieeetr}
\bibliography{reference}

\begin{thebibliography}{10}

\bibitem{long2015fully}
J.~Long, E.~Shelhamer, and T.~Darrell, ``Fully convolutional networks for
  semantic segmentation,'' in {\em Proceedings of the IEEE Conference on
  Computer Vision and Pattern Recognition}, pp.~3431--3440, 2015.

\bibitem{chen2017deeplab}
L.-C. Chen, G.~Papandreou, I.~Kokkinos, K.~Murphy, and A.~L. Yuille, ``Deeplab:
  Semantic image segmentation with deep convolutional nets, atrous convolution,
  and fully connected crfs,'' {\em IEEE Transactions on Pattern Analysis and
  Machine Intelligence}, vol.~40, no.~4, pp.~834--848, 2017.

\bibitem{chen2017rethinking}
L.-C. Chen, G.~Papandreou, F.~Schroff, and H.~Adam, ``Rethinking atrous
  convolution for semantic image segmentation,'' {\em arXiv preprint
  arXiv:1706.05587}, 2017.

\bibitem{liu2015parsenet}
W.~Liu, A.~Rabinovich, and A.~C. Berg, ``Parsenet: Looking wider to see
  better,'' {\em arXiv preprint arXiv:1506.04579}, 2015.

\bibitem{zhao2017pyramid}
H.~Zhao, J.~Shi, X.~Qi, X.~Wang, and J.~Jia, ``Pyramid scene parsing network,''
  in {\em Proceedings of the IEEE Conference on Computer Vision and Pattern
  Recognition}, pp.~2881--2890, 2017.

\bibitem{peng2017large}
C.~Peng, X.~Zhang, G.~Yu, G.~Luo, and J.~Sun, ``Large kernel matters--improve
  semantic segmentation by global convolutional network,'' in {\em Proceedings
  of the IEEE conference on computer vision and pattern recognition},
  pp.~4353--4361, 2017.

\bibitem{fu2019dual}
J.~Fu, J.~Liu, H.~Tian, Y.~Li, Y.~Bao, Z.~Fang, and H.~Lu, ``Dual attention
  network for scene segmentation,'' in {\em Proceedings of the IEEE/CVF
  Conference on Computer Vision and Pattern Recognition}, pp.~3146--3154, 2019.

\bibitem{yuan2021ocnet}
Y.~Yuan, L.~Huang, J.~Guo, C.~Zhang, X.~Chen, and J.~Wang, ``Ocnet: Object
  context for semantic segmentation,'' {\em International Journal of Computer
  Vision}, pp.~1--24, 2021.

\bibitem{yin2020disentangled}
M.~Yin, Z.~Yao, Y.~Cao, X.~Li, Z.~Zhang, S.~Lin, and H.~Hu, ``Disentangled
  non-local neural networks,'' in {\em European Conference on Computer Vision},
  pp.~191--207, Springer, 2020.

\bibitem{vaswani2017attention}
A.~Vaswani, N.~Shazeer, N.~Parmar, J.~Uszkoreit, L.~Jones, A.~N. Gomez,
  L.~Kaiser, and I.~Polosukhin, ``Attention is all you need,'' {\em arXiv
  preprint arXiv:1706.03762}, 2017.

\bibitem{dosovitskiy2020image}
A.~Dosovitskiy, L.~Beyer, A.~Kolesnikov, D.~Weissenborn, X.~Zhai,
  T.~Unterthiner, M.~Dehghani, M.~Minderer, G.~Heigold, S.~Gelly, {\em et~al.},
  ``An image is worth 16x16 words: Transformers for image recognition at
  scale,'' {\em arXiv preprint arXiv:2010.11929}, 2020.

\bibitem{touvron2020training}
H.~Touvron, M.~Cord, M.~Douze, F.~Massa, A.~Sablayrolles, and H.~J{\'e}gou,
  ``Training data-efficient image transformers \& distillation through
  attention,'' {\em arXiv preprint arXiv:2012.12877}, 2020.

\bibitem{wang2021pyramid}
W.~Wang, E.~Xie, X.~Li, D.-P. Fan, K.~Song, D.~Liang, T.~Lu, P.~Luo, and
  L.~Shao, ``Pyramid vision transformer: A versatile backbone for dense
  prediction without convolutions,'' {\em arXiv preprint arXiv:2102.12122},
  2021.

\bibitem{liu2021swin}
Z.~Liu, Y.~Lin, Y.~Cao, H.~Hu, Y.~Wei, Z.~Zhang, S.~Lin, and B.~Guo, ``Swin
  transformer: Hierarchical vision transformer using shifted windows,'' {\em
  arXiv preprint arXiv:2103.14030}, 2021.

\bibitem{zheng2021rethinking}
S.~Zheng, J.~Lu, H.~Zhao, X.~Zhu, Z.~Luo, Y.~Wang, Y.~Fu, J.~Feng, T.~Xiang,
  P.~H. Torr, {\em et~al.}, ``Rethinking semantic segmentation from a
  sequence-to-sequence perspective with transformers,'' in {\em Proceedings of
  the IEEE/CVF Conference on Computer Vision and Pattern Recognition},
  pp.~6881--6890, 2021.

\bibitem{xie2021segformer}
E.~Xie, W.~Wang, Z.~Yu, A.~Anandkumar, J.~M. Alvarez, and P.~Luo, ``Segformer:
  Simple and efficient design for semantic segmentation with transformers,''
  {\em arXiv preprint arXiv:2105.15203}, 2021.

\bibitem{ranftl2021vision}
R.~Ranftl, A.~Bochkovskiy, and V.~Koltun, ``Vision transformers for dense
  prediction,'' {\em arXiv preprint arXiv:2103.13413}, 2021.

\bibitem{strudel2021segmenter}
R.~Strudel, R.~Garcia, I.~Laptev, and C.~Schmid, ``Segmenter: Transformer for
  semantic segmentation,'' {\em arXiv preprint arXiv:2105.05633}, 2021.

\bibitem{yan2022lawin}
H.~Yan, C.~Zhang, and M.~Wu, ``Lawin transformer: Improving semantic
  segmentation transformer with multi-scale representations via large window
  attention,'' {\em arXiv preprint arXiv:2201.01615}, 2022.

\bibitem{jain2021semask}
J.~Jain, A.~Singh, N.~Orlov, Z.~Huang, J.~Li, S.~Walton, and H.~Shi, ``Semask:
  Semantically masked transformers for semantic segmentation,'' {\em arXiv
  preprint arXiv:2112.12782}, 2021.

\bibitem{lin2022structtoken}
F.~Lin, Z.~Liang, J.~He, M.~Zheng, S.~Tian, and K.~Chen, ``Structtoken:
  Rethinking semantic segmentation with structural prior,'' {\em arXiv preprint
  arXiv:2203.12612}, 2022.

\bibitem{chen2022vision}
Z.~Chen, Y.~Duan, W.~Wang, J.~He, T.~Lu, J.~Dai, and Y.~Qiao, ``Vision
  transformer adapter for dense predictions,'' {\em arXiv preprint
  arXiv:2205.08534}, 2022.

\bibitem{park2021vision}
N.~Park and S.~Kim, ``How do vision transformers work?,'' in {\em International
  Conference on Learning Representations}, 2021.

\bibitem{gong2021vision}
C.~Gong, D.~Wang, M.~Li, V.~Chandra, and Q.~Liu, ``Vision transformers with
  patch diversification,'' {\em arXiv preprint arXiv:2104.12753}, 2021.

\bibitem{bao2021beit}
H.~Bao, L.~Dong, and F.~Wei, ``Beit: Bert pre-training of image transformers,''
  {\em arXiv preprint arXiv:2106.08254}, 2021.

\bibitem{steiner2021train}
A.~Steiner, A.~Kolesnikov, X.~Zhai, R.~Wightman, J.~Uszkoreit, and L.~Beyer,
  ``How to train your vit? data, augmentation, and regularization in vision
  transformers,'' {\em arXiv preprint arXiv:2106.10270}, 2021.

\bibitem{chu2021conditional}
X.~Chu, Z.~Tian, B.~Zhang, X.~Wang, X.~Wei, H.~Xia, and C.~Shen, ``Conditional
  positional encodings for vision transformers,'' {\em arXiv preprint
  arXiv:2102.10882}, 2021.

\bibitem{chu2021twins}
X.~Chu, Z.~Tian, Y.~Wang, B.~Zhang, H.~Ren, X.~Wei, H.~Xia, and C.~Shen,
  ``Twins: Revisiting the design of spatial attention in vision transformers,''
  {\em arXiv preprint arXiv:2104.13840}, 2021.

\bibitem{dai2021coatnet}
Z.~Dai, H.~Liu, Q.~V. Le, and M.~Tan, ``Coatnet: Marrying convolution and
  attention for all data sizes,'' {\em arXiv preprint arXiv:2106.04803}, 2021.

\bibitem{dong2021cswin}
X.~Dong, J.~Bao, D.~Chen, W.~Zhang, N.~Yu, L.~Yuan, D.~Chen, and B.~Guo,
  ``Cswin transformer: A general vision transformer backbone with cross-shaped
  windows,'' in {\em Proceedings of the IEEE/CVF Conference on Computer Vision
  and Pattern Recognition}, pp.~12124--12134, 2022.

\bibitem{yuan2022volo}
L.~Yuan, Q.~Hou, Z.~Jiang, J.~Feng, and S.~Yan, ``Volo: Vision outlooker for
  visual recognition,'' {\em IEEE Transactions on Pattern Analysis and Machine
  Intelligence}, 2022.

\bibitem{yuan2021hrformer}
Y.~Yuan, R.~Fu, L.~Huang, W.~Lin, C.~Zhang, X.~Chen, and J.~Wang, ``Hrformer:
  High-resolution transformer for dense prediction,'' {\em arXiv preprint
  arXiv:2110.09408}, 2021.

\bibitem{gu2021multi}
J.~Gu, H.~Kwon, D.~Wang, W.~Ye, M.~Li, Y.-H. Chen, L.~Lai, V.~Chandra, and
  D.~Z. Pan, ``Multi-scale high-resolution vision transformer for semantic
  segmentation,'' {\em arXiv preprint arXiv:2111.01236}, 2021.

\bibitem{touvron2022deit}
H.~Touvron, M.~Cord, and H.~J{\'e}gou, ``Deit iii: Revenge of the vit,'' {\em
  arXiv preprint arXiv:2204.07118}, 2022.

\bibitem{he2021masked}
K.~He, X.~Chen, S.~Xie, Y.~Li, P.~Doll{\'a}r, and R.~Girshick, ``Masked
  autoencoders are scalable vision learners,'' {\em arXiv preprint
  arXiv:2111.06377}, 2021.

\bibitem{xiao2018unified}
T.~Xiao, Y.~Liu, B.~Zhou, Y.~Jiang, and J.~Sun, ``Unified perceptual parsing
  for scene understanding,'' in {\em Proceedings of the European conference on
  computer vision (ECCV)}, pp.~418--434, 2018.

\bibitem{carion2020end}
N.~Carion, F.~Massa, G.~Synnaeve, N.~Usunier, A.~Kirillov, and S.~Zagoruyko,
  ``End-to-end object detection with transformers,'' in {\em European
  Conference on Computer Vision}, pp.~213--229, Springer, 2020.

\bibitem{cheng2021per}
B.~Cheng, A.~Schwing, and A.~Kirillov, ``Per-pixel classification is not all
  you need for semantic segmentation,'' {\em Advances in Neural Information
  Processing Systems}, vol.~34, 2021.

\bibitem{cheng2022masked}
B.~Cheng, I.~Misra, A.~G. Schwing, A.~Kirillov, and R.~Girdhar,
  ``Masked-attention mask transformer for universal image segmentation,'' in
  {\em Proceedings of the IEEE/CVF Conference on Computer Vision and Pattern
  Recognition}, pp.~1290--1299, 2022.

\bibitem{ronneberger2015u}
O.~Ronneberger, P.~Fischer, and T.~Brox, ``U-net: Convolutional networks for
  biomedical image segmentation,'' in {\em International Conference on Medical
  image computing and computer-assisted intervention}, pp.~234--241, Springer,
  2015.

\bibitem{9133304}
Z.~Huang, X.~Wang, Y.~Wei, L.~Huang, H.~Shi, W.~Liu, and T.~S. Huang, ``Ccnet:
  Criss-cross attention for semantic segmentation,'' {\em IEEE Transactions on
  Pattern Analysis and Machine Intelligence}, pp.~1--1, 2020.

\bibitem{9206128}
Y.~Liu, Y.~Chen, P.~Lasang, and Q.~Sun, ``Covariance attention for semantic
  segmentation,'' {\em IEEE Transactions on Pattern Analysis and Machine
  Intelligence}, vol.~44, no.~4, pp.~1805--1818, 2022.

\bibitem{9633155}
Z.~Li, Y.~Sun, L.~Zhang, and J.~Tang, ``Ctnet: Context-based tandem network for
  semantic segmentation,'' {\em IEEE Transactions on Pattern Analysis and
  Machine Intelligence}, vol.~44, no.~12, pp.~9904--9917, 2022.

\bibitem{lin2017refinenet}
G.~Lin, A.~Milan, C.~Shen, and I.~Reid, ``Refinenet: Multi-path refinement
  networks for high-resolution semantic segmentation,'' in {\em Proceedings of
  the IEEE conference on computer vision and pattern recognition},
  pp.~1925--1934, 2017.

\bibitem{kirillov2019panoptic}
A.~Kirillov, R.~Girshick, K.~He, and P.~Doll{\'a}r, ``Panoptic feature pyramid
  networks,'' in {\em Proceedings of the IEEE/CVF conference on computer vision
  and pattern recognition}, pp.~6399--6408, 2019.

\bibitem{li2020semantic}
X.~Li, A.~You, Z.~Zhu, H.~Zhao, M.~Yang, K.~Yang, S.~Tan, and Y.~Tong,
  ``Semantic flow for fast and accurate scene parsing,'' in {\em European
  Conference on Computer Vision}, pp.~775--793, Springer, 2020.

\bibitem{takikawa2019gated}
T.~Takikawa, D.~Acuna, V.~Jampani, and S.~Fidler, ``Gated-scnn: Gated shape
  cnns for semantic segmentation,'' in {\em Proceedings of the IEEE/CVF
  international conference on computer vision}, pp.~5229--5238, 2019.

\bibitem{li2020improving}
X.~Li, X.~Li, L.~Zhang, G.~Cheng, J.~Shi, Z.~Lin, S.~Tan, and Y.~Tong,
  ``Improving semantic segmentation via decoupled body and edge supervision,''
  {\em arXiv preprint arXiv:2007.10035}, 2020.

\bibitem{wang2021active}
C.~Wang, Y.~Zhang, M.~Cui, J.~Liu, P.~Ren, Y.~Yang, X.~Xie, X.~Hua, H.~Bao, and
  W.~Xu, ``Active boundary loss for semantic segmentation,'' {\em arXiv
  preprint arXiv:2102.02696}, 2021.

\bibitem{yuan2020segfix}
Y.~Yuan, J.~Xie, X.~Chen, and J.~Wang, ``Segfix: Model-agnostic boundary
  refinement for segmentation,'' in {\em European Conference on Computer
  Vision}, pp.~489--506, Springer, 2020.

\bibitem{bai2022improving}
J.~Bai, L.~Yuan, S.-T. Xia, S.~Yan, Z.~Li, and W.~Liu, ``Improving vision
  transformers by revisiting high-frequency components,'' {\em arXiv preprint
  arXiv:2204.00993}, 2022.

\bibitem{huang2022car}
Y.~Huang, D.~Kang, L.~Chen, X.~Zhe, W.~Jia, X.~He, and L.~Bao, ``Car:
  Class-aware regularizations for semantic segmentation,'' {\em arXiv preprint
  arXiv:2203.07160}, 2022.

\bibitem{ba2016layer}
J.~L. Ba, J.~R. Kiros, and G.~E. Hinton, ``Layer normalization,'' {\em arXiv
  preprint arXiv:1607.06450}, 2016.

\bibitem{hendrycks2016gaussian}
D.~Hendrycks and K.~Gimpel, ``Gaussian error linear units (gelus),'' {\em arXiv
  preprint arXiv:1606.08415}, 2016.

\bibitem{mmseg2020}
M.~Contributors, ``{MMSegmentation}: Openmmlab semantic segmentation toolbox
  and benchmark,'' 2020.

\bibitem{chen2018encoder}
L.-C. Chen, Y.~Zhu, G.~Papandreou, F.~Schroff, and H.~Adam, ``Encoder-decoder
  with atrous separable convolution for semantic image segmentation,'' in {\em
  Proceedings of the European conference on computer vision (ECCV)},
  pp.~801--818, 2018.

\bibitem{liu2022convnet}
Z.~Liu, H.~Mao, C.-Y. Wu, C.~Feichtenhofer, T.~Darrell, and S.~Xie, ``A convnet
  for the 2020s,'' in {\em Proceedings of the IEEE/CVF Conference on Computer
  Vision and Pattern Recognition}, pp.~11976--11986, 2022.

\bibitem{yu2018bisenet}
C.~Yu, J.~Wang, C.~Peng, C.~Gao, G.~Yu, and N.~Sang, ``Bisenet: Bilateral
  segmentation network for real-time semantic segmentation,'' in {\em
  Proceedings of the European conference on computer vision (ECCV)},
  pp.~325--341, 2018.

\bibitem{yu2021bisenet}
C.~Yu, C.~Gao, J.~Wang, G.~Yu, C.~Shen, and N.~Sang, ``Bisenet v2: Bilateral
  network with guided aggregation for real-time semantic segmentation,'' {\em
  International Journal of Computer Vision}, vol.~129, no.~11, pp.~3051--3068,
  2021.

\bibitem{hong2021deep}
Y.~Hong, H.~Pan, W.~Sun, Y.~Jia, {\em et~al.}, ``Deep dual-resolution networks
  for real-time and accurate semantic segmentation of road scenes,'' {\em arXiv
  preprint arXiv:2101.06085}, 2021.

\bibitem{pang2019towards}
Y.~Pang, Y.~Li, J.~Shen, and L.~Shao, ``Towards bridging semantic gap to
  improve semantic segmentation,'' in {\em Proceedings of the IEEE/CVF
  International Conference on Computer Vision}, pp.~4230--4239, 2019.

\bibitem{huang2021alignseg}
Z.~Huang, Y.~Wei, X.~Wang, W.~Liu, T.~S. Huang, and H.~Shi, ``Alignseg:
  Feature-aligned segmentation networks,'' {\em IEEE Transactions on Pattern
  Analysis and Machine Intelligence}, vol.~44, no.~1, pp.~550--557, 2021.

\bibitem{zhou2021decoupled}
J.~Zhou, V.~Jampani, Z.~Pi, Q.~Liu, and M.-H. Yang, ``Decoupled dynamic filter
  networks,'' in {\em Proceedings of the IEEE/CVF Conference on Computer Vision
  and Pattern Recognition}, pp.~6647--6656, 2021.

\bibitem{mazzini2018guided}
D.~Mazzini, ``Guided upsampling network for real-time semantic segmentation,''
  {\em arXiv preprint arXiv:1807.07466}, 2018.

\bibitem{fan2021rethinking}
M.~Fan, S.~Lai, J.~Huang, X.~Wei, Z.~Chai, J.~Luo, and X.~Wei, ``Rethinking
  bisenet for real-time semantic segmentation,'' in {\em Proceedings of the
  IEEE/CVF conference on computer vision and pattern recognition},
  pp.~9716--9725, 2021.

\bibitem{yuan2020object}
Y.~Yuan, X.~Chen, and J.~Wang, ``Object-contextual representations for semantic
  segmentation,'' in {\em Computer Vision--ECCV 2020: 16th European Conference,
  Glasgow, UK, August 23--28, 2020, Proceedings, Part VI 16}, pp.~173--190,
  Springer, 2020.

\bibitem{cordts2016cityscapes}
M.~Cordts, M.~Omran, S.~Ramos, T.~Rehfeld, M.~Enzweiler, R.~Benenson,
  U.~Franke, S.~Roth, and B.~Schiele, ``The cityscapes dataset for semantic
  urban scene understanding,'' in {\em Proceedings of the IEEE Conference on
  Computer Vision and Pattern Recognition}, pp.~3213--3223, 2016.

\bibitem{zhou2017scene}
B.~Zhou, H.~Zhao, X.~Puig, S.~Fidler, A.~Barriuso, and A.~Torralba, ``Scene
  parsing through ade20k dataset,'' in {\em Proceedings of the IEEE conference
  on computer vision and pattern recognition}, pp.~633--641, 2017.

\bibitem{mottaghi2014role}
R.~Mottaghi, X.~Chen, X.~Liu, N.-G. Cho, S.-W. Lee, S.~Fidler, R.~Urtasun, and
  A.~Yuille, ``The role of context for object detection and semantic
  segmentation in the wild,'' in {\em Proceedings of the IEEE conference on
  computer vision and pattern recognition}, pp.~891--898, 2014.

\bibitem{caesar2018coco}
H.~Caesar, J.~Uijlings, and V.~Ferrari, ``Coco-stuff: Thing and stuff classes
  in context,'' in {\em Proceedings of the IEEE conference on computer vision
  and pattern recognition}, pp.~1209--1218, 2018.

\bibitem{deng2009imagenet}
J.~Deng, W.~Dong, R.~Socher, L.-J. Li, K.~Li, and L.~Fei-Fei, ``Imagenet: A
  large-scale hierarchical image database,'' in {\em 2009 IEEE conference on
  computer vision and pattern recognition}, pp.~248--255, Ieee, 2009.

\bibitem{zhang2018context}
H.~Zhang, K.~Dana, J.~Shi, Z.~Zhang, X.~Wang, A.~Tyagi, and A.~Agrawal,
  ``Context encoding for semantic segmentation,'' in {\em Proceedings of the
  IEEE conference on Computer Vision and Pattern Recognition}, pp.~7151--7160,
  2018.

\bibitem{wang2020deep}
J.~Wang, K.~Sun, T.~Cheng, B.~Jiang, C.~Deng, Y.~Zhao, D.~Liu, Y.~Mu, M.~Tan,
  X.~Wang, {\em et~al.}, ``Deep high-resolution representation learning for
  visual recognition,'' {\em IEEE transactions on pattern analysis and machine
  intelligence}, vol.~43, no.~10, pp.~3349--3364, 2020.

\bibitem{gu2021hrvit}
J.~Gu, H.~Kwon, D.~Wang, W.~Ye, M.~Li, Y.-H. Chen, L.~Lai, V.~Chandra, and
  D.~Z. Pan, ``Hrvit: Multi-scale high-resolution vision transformer,'' {\em
  arXiv preprint arXiv:2111.01236}, 2021.

\bibitem{zhao2018psanet}
H.~Zhao, Y.~Zhang, S.~Liu, J.~Shi, C.~C. Loy, D.~Lin, and J.~Jia, ``Psanet:
  Point-wise spatial attention network for scene parsing,'' in {\em Proceedings
  of the European Conference on Computer Vision (ECCV)}, pp.~267--283, 2018.

\bibitem{yu2020context}
C.~Yu, J.~Wang, C.~Gao, G.~Yu, C.~Shen, and N.~Sang, ``Context prior for scene
  segmentation,'' in {\em Proceedings of the IEEE/CVF Conference on Computer
  Vision and Pattern Recognition}, pp.~12416--12425, 2020.

\bibitem{zhu2021learning}
L.~Zhu, D.~Ji, S.~Zhu, W.~Gan, W.~Wu, and J.~Yan, ``Learning statistical
  texture for semantic segmentation,'' in {\em Proceedings of the IEEE/CVF
  Conference on Computer Vision and Pattern Recognition}, pp.~12537--12546,
  2021.

\bibitem{bousselham2021efficient}
W.~Bousselham, G.~Thibault, L.~Pagano, A.~Machireddy, J.~Gray, Y.~H. Chang, and
  X.~Song, ``Efficient self-ensemble framework for semantic segmentation,''
  {\em arXiv preprint arXiv:2111.13280}, 2021.

\bibitem{zhang2022segvit}
B.~Zhang, Z.~Tian, Q.~Tang, X.~Chu, X.~Wei, C.~Shen, and Y.~Liu, ``Segvit:
  Semantic segmentation with plain vision transformers,'' {\em arXiv preprint
  arXiv:2210.05844}, 2022.

\bibitem{he2022rankseg}
H.~He, Y.~Yuan, X.~Yue, and H.~Hu, ``Rankseg: Adaptive pixel classification
  with image category ranking for segmentation,'' in {\em European Conference
  on Computer Vision}, pp.~682--700, Springer, 2022.

\bibitem{cui2022region}
J.~Cui, Y.~Yuan, Z.~Zhong, Z.~Tian, H.~Hu, S.~Lin, and J.~Jia, ``Region
  rebalance for long-tailed semantic segmentation,'' {\em arXiv preprint
  arXiv:2204.01969}, 2022.

\bibitem{huang2021channelized}
Y.~Huang, D.~Kang, W.~Jia, X.~He, and L.~Liu, ``Channelized axial attention for
  semantic segmentation--considering channel relation within spatial attention
  for semantic segmentation,'' {\em arXiv preprint arXiv:2101.07434}, 2021.

\bibitem{kamann2020benchmarking}
C.~Kamann and C.~Rother, ``Benchmarking the robustness of semantic segmentation
  models,'' in {\em Proceedings of the IEEE/CVF Conference on Computer Vision
  and Pattern Recognition}, pp.~8828--8838, 2020.

\bibitem{zhou2022understanding}
D.~Zhou, Z.~Yu, E.~Xie, C.~Xiao, A.~Anandkumar, J.~Feng, and J.~M. Alvarez,
  ``Understanding the robustness in vision transformers,'' in {\em
  International Conference on Machine Learning}, pp.~27378--27394, PMLR, 2022.

\bibitem{bhojanapalli2021understanding}
S.~Bhojanapalli, A.~Chakrabarti, D.~Glasner, D.~Li, T.~Unterthiner, and
  A.~Veit, ``Understanding robustness of transformers for image
  classification,'' in {\em Proceedings of the IEEE/CVF International
  Conference on Computer Vision}, pp.~10231--10241, 2021.

\bibitem{mahmood2021robustness}
K.~Mahmood, R.~Mahmood, and M.~Van~Dijk, ``On the robustness of vision
  transformers to adversarial examples,'' in {\em Proceedings of the IEEE/CVF
  International Conference on Computer Vision}, pp.~7838--7847, 2021.

\bibitem{paul2022vision}
S.~Paul and P.-Y. Chen, ``Vision transformers are robust learners,'' in {\em
  Proceedings of the AAAI Conference on Artificial Intelligence}, vol.~36,
  pp.~2071--2081, 2022.

\bibitem{mao2022towards}
X.~Mao, G.~Qi, Y.~Chen, X.~Li, R.~Duan, S.~Ye, Y.~He, and H.~Xue, ``Towards
  robust vision transformer,'' in {\em Proceedings of the IEEE/CVF Conference
  on Computer Vision and Pattern Recognition}, pp.~12042--12051, 2022.

\end{thebibliography}

\end{document}